\pdfoutput=1
\documentclass[11pt]{article}
\usepackage[table,xcdraw]{xcolor}
\usepackage{tikz}
\definecolor{peiGreen}{RGB}{235, 250, 235} % 极浅的薄荷绿
\definecolor{peiRed}{RGB}{250, 235, 235}   % 极浅的樱花红
\definecolor{headerGray}{RGB}{240, 240, 240} % 表头浅灰
\usepackage{acl}
\usepackage{times}
\usepackage{latexsym}

\usepackage[T1]{fontenc}
\usepackage[table,xcdraw]{xcolor}

\usepackage{booktabs}

\usepackage{pifont}

\definecolor{cGreen}{HTML}{2ECC71}  % 翠绿 (比纯绿更护眼)
\definecolor{cRed}{HTML}{E74C3C}    % 柔和红
\definecolor{cGray}{HTML}{95A5A6}   % 灰色
\definecolor{rowGray}{HTML}{F9F9F9} % 极淡的斑马纹背景色
\definecolor{colHighlight}{HTML}{EAF2F8} % PEI列的高亮背景色(淡蓝)

\newcommand{\cmark}{\textcolor{cGreen}{\ding{51}}}
\newcommand{\xmark}{\textcolor{cRed}{\ding{55}}}
\usepackage[utf8]{inputenc}

\usepackage{microtype}

\usepackage{inconsolata}
\usepackage{booktabs} % 用于绘制美观的三线表
\usepackage{graphicx} % 用于 \resizebox 缩放表格
\usepackage{amssymb} % For \checkmark and \times symbols

\usepackage{bm}
\usepackage{array}
\usepackage{arydshln}
\usepackage{xspace}
\usepackage{mathrsfs}
\usepackage{amssymb}
\usepackage{amsmath}
\usepackage{enumitem}
\usepackage{booktabs}
\usepackage{graphicx}
\usepackage{multirow}
\usepackage{multicol}
\usepackage{makecell}
\usepackage{listings}
\usepackage{subfigure}
\usepackage{subcaption}
\usepackage{tablefootnote}
\usepackage[normalem]{ulem}
\usepackage{algorithm}
\usepackage{algpseudocode}

\usepackage{nicematrix}
\usepackage{booktabs}
\usepackage[most]{tcolorbox}
\usepackage{color,colortbl}
\definecolor{color1}{HTML}{006EB8}
\definecolor{color2}{HTML}{009B55}
\definecolor{color3}{HTML}{00A99A}
\definecolor{color4}{HTML}{3C8031}
\definecolor{color5}{HTML}{006795}
\definecolor{color6}{HTML}{00AEB3}
\definecolor{grayrow}{gray}{0.95}

\setlist[itemize]{leftmargin=5mm, itemsep=0mm}

\definecolor{comment}{rgb}{0,0.6,0}
\definecolor{key}{rgb}{0.84, 0.44, 0.84}
\definecolor{func}{rgb}{0.16, 0.72, 0.86}
\definecolor{var}{rgb}{1, 0.33, 0.33}
\definecolor{true}{rgb}{1, 0.78, 0.02}

\newcommand{\method}{SlidesGen-Bench\xspace}

\newcommand{\etc}{\emph{etc.}\xspace}

\title{SlidesGen-Bench: Evaluating  Slides Generation via Computational and Quantitative Metrics}

\author{%
    Yunqiao Yang$^{1}$\thanks{Equal contribution.} \quad Wenbo Li$^{1}$\footnotemark[1] \quad Houxing Ren$^{1}$\footnotemark[1]\thanks{Project Lead.} \quad Zimu Lu$^{1}$ \quad Ke Wang$^{1}$ \\ \quad \textbf{Zhiyuan Huang}$^{2}$ \quad
    \textbf{Zhuofan Zong}$^{1}$ \quad \textbf{Mingjie Zhan}$^{2}$\footnotemark[3] \quad \textbf{Hongsheng Li}$^{1,3,4}$ \thanks{Corresponding author.} \\
    $^{1}$CUHK MMLab, $^{2}$SenseTime Research \\ $^{3}$CPII under InnoHK, $^{4}$Shanghai AI Laboratory  \\
    \texttt{yangyunqiao7@gmail.com} \quad \texttt{zhanmingjie@sensetime.com} \quad \texttt{hsli@ee.cuhk.edu.hk}
}

\begin{document}
\maketitle
\begin{abstract}

The rapid evolution of Large Language Models (LLMs) has fostered diverse paradigms for automated slide generation, ranging from template-based file generation and code-driven layouts to image-centric synthesis. However, evaluating these heterogeneous systems remains challenging, as existing protocols often struggle to provide comparable scores across architectures or rely on uncalibrated judgments. In this paper, we introduce \textbf{SlidesGen-Bench}, a structured evaluation protocol that assesses slide generation along three practically important dimensions: \textit{content fidelity}, \textit{visual aesthetics}, and \textit{editability}. To compare heterogeneous systems under one framework, all outputs are projected into the visual domain for content and aesthetic assessment, while editability is evaluated through format-conditional structural parsing of the generated files. Each dimension is quantified with structured, verifiable pipelines rather than monolithic LLM judgment. For the highly subjective aesthetic dimensions, we construct the Slides-Align1.5k human preference dataset and show that our computational metrics align with human judgment substantially better than standard LLM-as-a-Judge baselines. Our code and data are released to support further research.

\end{abstract}
\begin{figure*}[ht]
    \centering
    \includegraphics[width=0.84\linewidth]{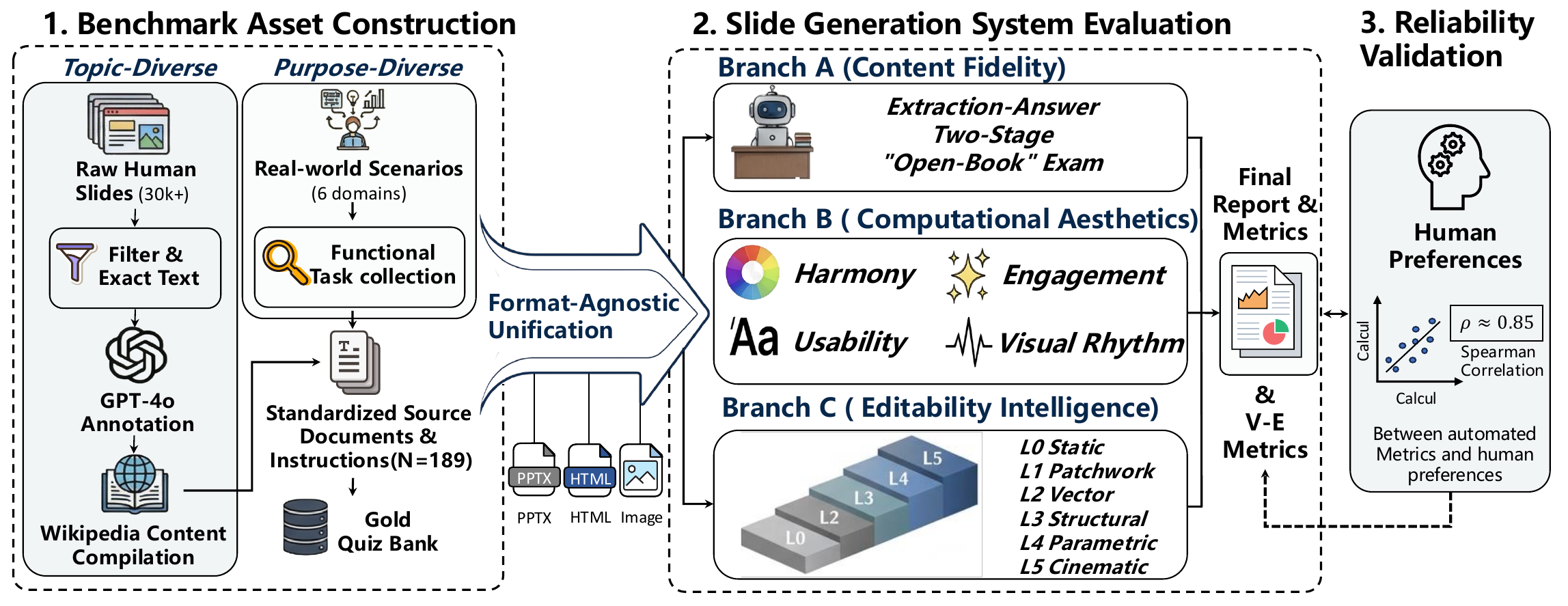}
    \caption{The main pipeline of \method.}
    \label{fig: pipeline}
    \vspace{-0.5cm}
\end{figure*}

\section{Introduction} \label{sec: intro}

Driven by recent breakthroughs in Large Language Models (LLMs)~\cite{GPT42023ABS230308774, LLama2023ABS230213971, LLama22023ABS230709288, Llama312024Dubey, Qwen2023ABS230916609, Mistral2023ABS231006825, Claude2023Anthropic, Qwen252024Yang} and the rapid evolution of code agents~\cite{wang2024executable,dong2025survey, wang2024opendevin}, researchers have increasingly explored the use of LLM-based agents for automated slide generation. This surge in interest has led to a wide array of research initiatives~\cite{ge2025autopresent, zheng2025pptagent, tang2025slidecoder, jung2025talk, yang2025auto, bandyopadhyay2024enhancing} and commercial applications. 
Current presentation generation approaches typically follow one of three paradigms: (1) Template-based Generation, which populates pre-defined schemas; (2) Code-driven Layouts, which utilize generated markup to render slide structures; and (3) Image-centric Generation, which synthesizes entire slide pages directly.

To evaluate automatically generated slides, prior work has proposed several pipelines that generally fall into two categories: reference-based comparisons~\cite{ge2025autopresent} and LLM-as-a-Judge frameworks~\cite{zheng2025pptagent}. Reference-based metrics rely heavily on source files or target images, restricting their applicability in open-ended generation scenarios. Conversely, while LLM-based evaluation offers flexibility, it is susceptible to inherent model behaviors---such as verbosity bias~\cite{saito2023verbosity,wang2024large} or reasoning failures~\cite{krumdick2025no,thakur2025judging}. The significant diversity in system architectures and output formats introduces three core challenges for systematic assessment: (1) achieving comparable evaluation across diverse architectures; (2) conducting a comprehensive and quantitative assessment; and (3) verifying the reliability and rationality of the evaluation scheme.

% To evaluate automatically generated slides, prior work has proposed several pipelines that generally fall into two categories: reference-based comparisons and LLM-as-a-Judge frameworks. Reference-based methods, such as components of AutoPresent~\cite{ge2025autopresent}, assess generated slides by comparing them with reference files or target images. Meanwhile, LLM-based evaluation frameworks, adopted by PPTAgent~\cite{zheng2025pptagent} and AutoPresent, leverage large language models to judge slide quality across multiple dimensions.

% Despite their effectiveness in certain settings, these approaches exhibit notable limitations. Reference-based metrics rely heavily on source files or target images, which restricts their applicability in open-ended generation scenarios where ground truth is unavailable. Conversely, while LLM-based evaluation offers flexibility, it is susceptible to inherent model behaviors---such as verbosity bias~\cite{saito2023verbosity,wang2024large} or reasoning failures~\cite{krumdick2025no,thakur2025judging}. Existing slide generation systems exhibit significant diversity in their underlying architectures and output formats, which introduces several key challenges for systematic assessment. Specifically, three core challenges must be addressed:
% (1) achieving comparable evaluation across diverse architectures;
% (2) conducting a comprehensive and quantitative assessment; and
% (3) verifying the reliability and rationality of the evaluation scheme.

To this end, we introduce \textbf{\method}, a benchmark designed to quantitatively evaluate slide generation by tackling these challenges. First, we leverage the observation that the final output of the existing slide generator is a visual rendering. By grounding our aesthetic and content evaluation in the image domain, our framework establishes a comparable baseline across paradigms, while carefully maintaining a format-conditional hierarchy to evaluate underlying structural editability. Second, we evaluate slides across Content, Aesthetics, and Editability using structured, verifiable pipelines. Rather than relying on a single monolithic LLM prompt~\cite{zheng2025pptagent}, we decouple content evaluation into a structured Vision-Language Model (VLM) extraction and question-answering pipeline, and ground aesthetic evaluation in cognitive and visual ergonomics formulas. Finally, to validate the reliability of our scheme, we apply a divide-and-conquer strategy. While the Content and Editability branches are grounded in a structured pipeline and verifiable structural metadata---yielding systematic and quantifiable metrics that enable consistent comparisons across disparate models---we address the inherent subjectivity of aesthetics through human alignment. Specifically, we incorporate human alignment studies on Slides-Align1.5k. This human preference-aligned dataset covers slides from nine mainstream generation systems across seven scenarios. Our experiments demonstrate a high consistency between our automated aesthetic metrics and human ratings. The whole pipeline is shown in Figure~\ref{fig: pipeline}.

In summary, our contributions are as follows:
\begin{itemize}[leftmargin=5mm, itemsep=0mm]
    \item We propose \method, a structured benchmark covering content fidelity, visual aesthetics, and editability. By grounding content and aesthetic evaluation in the visual domain and evaluating editability through format-conditional structural parsing, the protocol yields comparable scores across template-based, code-driven, and image-centric systems under one framework.

    \item We construct Slides-Align1.5k, a human preference-aligned dataset covering slides from nine mainstream generation systems, bridging the gap between automated metrics and perceptual quality.

    \item Comprehensive experiments demonstrate that our structured evaluation pipeline aligns with human preferences substantially better than standard LLM-as-a-judge protocols, with robustness analyses (bootstrap confidence intervals, oracle-context QA, and style-transfer validation) supporting the reliability of the reported scores.
\end{itemize}

\section{\method} \label{sec: met}
% In this section, we first present the Data Collection procedure, followed by the Content and Aesthetics Evaluation pipeline. Finally, we introduce the presentation editability intelligence evaluation framework.
To achieve comparable evaluation across diverse architectures (Challenge 1) and quantitative assessment (Challenge 2), \method projects all outputs into consistent visual renderings and structural schemas before evaluation. Concretely, content and aesthetic evaluation operate on rendered visual outputs, while editability is measured through format-conditional structural parsing of the generated files, allowing template-based PPTX, HTML/code-generation, and image-centric systems to be compared under one framework. The pipeline comprises data curation (Sec. 2.1) and three evaluation branches (Sec. 2.2-2.4). Furthermore, to ensure the reliability of these metrics (Challenge 3), we systematically validate them against human preferences (Sec. 2.5).

\subsection{Data collection} \label{subsec: instruction}

We curate our benchmark based on two primary dimensions: \textit{topics} and \textit{purposes}.
Topic-based instructions assess the model's ability to process diverse content types and generate relevant multi-modal elements, such as images and illustrations. Purpose-based instructions evaluate the capacity to control stylistic coherence and produce visually appealing presentations tailored to specific scenarios and audiences. Constructing the dataset through this dual lens enables a systematic examination of a generative tool's comprehension of requirements and its proficiency in slide creation.

\paragraph{Determining the Data Format.}
We choose document-based inputs (comprehensive outlines with required text and images) as our primary data format to test the models' long-context summarization and formatting stability.

\paragraph{Topic-Diverse Instruction Collection.}
To ensure instructional diversity, we initially collected approximately 30k human-authored slides and templates from various sources. We applied a length filter to exclude decks shorter than 5 pages or longer than 40 pages. We employed GPT-4o to annotate the slides regarding their topics and purposes. The detailed topic distribution analysis is provided in Appendix~\ref{app:distribution}.
Subsequently, we compiled detailed source documents for each topic to serve as input. We leveraged Wikipedia to gather core content, including detailed text, hierarchical subtitles, and image descriptions, while preserving the original structural integrity. Filtering invalid or sparse entries yielded a final set of 94 instructions.

\begin{figure}[t]
    \centering
    \includegraphics[width=0.9\linewidth]{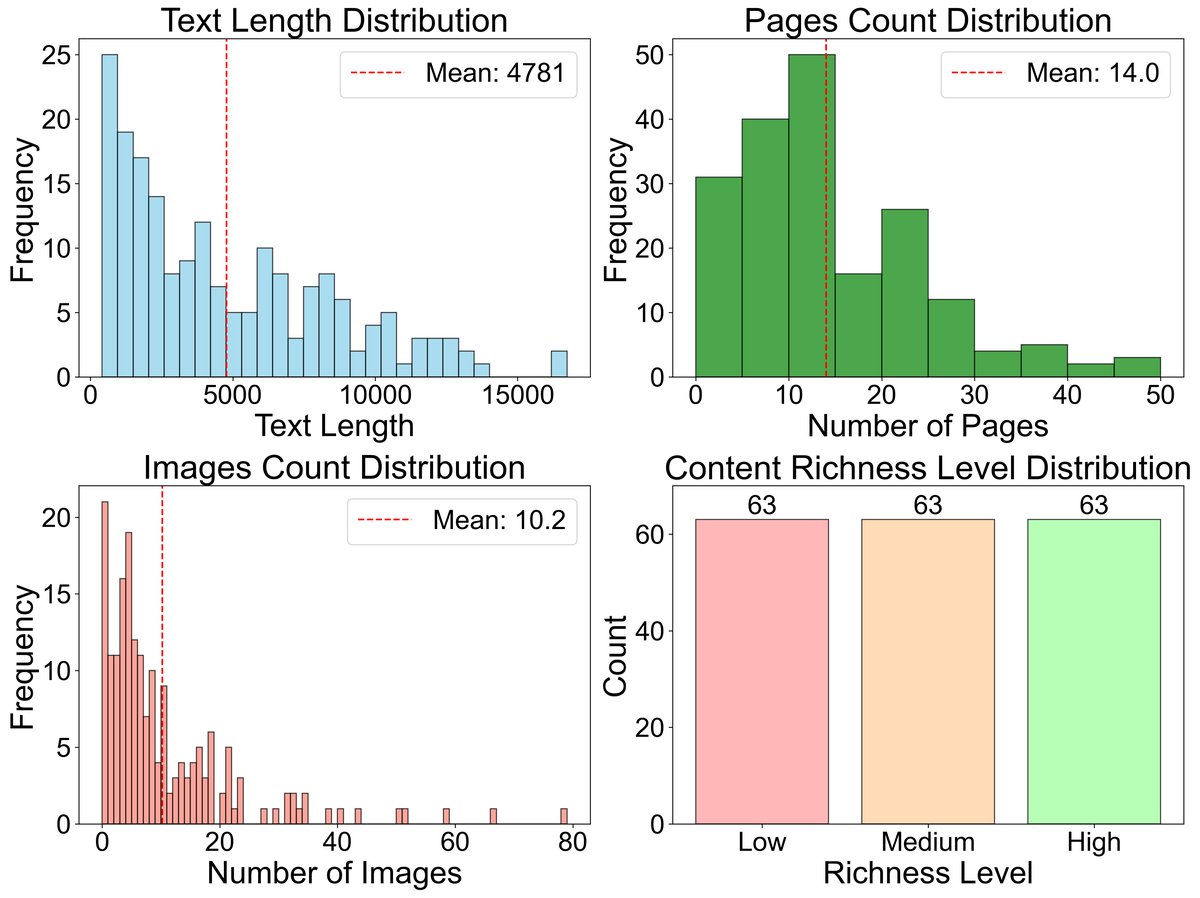}
    \caption{Statistical distribution of content metrics. }
    \label{fig:content_dist}
    \vspace{-0.3cm}
\end{figure}

\paragraph{Purpose-Diverse Instruction Collection.}
The practical utility of automated slide generation depends on its ability to adapt to different contexts. To evaluate this, we expanded our benchmark to look at functional tasks, not just different topics. We collected another 95 instructions across six real-world scenarios: brand promotion, business plans, course preparation, personal statements, product launches, and work reports. These categories cover a wide range of goals, from the narrative style of personal statements to the logical structure of business plans. While the final benchmark consists of 189 filtered instructions, each is a highly complex, document-level synthesis task (mean: 4,525 chars, requiring multi-modal image integration; Figure~\ref{fig:content_dist}).

\paragraph{Statistics of the Instructions.}

As illustrated in Figure~\ref{fig:content_dist}, the curated dataset exhibits significant diversity in text length (mean: 4,525 chars), page count, and image count. To quantitatively categorize difficulty, we introduce a Content Richness metric, stratifying the dataset into three balanced levels (Low, Medium, High). Detailed formulations and distributions are provided in Appendix~\ref{app: richness}. While the final benchmark consists of $N=189$ filtered instructions, each input represents a highly complex, document-level synthesis task. Although this scale might appear modest compared to standard text-only NLP benchmarks, it is a deliberate design choice driven by the high operational cost and complexity of end-to-end multimodal slide generation.  A detailed justification is provided in Appendix \ref{sec:dataset_size_justification}.

\subsection{Content Evaluation} \label{subsec: quiz}

To assess salient information retention, we propose a structured VLM-based QuizBank evaluation framework. Given that presentation slides represent a condensed abstraction of source documents, ensuring critical information preservation is a primary challenge. Rather than relying on a single black-box LLM prompt to judge "content quality"~\cite{zheng2025pptagent}, we construct a reference QuizBank derived directly from the source text. We then employ a decoupled Vision-Language Model (VLM) pipeline to extract text from generated slides and answer the questions. While this approach fundamentally relies on the capabilities of the underlying VLM, decoupling extraction from reasoning minimizes hallucination artifacts inherent in standard LLM-as-a-judge approaches.

\paragraph{QuizBank Construction.}
The construction of the QuizBank is a multi-agent pipeline designed to extract ground-truth knowledge with high precision. As illustrated in Figure \ref{fig:quizbank_pipeline}, the process moves through three distinct phases:

\begin{figure}[t]
    \centering
    \includegraphics[width=0.92\linewidth]{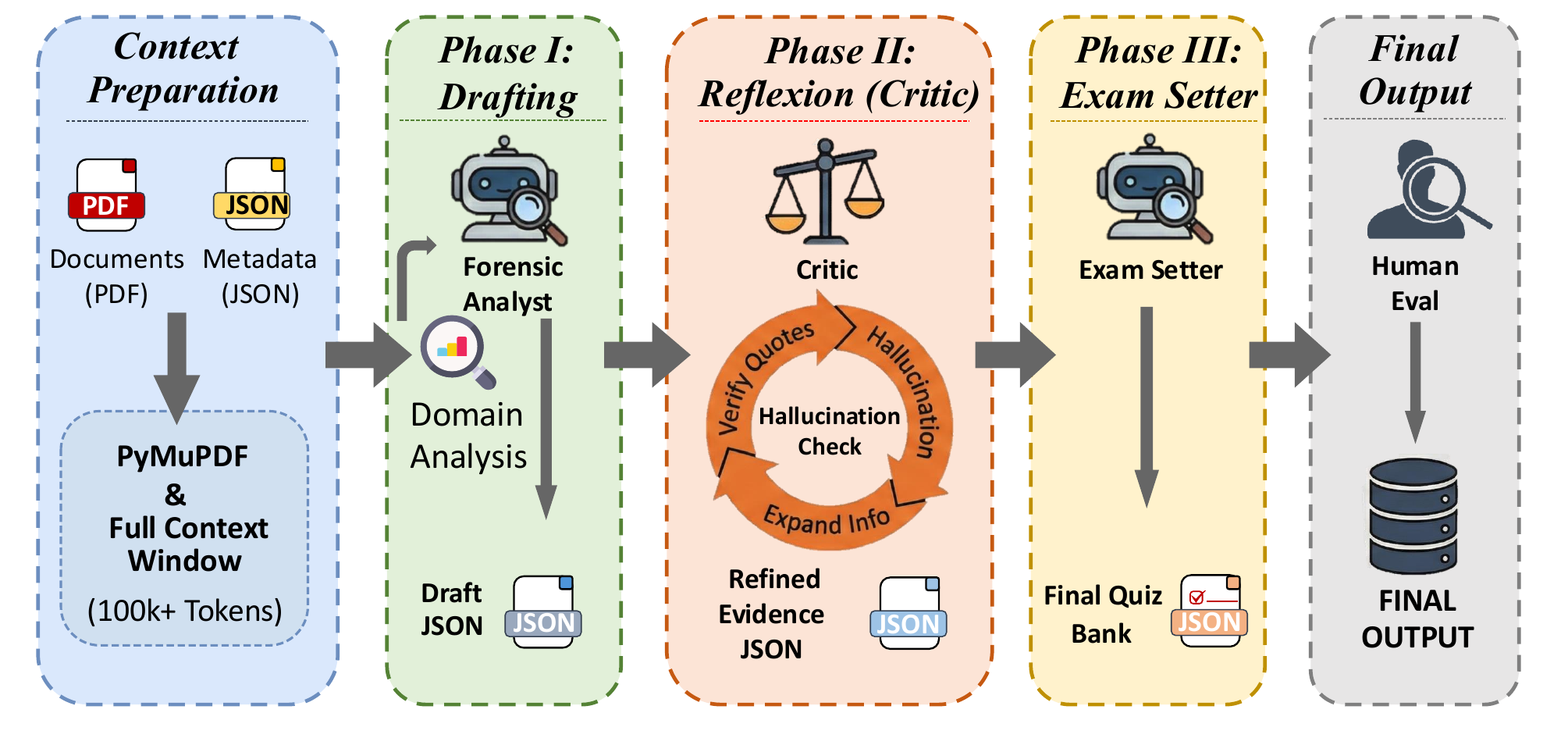}
    \caption{The QuizBank construction pipeline. }
    \label{fig:quizbank_pipeline}
    \vspace{-0.5cm}
\end{figure}

% \begin{itemize}
%     \item \textbf{Phase I: Domain-Adaptive Extraction (Drafting).}
%     The process uses the Forensic Analyst agent, scanning the full context window (100k+ tokens) of the raw PDF documents. Its objective is to identify high-value information segments to produce a draft JSON of key points.

%     \item \textbf{Phase II: Reflection \& Verification (The Critic).}
%     This phase implements a cyclic feedback loop involving hallucination checks, quote verification, and the expansion of missing information, producing a Refined Evidence JSON.

%     \item \textbf{Phase III: Dynamic Generation (Exam Setting).}
%     Finally, the Exam Setter agent converts the refined evidence into a structured assessment. To ensure an equitable and length-agnostic evaluation across all baseline models, the system dynamically samples a standardized probe of exactly 10 Multiple-Choice Questions (MCQs) per document: 5 Concept Questions focus on high-level understanding, while 5 Data Questions focus on specific details. This standardized length effectively prevents evaluation bias toward overly verbose generations and strictly tests core summarization abilities. Distractors are carefully generated, and all questions were manually verified by authors to ensure they cannot be answered using general world knowledge alone.
% \end{itemize}
Phase I: Drafting. The process uses the Forensic Analyst agent, scanning the full context window (100k+ tokens) of the raw PDF documents. Its objective is to identify high-value information segments to produce a draft JSON of key points.

Phase II: Critic. This phase runs a cyclic feedback loop of hallucination checks, quote verification, and gap expansion, producing a Refined Evidence JSON.

Phase III: Exam Setting. Finally, the Exam Setter agent converts the refined evidence into a structured assessment. To ensure an equitable and length-agnostic evaluation across all baseline models, the system dynamically samples a standardized probe of exactly 10 Multiple-Choice Questions (MCQs) per document: 5 Concept Questions focus on high-level understanding, while 5 Data Questions focus on specific details. This standardized length effectively prevents evaluation bias toward overly verbose generations and strictly tests core summarization abilities. 
In the end, all questions were manually verified by authors to ensure they cannot be answered using general world knowledge alone.

\paragraph{QuizBank Evaluation.} 
To quantify information preservation, we employ a VLM-based "open-book" exam protocol. Instead of directly feeding slide images to the VLM to generate answers, we adopt a two-stage pipeline to decouple visual content extraction from semantic comprehension. This design enables us to accurately isolate whether an incorrect answer stems from a visual extraction failure or a comprehension shortfall. We first parse all slide images into structured Markdown via a VLM. Then, the VLM is explicitly constrained with the prompt to answer QuizBank questions relying solely on this slide-derived context. Accuracy rates serve as a proxy for content quality. Our Error Analysis in Appendix~\ref{app:quizbank_details} shows that only 1.2\% of errors stem from the VLM extraction failures, ensuring the robustness of our evaluation against VLM bottlenecks. More details are in Appendix~\ref{app: quiz_eval}.

\subsection{Aesthetics Evaluation} \label{subsec: aesthetics}

To quantitatively evaluate the aesthetic quality of presentation slides against globally recognized professional standards, we introduce a four-dimensional evaluation framework grounded in cognitive load theory and visual ergonomics. Unlike traditional image quality assessment (IQA) metrics that treat images in isolation, by directly operating on the rendered visual domain, our framework remains architecture-agnostic and models the presentation as a temporal sequence, assessing both the \textit{spatial quality} (single slide) and the \textit{temporal coherence} (deck pacing). We focus on four core dimensions: Harmony, Engagement, Usability, and Visual Rhythm. The parameter decision process is shown in Appendix~\ref{app: parameter_tuning}. Detailed diagnostic patterns and examples are provided in Appendix~\ref{app:aesthetic_insights}.

\paragraph{Harmony Score.}
We compute a normalized per-slide harmony score $S_{slide}^{(i)}$ by optimizing its fit to hue templates in HSV space~\cite{cohen2006color}. To ensure consistency between our main formulation and the empirical deck-level calibration (detailed in Appendix~\ref{subsec:color_harmony_params}), the overall Harmony Score aggregates the mean harmony $\mu_{deck}$ and penalizes cross-slide inconsistency (standard deviation $\sigma_{deck}$):
\begin{equation}
    S_{harmony} = w_m\times \mu_{deck} - w_d\times\sigma_{deck}
\end{equation}

\paragraph{Engagement Score.}
We adapt Hasler and Süsstrunk’s colorfulness metric~\cite{hasler2003measuring} using opponent channels $rg = R-G$ and $yb = \frac{1}{2}(R+G) - B$:
\begin{equation}
    M_{slide} = \sqrt{\sigma_{rg}^2 + \sigma_{yb}^2} + 0.3 \sqrt{\mu_{rg}^2 + \mu_{yb}^2}
\end{equation}
We further define a deck-level pacing score from the standard deviation of slide colorfulness $\sigma_{pacing}$:
\begin{equation}
    \text{Score}_{pacing} = e^{ - \frac{(\sigma_{pacing} - \mu_{target})^2}{2w^2} }
\end{equation}

\paragraph{Usability Score.}
We evaluate figure-ground contrast within detected text regions using a layout analysis model~\cite{cui2025paddleocr}. Relative luminance $L$ is computed with sRGB linearization. With contrast ratio $c = (L_{max} + 0.05)/(L_{min} + 0.05)$, we map contrast into $S_{contrast} \in [0, 1]$ via logarithmic normalization (mapping the $21:1$ ratio to $1.0$):
\begin{equation}
    S_{contrast} = \frac{\ln(c)}{\ln(21)}
\end{equation}

\paragraph{Visual Rhythm Score.}
We introduce Visual Heart Rate Variability (VisualHRV), combining per-slide clutter (Subband Entropy) and temporal variability (RMSSD). 
Subband Entropy is computed as: $E_{SE} = \frac{E_L + w_c (E_a + E_b)}{1 + 2w_c}$.
% \begin{equation}\small
%     E_{SE} = \frac{E_L + w_c (E_a + E_b)}{1 + 2w_c}
% \end{equation}
For a raw entropy value $x$, calculate the normalized Score $S$ (0 to 1): $S(x) = \frac{1}{1 + e^{-k \cdot (x - \mu)}}$
For the valid sequence of entropy-derived scores ($N \geq 5$, as standard presentations filtered in Sec 2.1), we compute the length-normalized RMSSD:
\begin{equation}\small
    \text{RMSSD} = \sqrt{\frac{1}{N-1} \sum_{i=1}^{N-1} (S_{i+1}-S_i)^2}
\end{equation}
We then combine:
\begin{equation}
    \text{Score}_{VHRV} = \lambda_1 \cdot \overline{S_{entropy}} + \lambda_2 \cdot \text{RMSSD}
\end{equation}

\paragraph{Overall Aesthetics score.}
The overall Aesthetics score is computed as the sum of its weighted component scores, integrating spatial and temporal evaluations into a metric. 
\begin{equation}
    Score_{total}=H_{score}+E_{score}+U_{score}+V_{score}
\end{equation}

\subsection{Editability Evaluation} \label{subsec: edit}
Recent generative models produce visually convincing slides, yet their structural properties vary significantly depending on their design objectives. To quantify this, we introduce the Presentation Editability Intelligence (PEI) framework, a hierarchical scale from static visual mimicry to fully editable, narrative presentations. Levels 1–2 capture surface fidelity without true semantic structure, Levels 3–4 add consistent document organization and native, data-driven objects (e.g., editable charts), and Level 5 models temporal and multimedia dynamics so the deck functions as a directed experience. Systems are scored with a dependency-based “knockout” rule: failing a lower level precludes credit at higher ones. It is important to note that while PEI levels describe user-facing operational capabilities, our benchmark evaluates them fully automatically by directly parsing the underlying structural files (e.g., Office Open XML schemas via python-pptx and DOM trees for HTML), requiring zero human intervention. To make clustered systems easier to compare, we additionally report parser-derived subcriteria (text object editability, grouping, native chart/table editability, narrative structure, and animation/multimedia support; Appendix~\ref{app:pei_subcriteria}) while keeping the main PEI levels unchanged. Note that PEI is format-conditional: it measures the editability features exposed by the generated file format and parser, so it captures operational editability rather than overall visual or semantic quality. Detailed definitions are provided in Appendix~\ref{app: PEI}.

\subsection{Reliability Verification via Human}
\label{sec:reliability_verification}
To systematically verify the rationality and reliability of our evaluation scheme (Challenge 3), we integrate a human-preference validation stage into our benchmark. We construct the \textbf{Slides-Align1.5k} dataset, a ground-truth testbed annotated by a diverse, standardized pool including CS Ph.D.\ students, art undergraduates, and office clerks (Annotation details in Appendix~\ref{app:annotator_details}). 
We quantify the reliability of automated evaluation methods against this human ground truth using three core indicators: Average Spearman correlation, its Standard Deviation, and the Average Identical ratio.

\paragraph{Scope of the Metrics.}
The three branches provide different levels of evidence. QuizBank is a practical proxy for content retention (source-grounded questions, human audit of failures, oracle-context QA, backend-sensitivity checks; Sec.~\ref{sec:quizbank_analysis}, Appendix~\ref{app:quizbank_robust}); the aesthetic metrics are human-aligned computational proxies validated on Slides-Align1.5k within the tested style distribution, not universal measures of visual quality (Appendix~\ref{app:style_transfer}); PEI is closer to a direct structural measurement because it is parsed from the generated file structures. This separation lets readers assign the right scope to each score rather than treating all components as equally direct measures of quality.

\section{Experiments} \label{sec: exp}

\subsection{Experimental Setup} \label{subsec: setup}

\paragraph{Slides Generation Frameworks.} 
We evaluate several prominent slides-generation frameworks, encompassing both commercial platforms and recent open-source academic pipelines. To ensure the evaluation reflects the most up-to-date capabilities, all commercial systems were tested between November and December 2025. Crucially, all generated artifacts from these evaluations have been open-sourced to facilitate reproducible research. We present the results as a dated \method\ v1.0 snapshot; system access details and the rerun/update protocol are documented in Appendix~\ref{app:snapshot}.

The evaluated frameworks can be categorized into three paradigms. \textbf{(1) Source File Generation} directly outputs standard \texttt{.pptx} files. Commercial systems such as Gamma.com.ai, Quark-PPT~\cite{Quark_PPT} and Kimi-PPT~\cite{MoonshotAI_Kimi_Slides} adopt a template-filling approach. In contrast, recent open-source frameworks employ code-centric manipulation strategies. To ensure these frameworks could be evaluated fairly under our zero-shot, end-to-end benchmark setting, we implemented necessary input adaptations. For instance, AutoPresent* \citep{ge2025autopresent} was augmented with an LLM pre-processing step to convert our raw PDF inputs into its required bullet-point format, and its image backend was standardized. Similarly, PPTAgent* \citep{zheng2025pptagent} requires a reference template; we adapted it by providing a curated local template library and an LLM routing mechanism to dynamically select appropriate styles (detailed configurations in Appendix~\ref{app:opensource_eval}). We denote these adapted pipelines with an asterisk (*) to distinguish them from their raw original versions.

\textbf{(2) HTML Code Generation} leverages LLMs to synthesize structural code (HTML/CSS/JS). Platforms like Zhipu PPT~\cite{ZhipuAI_ChatGLM_Slides} and Skywork~\cite{SkyworkAI_Slides} utilize this method to achieve superior typographic precision and layout flexibility; however, they face limitations regarding fidelity loss when converting to standard \texttt{.pptx} formats. 

\textbf{(3) Image Generation Models} represent a shift toward visual synthesis, employing multi-modal models to generate slides directly as images. Exemplified by NotebookLM, Skywork-Banana Mode~\cite{SkyworkAI_Slides} and Kimi-Banana Mode~\cite{MoonshotAI_Kimi_Slides}, this method achieves studio-grade design aesthetics often unattainable by standard templates, but treating slides as pixel data compromises text editability and necessitates additional OCR layers for accessibility.

\subsection{Experimental Results} \label{subsec: result}

\paragraph{Content Results.}

The content quality of the slides is reflected by the accuracy of the QuizBank test. Table~\ref{tab: quizbank_full} presents the QuizBank accuracy results. Zhipu emerges as the top performer with an overall average of 88.29\%, demonstrating broad competency across topics. Furthermore, it proves to be the most robust model for complex content, achieving the highest accuracy on both High and Medium difficulty questions. In terms of domain-specific strengths, Skywork-Banana leads in `Brand' and `Report' generation, while Kimi-Standard achieves near-perfect performance (96.47\%) in `Personal' topics. The data highlights a trend where `Business' topics remain the primary bottleneck for current models, yielding the lowest average accuracy of 61.61\%.

To gauge the stability of these comparisons, we computed 95\% bootstrap confidence intervals (CIs) and the evaluated $n$ for each system (Appendix~\ref{app:quizbank_robust}). The intervals are tight (typically $\pm$2--5\%), and the top-ranked system's lower bound remains at or above the runner-up's upper bound, so the headline conclusions are not driven by small samples. The evaluation is also robust to the VLM backend: an all-question comparison of DeepSeek-v4-Pro and MiniMax-M2.7 yields nearly identical accuracies (60.8\% vs.\ 61.1\%), and an oracle-context QA test, where the evaluator answers the same MCQs from ground-truth source evidence instead of slide context, reaches 100\%. Observed failures thus primarily reflect generated-slide content loss or mismatch rather than evaluator error (a full-bank coverage and bias audit is in Appendix~\ref{app:quizbank_audit}).

\begin{table*}[t]
\small
\centering
\setlength{\tabcolsep}{3pt} % Reduced spacing to fit the extra column
\begin{NiceTabular}{l|cccccc|c|ccc}
\toprule
\multirow{3}{*}{\bf Product} & \multicolumn{6}{c|}{\bf Purpose Categories} & \multicolumn{1}{c}{\bf Topic} & \multicolumn{3}{c}{\bf Difficulty Level} \\
& Brand & Business & Personal & Product & Course & Work & Topic & & & \\
& Promote & Plan & Statement & Launch & Preparation & Report & Intro. & High & Low & Med \\
\midrule
\rowcolor{white}
Gamma & 63.57 & 52.50 & 74.00 & 54.44 & 85.00 & 48.46 & 76.34 & 67.26 & 75.56 & 68.45 \\
\rowcolor{white}
Kimi-Banana & 84.29 & \textbf{77.50} & 96.00 & \textbf{90.43} & 84.00 & 85.38 & 86.22 & 83.33 & 91.58 & 86.67 \\
\rowcolor{white}
Kimi-Smart & N/A & N/A & 95.00 & N/A & N/A & 84.29 & 79.68 & 78.78 & 88.50 & 78.57 \\
\rowcolor{white}
Kimi-Standard & N/A & N/A & \textbf{96.47} & N/A & N/A & 80.00 & 76.88 & 75.71 & 88.05 & 75.43 \\
\rowcolor{white}
NotebookLM & 83.00 & 45.00 & N/A & 84.21 & \textbf{93.00} & 86.15 & 68.33 & 69.81 & 86.00 & 71.16 \\
\rowcolor{white}
Quark & 84.29 & 51.25 & 94.00 & 79.05 & N/A & 68.00 & 82.13 & 69.78 & 90.52 & 82.12 \\
\rowcolor{white}
Skywork & 80.00 & 68.75 & 78.50 & 76.52 & 84.67 & 86.15 & 83.12 & 80.00 & 82.58 & 81.31 \\
\rowcolor{white}
Skywork-Banana & \textbf{92.14} & 67.50 & 91.50 & 89.57 & 87.69 & \textbf{90.00} & 79.33 & 79.67 & 88.00 & 84.31 \\
\rowcolor{white}
Zhipu & 80.00 & 68.75 & 96.00 & 86.08 & 90.00 & 84.17 & \textbf{90.44} & \textbf{84.07} & \textbf{92.00} & \textbf{88.97} \\
\hdashline
\rowcolor{white}
\shortstack[l]{PPTAgent*} & 62.25 & 47.37 & 69.15 & 61.80 & 66.86 & 61.27 & 61.94 & 58.94 & 66.76 & 61.68 \\
\rowcolor{white}
AutoPresent* & 56.13 & 42.72 & 62.35 & 55.73 & 60.29 & 55.25 & 55.85 & 53.15 & 60.20 & 55.62 \\
\midrule
\rowcolor{gray!15}
\it Average & \it 76.19 & \it 57.93 & \it 85.30 & \it 75.31 & \it 81.44 & \it 75.37 & \it 76.39 & \it 72.77 & \it 82.70 & \it 75.84 \\
\bottomrule
\end{NiceTabular}
\caption{The QuizBank accuracy(\%) for different products by topic and difficulty level. The \textit{Purpose Categories} group contains six functional-task scenarios, while the \textit{Topic Intro.} column reports topic-introduction instructions; \texttt{N/A} denotes task types not supported by the corresponding system (a true constraint rather than a missing evaluation).}
\label{tab: quizbank_full}
\end{table*}

\begin{table}[htb]
    \footnotesize
    \centering
    \begin{NiceTabular}{l|ccc}
         \toprule
         \multirow{2}{*}{\bf Method} & \multicolumn{2}{c}{\bf Spearman} & {\bf Identical} \\
          & {\bf Avg($\uparrow$)} & {\bf Std} ($\downarrow$)  &  {\bf Avg($\uparrow$)} \\
         \midrule
         \rowcolor{gray!7} \method & \textbf{0.71} & \textbf{0.16} & \textbf{32.6} \\
         \rowcolor{white!7} LLM-as-Judge Rating & 0.57 & 0.23 & 20.7 \\
         \rowcolor{white!7} LLM-as-Judge Arena & 0.52 & 0.27 & 17.3 \\
         \rowcolor{white!7} PPTAgent & 0.53 & 0.26 & 17.8 \\
         \midrule
         \rowcolor{white!7} Humans & 0.85 & 0.12 & 45.3 \\
         \bottomrule
    \end{NiceTabular}
    \caption{Human alignment results on SlidesGen-Bench.}
    \label{tab:SlidesGen_results}
\end{table}

\paragraph{Aesthetics Results.}

As shown in Table \ref{tab:aesthetics_eval}, Skywork-Banana outperforms all baselines in the aggregate evaluation. Notably, Skywork-Banana achieves the highest scores in the overall Aesthetics Score (27.28 vs. 26.58 for the runner-up) and Engagement (8.30 vs. 6.41), suggesting that Skywork-Banana is particularly effective at generating high-fidelity, visually engaging slides. Models with lower aesthetic scores, such as Quark and AutoPresent, have a relatively flat narrative pace in the generated slides, resulting in significantly lower Rhythm scores. The analysis of the aesthetics results is in Appendix~\ref{app: visual}.

To examine style transferability, we ran a style-transfer validation over the 1,435 Slides-Align1.5k aesthetics records, covering six style buckets---standard corporate/mixed (66.8\%), minimalist-light (14.8\%), image-rich marketing (8.6\%), diagrammatic/scientific (6.3\%), data-heavy report (2.6\%), and dark-mode minimalist (0.8\%)---with full diagnostics in Appendix~\ref{app:style_transfer}. We find no systematic dark-mode penalty (dark decks receive slightly better human mean ranks than light decks, 3.84 vs.\ 4.27), and removing any single style bucket changes the composite human-alignment score by less than 0.05. Boundary cases (e.g., image-rich marketing decks and creative palettes) show larger rank errors, so we present these metrics as human-aligned proxies within the tested style distribution rather than universal measures of visual quality.

\begin{table*}[htb]
    \small
    \centering
    % Adjusted colsep to fit columns gracefully
    \setlength{\tabcolsep}{8.0pt} 
    % Updated column definition: Removed Avg. Rank column
    \begin{NiceTabular}{l|cccc:c}
         \toprule
         {\bf Product} & {\bf Usability} ($\uparrow$) & {\bf Engagement} ($\uparrow$) & {\bf Harmony} ($\uparrow$) & {\bf Rhythm} ($\uparrow$) & {\bf Aesthetics} ($\uparrow$) \\
         \midrule
         % Highlighting the top performer (Skywork-Banana)
         \rowcolor{gray!7} Skywork-Banana & 5.62 & \textbf{8.30} & -0.47 & 13.84 & \textbf{27.28} \\
         \rowcolor{white} Kimi-Banana & \textbf{5.72} & 6.41 & -0.55 & \textbf{15.01} & 26.58 \\
         \rowcolor{white} NotebookLM & 4.13 & 7.32 & \textbf{-0.35} & 11.72 & 22.82 \\
         \rowcolor{white} Zhipu & 4.87 & 7.52 & -1.60 & 11.27 & 22.06 \\
         \rowcolor{white} Skywork & 4.83 & 7.60 & -1.18 & 9.44 & 20.69 \\
         \rowcolor{white} Kimi-Standard & 4.61 & 6.28 & -1.75 & 10.12 & 19.25 \\
         \rowcolor{white} Kimi-Smart & 4.13 & 7.99 & -1.88 & 8.06 & 18.30 \\
         \rowcolor{white} Gamma & 5.31 & 6.31 & -1.51 & 6.99 & 17.09 \\
         \rowcolor{white} Quark & 5.03 & 7.41 & -1.91 & 6.33 & 16.86 \\
         \hdashline
         \rowcolor{white} \shortstack[l]{PPTAgent*} & 5.01 & 6.12 & -1.43 & 5.62 & 15.32 \\
         \rowcolor{white} AutoPresent* & 4.28 & 5.93 & -2.15 & 4.57 & 12.63 \\
         \bottomrule
    \end{NiceTabular}
    \caption{Slides Aesthetics evaluation results. We report detailed component scores ($\uparrow$) including Usability, Engagement, Harmony, and Rhythm. The table is sorted by the Aesthetics score (sum of these components). }
    \label{tab:aesthetics_eval}
\end{table*}

\paragraph{Human Alignment Results.} 
Table~\ref{tab:SlidesGen_results} presents the evaluation on the Slides-Align1.5k dataset. This protocol achieved a robust inter-annotator agreement (Average Spearman $\rho = 0.85$, Std $= 0.12$), serving as a reliable gold standard. For baselines, we compare against PPT-Eval from PPTAgent \citep{zheng2025pptagent}, as well as standard LLM-as-a-Judge paradigms (Rating and Arena) \citep{zheng2023judging}. To ensure a strong upper bound, the LLM judges are powered by a state-of-the-art multimodal model (gemini-3-flash-preview), taking the rendered slide images as direct inputs with prompts mirroring our aesthetic criteria (Appendix~\ref{app: Judge}).

As shown in the table, SlidesGen-Bench demonstrates superior reliability across all metrics. It achieves the highest Average Spearman correlation of 0.71, significantly surpassing the runner-up LLM-as-Judge Rating (0.57) and PPTAgent (0.53). Furthermore, our method exhibits the most stable performance with the lowest Standard Deviation (0.16). Notably, in terms of the Average Identical ratio, our approach reaches 32.6\%, delivering a substantial improvement over existing LLM-based evaluators (which struggle around 20\%). These results verify that our computational methodology aligns more closely and robustly with human perceptual standards than existing protocols.

\paragraph{Editability Results.}
\begin{table}[t]
\centering
\footnotesize

\resizebox{\columnwidth}{!}{\begin{tabular}{l ccccc >{\columncolor{colHighlight}}c } % 给 PEI Level 列加背景色
\toprule
% 表头行
\textbf{System} & 
\textbf{T1} & \textbf{T2} & \textbf{T3} & \textbf{T4} & \textbf{T5} & 
\textbf{PEI} \\

% 副表头 (解释 T1-T5)
% \rowcolor{white} % 覆盖背景色
% \scriptsize\textit{(Core Metric)} & 
% \scriptsize\textit{Text Sep.} & 
% \scriptsize\textit{Vector} & 
% \scriptsize\textit{Structural} & 
% \scriptsize\textit{Parametic} & 
% \scriptsize\textit{Cinematic} & 
% \scriptsize\textit{(Ours)} & 
% \scriptsize\textit{Score} \\

\midrule
\textbf{PPTAgent*}           & \cmark & \cmark & \cmark & \cmark & \cmark & \textbf{L5} \\

\textbf{AutoPresent*}           & \cmark & \cmark & \cmark & \cmark & \xmark & \textbf{L4} \\
\arrayrulecolor{black!10}\midrule % 极细的浅灰分割线
\arrayrulecolor{black} % 恢复黑色

% --- L3 Group ---
\textbf{Quark}           & \cmark & \cmark & \cmark & \xmark & \xmark & \textbf{L3} \\

\arrayrulecolor{black!10}\midrule % 极细的浅灰分割线
\arrayrulecolor{black} % 恢复黑色

\textbf{Gamma}           & \cmark & \cmark & \xmark & \xmark & \xmark & \textbf{L2}  \\
\textbf{Skywork}         & \cmark & \cmark & \xmark & \xmark & \xmark & \textbf{L2}  \\
\textbf{Kimi (Standard)} & \cmark & \cmark & \xmark & \xmark & \xmark & \textbf{L2} \\
\textbf{Kimi (Smart)}    & \cmark & \cmark & \xmark & \xmark & \xmark & \textbf{L2}  \\
\textbf{Zhipu}   & \cmark & \cmark & \xmark & \xmark & \xmark & \textbf{L2} \\

\arrayrulecolor{black!10}\midrule % 极细的浅灰分割线
\arrayrulecolor{black} % 恢复黑色

% --- L1 Group ---
\textbf{Kimi (Banana)}           & \cmark & \xmark & \xmark & \xmark & \xmark & \textbf{L1}  \\
\textbf{Skywork (Banana)}& \cmark & \xmark & \xmark & \xmark & \xmark & \textbf{L1} \\

\arrayrulecolor{black!10}\midrule
\arrayrulecolor{black}

% --- L0 Group (重点对比对象) ---
\textbf{NotebookLM}      & \xmark & \xmark & \xmark & \xmark & \xmark & \textbf{L0} \\

\bottomrule
\end{tabular}}

\caption{\textbf{Evaluation Results against the PEI Framework.} 
Columns T1--T5 represent the hierarchical pass/fail criteria (\cmark/\xmark). }
\label{tab:pei_results}
\end{table}

Table \ref{tab:pei_results} reveals a distinct "Structural Barrier" in the landscape of presentation generation. While the majority of evaluating systems (e.g., Gamma, Skywork, Kimi-Smart) have mastered visual fidelity, achieving Level 2 (Vector Visual), they suffer from Structural Amnesia. These systems generate slides as isolated artistic canvases without global logic (e.g., $<p:sldMaster>$), meaning layout changes cannot be propagated system-wide. Quark stands out as the sole system to breach the "Toy-to-Tool" threshold (L3), demonstrating the capability to generate cohesive, grouped, and master-based hierarchies that support professional editing workflows. Although the Parametric Gap at Level 4 remains universally unsolved by commercial models, Open-source academic frameworks uniquely bridge these gaps: AutoPresent achieves L4 by natively generating programmable chart objects ($<c:chart>$), while PPTAgent reaches L5 by utilizing raw templates as structural priors to inherit temporal dynamics (animations and slide transitions). Notably, PPTAgent's L5 capability comes with a caveat: these inherited dynamic behaviors act as rigid priors and cannot be parametrically modified.

We also note that the current PEI levels cluster most commercial systems at L1--L2, reflecting limited discriminative granularity, with the substantive implication that many visually strong systems still expose limited structural editability under parseable file formats. To support finer-grained comparison of clustered systems, we report parser-derived subcriteria in Appendix~\ref{app:pei_subcriteria}.

\begin{figure}[t]
    \centering
    \includegraphics[width=0.88\linewidth]{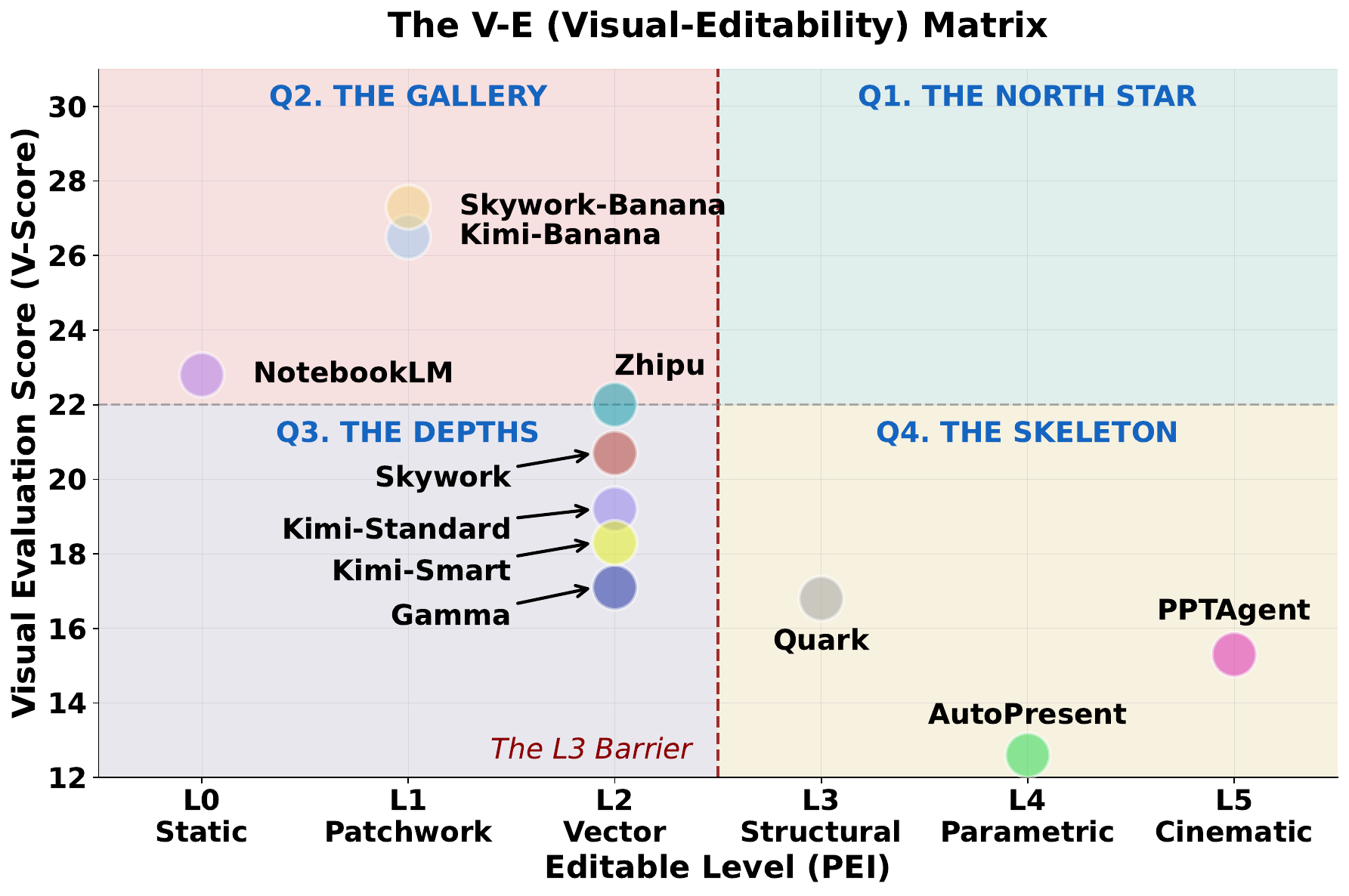}
    \caption{\textbf{The V-E (Visual-Editability) Matrix.}}
    \label{fig:ve_matrix}
\end{figure}

\paragraph{The V-E metrics Results.}
To quantify the interplay between aesthetics and functional utility, we map the evaluated systems onto the Visual-Editability (V-E) Matrix (Figure \ref{fig:ve_matrix}). Commercial systems dominate the left hemisphere (L0-L2), prioritizing visual polish over structural depth, while the lower-right quadrant (Q4) houses Quark (L3) and academic frameworks like AutoPresent (L4) and PPTAgent (L5), which remain confined there due to aesthetic limitations or reliance on unmodifiable rigid templates.
Crucially, the upper-right quadrant (Q1) remains vacant: no system achieves peak performance in both dimensions simultaneously. This gap highlights the trade-off between visual expressiveness and structural depth in current methodologies, suggesting future opportunities: bridging static visual proxies and functionally rigid schemas to enable fully controllable, dynamic visual storytelling.

\subsection{Ablation Studies} \label{subsec: ablation}

\subsubsection{Analysis of the Aesthetics Metrics}
To assess the individual contributions of our proposed aesthetic metrics, we conducted a comprehensive ablation study, as presented in Table~\ref{tab:ablation_metrics}. The parameters for our computational metrics were optimized on a completely independent verification set of 1000 slide pairs from human-generated slides (Detailed results are shown in Appendix~\ref{app: parameter_tuning}). 

\begin{table}[htp]
    \footnotesize
    \centering
    % Note: Requires \usepackage{nicematrix} and \usepackage{booktabs}
    % \setlength{\tabcolsep}{8.0pt}
    \begin{NiceTabular}{l|ccc}
         \toprule
         \multirow{2}{*}{\bf Metric Configuration} & \multicolumn{2}{c}{\bf Spearman} & {\bf Identical} \\
          & {\bf Avg($\uparrow$)} & {\bf Std} ($\downarrow$)  &  {\bf Avg($\uparrow$)} \\
         \midrule
         % Single Metrics
         Only Engagement & 0.224 & 0.349 & 14.8 \\
         Only Harmony & 0.312 & 0.414 & 15.6 \\
         Only Usability & 0.574 & 0.207 & 21.5 \\
         \rowcolor{white!7} Only Visual HRV & 0.618 & 0.198 & 24.4 \\
         \midrule
         % Combinations
         Harmony + Engagement & 0.371 & 0.440 & 20.7 \\
         Usability + Engagement & 0.612 & 0.212 & 23.7 \\
         Usability+ Har. + Eng. & 0.667 & 0.206 & 24.4 \\
         \midrule
         % Full Method
         \rowcolor{gray!10} \textbf{Full Method (All)} & \textbf{0.710} & \textbf{0.160} & \textbf{32.6} \\
         \bottomrule
    \end{NiceTabular}
    \caption{Ablation study of aesthetic metrics.}
    \label{tab:ablation_metrics}
\end{table}

Specifically, we observe that Usability serves as a strong baseline among the traditional heuristic metrics (Avg. Spearman 0.574), significantly outperforming Engagement (0.224) and Harmony (0.312) when used in isolation. This suggests that readability and visual clarity are the primary drivers of user preference in slide design. 

Furthermore, the synergy between metrics is evident. Combining Harmony, Engagement, and Usability yields a correlation of 0.667. Yet, the Full Method, which includes all metrics, achieves the highest performance across all dimensions: an Average Spearman correlation of 0.710, the lowest standard deviation of 0.160, and an Average Identical ratio of 32.6\%. This confirms that our multi-dimensional approach captures the complex interplay of perceptual ease and visual interest better than any single modality.

\subsubsection{Analysis of Errors in QuizBank}
\label{sec:quizbank_analysis}

We conduct a failure analysis on the $2499$ incorrect instances ($19.1\%$ of $13,023$ total samples) to understand generation limitations. All incorrect instances were categorized by human annotators, who inspected the rendered slide images against the ground truth and assigned each case to the error taxonomy. A full taxonomy of error types and model breakdowns is provided in Appendix~\ref{app:quizbank_details}. As shown in Table~\ref{tab:error_distribution}, \textit{Missing Content} is the primary bottleneck, accounting for $66.3\%$ of errors. This indicates that current models prioritize high-level visual narratives over the retrieval of granular facts (e.g., specific dates or metrics) required by the ground truth. \textit{Content Value Mismatch} ($31.7\%$) constitute the secondary error sources. Comparing model performance (Table~\ref{tab:model_breakdown}, Appendix) reveals a distinct trade-off: top-performing systems (e.g., Zhipu, Kimi-Banana) significantly reduce content omission but exhibit higher rates of value mismatch ($49\text{--}52\%$). This suggests that as models improve at retrieving detailed content, the challenge shifts from data recall to maintaining factual consistency.

\section{Related work} \label{sec: rel}

\paragraph{Slides Evaluation.}
Existing methodologies predominantly fall into reference-based metrics and LLM-based assessments. Early approaches, such as SlideCoder~\cite{tang2025slidecoder} and parts of AutoPresent~\cite{ge2025autopresent}, frame evaluation as a reconstruction problem, calculating fidelity against ground-truth references. While quantifiable, this reliance on paired data precludes their use in open-ended generation where canonical references are absent. To mitigate this, recent frameworks like PPTAgent~\cite{zheng2025pptagent} employ LLMs to judge content and aesthetics. However, these LLM-based evaluators introduce stochasticity, exhibiting susceptibility to bias and reasoning failures~\cite{thakur2025judging, krumdick2025no}. Critically, the rare paradigm incorporates a human alignment stage to validate metric reliability. This absence of correlation with human preference highlights the need for a consistent, reference-free evaluation standard that is both quantitatively robust and perceptually aligned.

% \paragraph{Multimodal Benchmarking.}
% The paradigm of utilizing LLMs to x generative outputs has gained significant traction \citep{zheng2023judging}. However, recent literature increasingly highlights the vulnerabilities of LLM-as-a-judge frameworks, particularly their susceptibility to verbosity bias, positional bias, and stochastic reasoning failures \citep{saito2023verbosity, wang2024executable, thakur2025judging}. In the multimodal domain, these issues are compounded by the semantic gap between visual rendering and textual code representations. By decoupling evaluation into a structured VLM-extraction pipeline and deterministic computational aesthetic formulas, SlidesGen-Bench addresses the call for more reproducible, metric-driven evaluation frameworks that mitigate the inherent biases of black-box LLM judges.

\paragraph{Computational Aesthetics.} Computational aesthetics has evolved from heuristic features to comprehensive frameworks like AIM \citep{oulasvirta2018aalto} and deep learning models such as NIMA \citep{talebi2018nima}. However, presentation slides constitute a unique multimodal domain requiring a distinct balance between high-contrast symbology and aesthetic cohesion, often rendering generic photography metrics insufficient. To address this, we distill evaluation into four dimensions that prioritize perceptual fluency. For visual appeal, we refine standard Color Harmony \citep{cohen2006color} via saturation weighting and adapt Hasler’s metric \citep{hasler2003measuring} into an Engagement Score to quantify vibrancy. Functionally, we adopt a Usability Score based on WCAG luminance \citep{caldwell2008web} to ensure deterministic legibility, and utilize Subband Entropy \citep{rosenholtz2007measuring} to model Visual HRV, providing a robust proxy for audience cognitive load.

\section{Conclusion} \label{sec: con}

In this work, we introduce \method, a standardized benchmark designed to systematically evaluate heterogeneous slide generation systems. By projecting diverse architectural outputs into a unified visual and structural space, we enable comparable, quantitative assessments across Content, Aesthetics, and Editability. Furthermore, validation on our novel Slides-Align1.5k dataset demonstrates that our structured computational metrics achieve significantly higher alignment with human preferences than existing evaluation pipelines. Ultimately, \method provides a reliable and reproducible framework to accelerate future research in automated presentation synthesis.

\section{Limitations} \label{sec: limit}

Despite the contributions of \method, there are limitations to our current framework. First, our evaluation focuses exclusively on static visual content, overlooking temporal dynamics such as animations and slide transitions; this limitation is prevalent across existing evaluation protocols, highlighting a broader community challenge for dynamic presentation flows. Second, content evaluation depends on the chosen VLM extraction and question-answering configuration: QuizBank is a structured content-retention evaluation under a specified evaluator setup rather than evaluator-independent ground truth, and although our human audit, oracle-context QA, and backend-sensitivity checks bound this risk (Sec.~\ref{sec:quizbank_analysis}, Appendix~\ref{app:quizbank_robust}), results may shift with different evaluator stacks. Third, PEI is format-conditional---it measures editability features exposed by the generated file format, so image- or web-output systems face a natural architectural ceiling unrelated to visual quality. Fourth, the aesthetic metrics carry cultural and domain assumptions: derived from visual-ergonomics principles and validated on our slide-preference distribution, they may need recalibration for styles outside it (Appendix~\ref{app:style_transfer}). Finally, Slides-Align1.5k remains primarily English-centric and optimized for globalized professional communication (e.g., corporate and academic standards); future iterations will explore how varying cultural tolerances for visual density affect human preference (e.g., medical or legal reports).

\bibliography{references}

\clearpage\appendix\section*{Appendix}

\section{Dataset Distribution Details}
\label{app:distribution}

\begin{figure}[h]
    \centering
    \includegraphics[width=\linewidth]{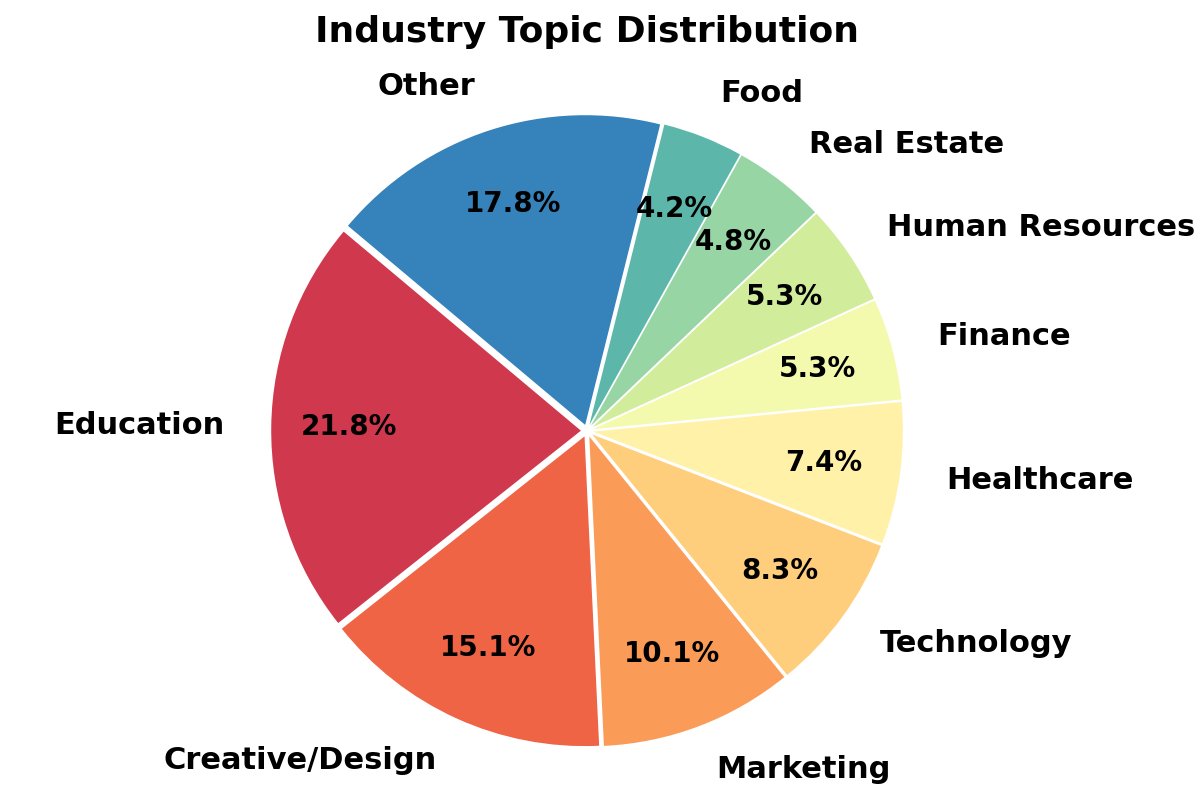}
    \caption{The topic distribution of 30k human-generated slides, highlighting prevalent categories such as Education and Technology.}
    \label{fig:industry}
\end{figure}

Figure~\ref{fig:industry} illustrates the topic distribution of the 30k collected human-authored slides used to guide our instruction generation process.

\begin{figure}[ht]
    \centering
    \includegraphics[width=\linewidth]{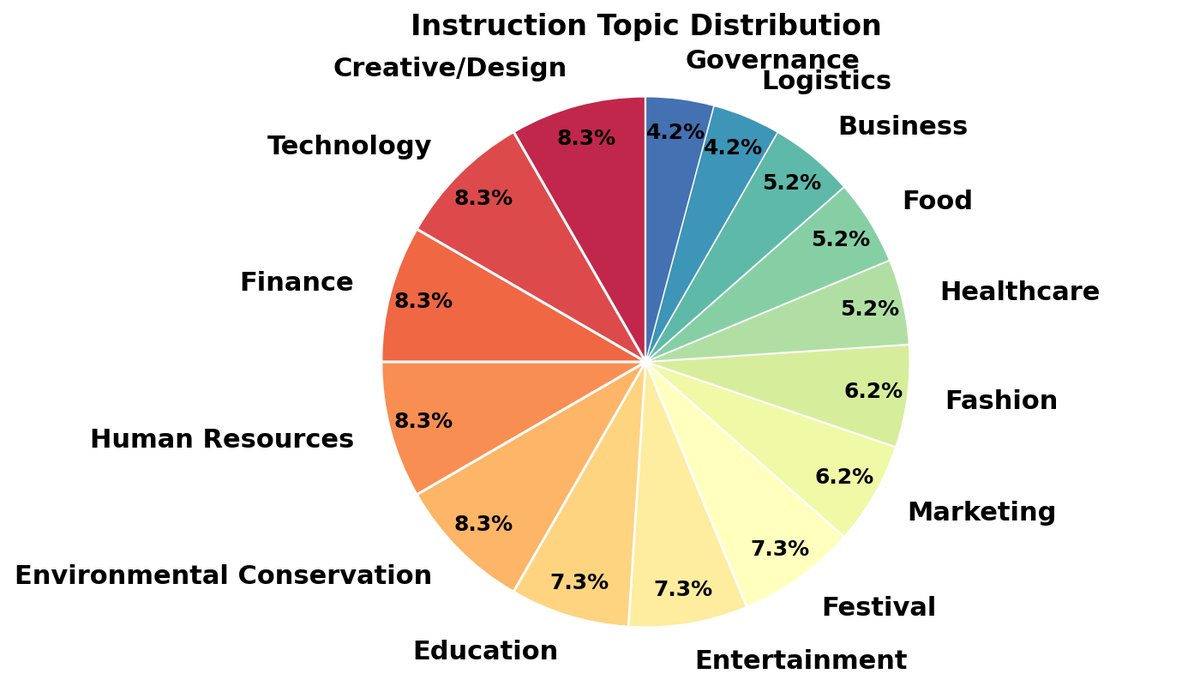}
    \caption{The distribution of instructions with different topics.
    }
    \label{fig: topic}
    \vspace{-0.5cm}
\end{figure}

Figure~\ref{fig: topic} depicts the topic distribution of the curated instructions.

Table \ref{tab: purpose-diverse} shows the domain distribution of the instructions.

\begin{table}[htp]
\small
\centering
\begin{tabular}{lr}
\toprule
\textbf{Domain} & \textbf{Count} \\
\midrule
Brand Promote & 15 \\
Business Plan & 8 \\
Knowledge Teaching & 15 \\
Personal Statement & 20 \\
Product Launch & 24 \\
Work Report & 13 \\
Topic Introduction & 94 \\
\midrule
\textbf{Total} & \textbf{189} \\
\bottomrule
\end{tabular}
\caption{Domain Distribution}\label{tab: purpose-diverse}
\end{table}

\section{Dataset Richness Category Details }\label{app: richness}
The richness score $S$ is defined as follows, where $T$ and $I$ denote the text length and image count:
\begin{equation} \small
    S = w_t \cdot \frac{T - \min(T)}{\max(T) - \min(T)} + w_i \cdot \frac{I - \min(I)}{\max(I) - \min(I)}
\end{equation}
where we set $w_t = 0.7$ and $w_i = 0.3$ to prioritize textual semantic density while accounting for visual requirements. 

\begin{figure}[t]
    \centering
    \includegraphics[width=\linewidth]{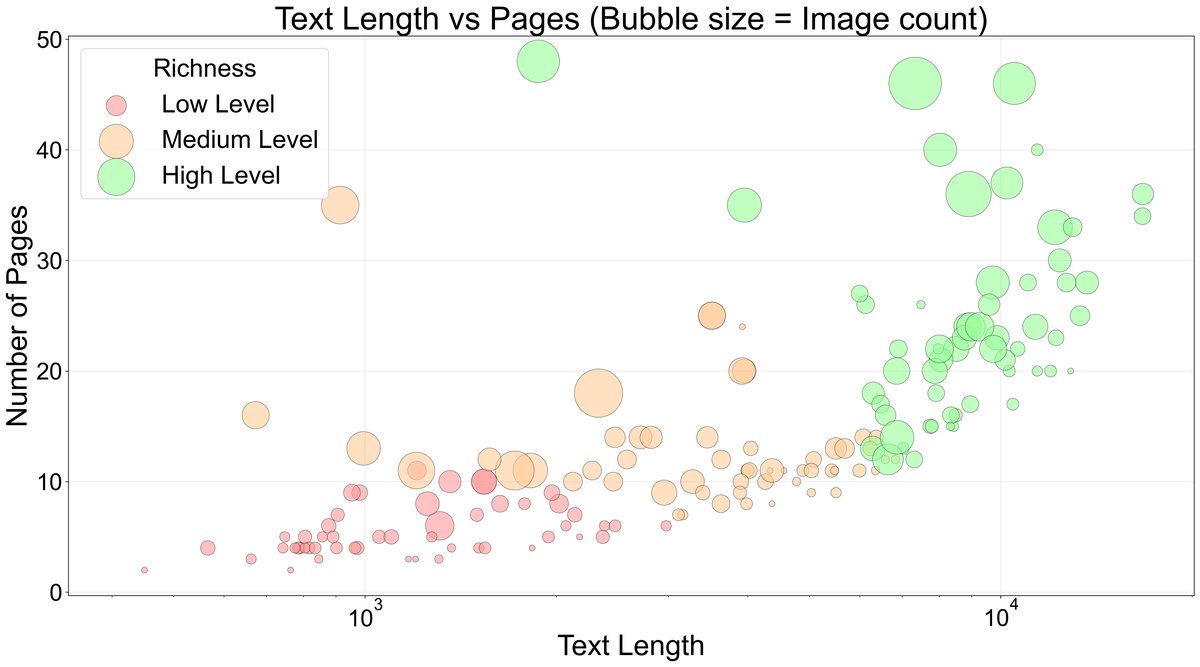}
    \caption{Analysis of Content Richness.}
    \label{fig:richness_dashboard}
    \vspace{-0.5cm}
\end{figure}

 Based on the calculated scores, we stratify the dataset into three balanced levels—Low, Medium, and High—each containing exactly 63 instructions (33.3\% of the total). As shown in the bubble plots in Figure~\ref{fig:richness_dashboard}, these levels correlate strongly with structural complexity: "Low" richness instructions typically require synthesizing fewer than 2,500 characters into brief decks, whereas "High" richness instructions demand the organization of over 7,500 characters and 20+ images into extensive presentations, effectively testing the model's stability and long-context summarization ability.

\section{Justification for the Benchmark Size ($N=189$)}
\label{sec:dataset_size_justification}

The dataset size of $N=189$ was a deliberate choice, necessitated by the high operational cost and technical complexity of evaluating professional slide generation. We offer the following clarifications regarding this design:

\paragraph{1. Practicality and Time Cost.} High-quality presentation generation is significantly more time-intensive than standard text generation. For instance, tools like Zhipu PPT require approximately 20 minutes to produce a multi-slide deck in our settings. Evaluating nine mainstream systems across 189 instructions already demands nearly 1,300 generation hours. Thus, $N=189$ represents a rigorous balance between evaluation depth and practical feasibility.

\paragraph{2. Representativeness and Scenario Richness.} We prioritized the quality, complexity, and diversity of each sample over sheer volume. Unlike benchmarks utilizing short prompts, SlidesGen-Bench employs comprehensive source documents (averaging 4,525 characters) to generate full-length decks (mean: 13.3 pages). The dataset covers seven distinct domains: 49.7\% focuses on general knowledge synthesis (\textit{Topic Introduction}), while the remaining 50.3\% comprises specialized \textit{Functional Tasks} (e.g., \textit{Business Plans, Personal Statements}) to evaluate domain-specific adaptability and logical rigidity.

\paragraph{3. Depth via Difficulty Stratification.} To ensure the benchmark remains sufficiently challenging, we implemented a Content Richness stratification (Low, Medium, and High). The ``High'' richness category specifically stress-tests a model's formatting stability and long-context summarization abilities by demanding the organization of over 7,500 characters and 20+ images into cohesive, extensive presentations.

\section{QuizBank Evaluation Details}
\label{app: quiz_eval}
\paragraph{A.1 Multimodal Content Extraction}
Rather than directly feeding slide images into the evaluator to generate answers, we adopt a two-stage pipeline that begins with explicit content extraction. The primary motivation for this design is to decouple visual content extraction from semantic comprehension. By isolating these two processes, we can accurately diagnose the root causes of evaluation failures, explicitly distinguishing whether an incorrect answer stems from a visual extraction error or a reasoning/understanding shortfall. To transform slides into machine-readable context, we employ a state-of-the-art multimodal model (gemini-3-flash-preview) to extract the contents from slide images (The prompt used is in Appendix~\ref{app: quiz_eval_prompt}). The extraction targets three specific modalities:
\begin{itemize}
\item \textbf{Textual Claims:} Verbatim headlines, core bullet points, and callout text.
\item \textbf{Quantitative Data:} Specific numerical values extracted from charts, tables, and in-text metrics.
\item \textbf{Visual Interpretation:} Semantic descriptions of non-textual elements (e.g., "green up-arrow indicating growth") to capture visual information flow.
\end{itemize}
Benefiting from this decoupled design, our Error Analysis in Appendix~\ref{app:quizbank_details} show that only 1.2\% of errors stem from the VLM extraction failures, ensuring the robustness of our evaluation against VLM bottlenecks.
\paragraph{A.2 The "Open-Book" Exam}
The evaluator VLM (gemini-3-flash-preview) is provided with the structured Markdown derived from the slides and the ground-truth QuizBank questions. Then the VLM is explicitly constrained with the prompt to \textit{"Answer strictly based on the provided slide context and cite the source"}, then attempts to answer QuizBank questions relying solely on this slide-derived context (The prompt used is in Appendix~\ref{app: quiz_eval_prompt}).
\paragraph{A.3 Scoring and Analysis}
The evaluation metric relies on the binary success of the VLM examinee:
\begin{itemize}
\item \textbf{Correct Answer:} Confirms the specific concept or data point was successfully transferred from the source document to the slides.
\item \textbf{Incorrect/Insufficient Information:} Indicates information loss. This distinction allows us to measure quality granularly, differentiating between the preservation of high-level concepts and specific data points.
\end{itemize}

\section{Mathematical Formulation of Color Harmony}
\label{app:color_harmony}
For a given slide image, we utilize the HSV color space and a set of eight geometric templates $\mathcal{T} = \{i, I, V, L, Y, X\}$ on the hue wheel~\cite{cohen2006color}. An optimization problem is solved to find the optimal template $T$ and rotation angle $\alpha$ that minimizes the weighted angular distance of the slide's pixels to the template sectors. The harmonic deviation $D$ is defined as:
\begin{equation}
    D(H, S) = \frac{\sum_{p} S_p \cdot \text{dist}(H_p, T_{\alpha})}{\sum_{p} S_p}
\end{equation}
where $H_p$ and $S_p$ represent the hue and saturation of pixel $p$, and $\text{dist}(\cdot)$ calculates the shortest angular distance (in degrees) to the valid sector of the rotated template $T_{\alpha}$~\cite{o2011color}. We map this raw deviation $D$ to the normalized quality probability score $S_{slide}^{(i)} \in [0, 1]$ used in the main paper via a Gaussian decay function.

\section{Aesthetic Metrics Parameter Tuning}
\label{app: parameter_tuning}

This appendix details the parameter decision process for four key aesthetic metrics: \textbf{SubbandEntropyMetric}, \textbf{ColorfulnessMetric}, \textbf{ColorHarmonyMetric}, and \textbf{VisualHRVMetric}. Our calibration follows a three-stage methodology: (1) empirical distribution analysis on human-generated slides, (2) initial parameter selection based on distribution characteristics, and (3) optimization against human preference rankings.

\subsection{Methodology Overview}

\subsubsection{Stage 1: Empirical Distribution Analysis}
We collected and analyzed over 3,000 human-generated PowerPoint slides from diverse professional contexts. For each metric, we computed raw scores across the corpus to establish baseline distributions (Table~\ref{tab:raw_score_distributions}).

\begin{table}[t]
\centering
\footnotesize
\caption{Raw Score Distributions}
\label{tab:raw_score_distributions}
\begin{tabular}{lcc}
\toprule
\textbf{Metric} & \textbf{$\mu$} & \textbf{$\sigma^2$} \\
\midrule
\multicolumn{3}{l}{\textit{Color Harmony}} \\
Mean Distance & 0.202 & 0.131 \\
Deck Mean Score & 0.869 & 0.039 \\
Deck Consistency & 0.147 & 0.018 \\
\midrule
\multicolumn{3}{l}{\textit{Colorfulness}} \\
Mean & 51.09 & 767.72 \\
Std & 11.28 & 72.89 \\
\midrule
\multicolumn{3}{l}{\textit{Subband Entropy}} \\
Mean & 3.878 & 0.533 \\
Std & 0.457 & 0.036 \\
\bottomrule
\end{tabular}
\end{table}

\subsubsection{Stage 2: Initial Parameter Selection}
\begin{figure}[t]
\centering
\includegraphics[width=\columnwidth]{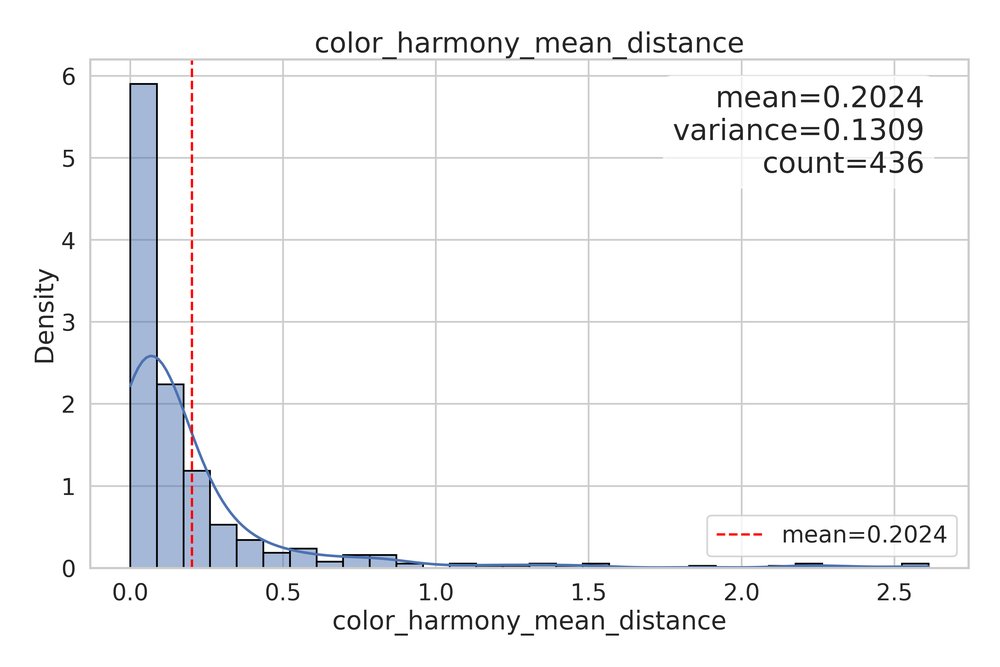}
\caption{Color Harmony distance distribution ($\bar{D} = 0.202$).}
\label{fig:color_harmony_dist}
\end{figure}

Based on the empirical distributions (Figures~\ref{fig:subband_entropy_dist}--\ref{fig:color_harmony_dist}), we established initial parameter values capturing international corporate design standards (e.g., Duarte's principles) to minimize localized stylistic bias.

\subsubsection{Stage 3: Human Preference Optimization}
We constructed a verification set of 1000 slide pairs sampled from the initial 3,000 slides with human preference rankings. Using Bayesian optimization, we tuned parameters to maximize Spearman correlation ($\rho$) between metric scores and human rankings.

%----------------------------------------------------------
\subsection{SubbandEntropyMetric Parameters}
\label{subsec:subband_entropy_params}

The Subband Entropy metric measures visual clutter through steerable pyramid decomposition~\cite{rosenholtz2007measuring}. The metric decomposes images in the CIE-LAB color space and computes Shannon entropy across subbands:
\begin{equation}
SE = w_L \bar{H}_L + w_{ab} \bar{H}_a + w_{ab} \bar{H}_b
\label{eq:subband_entropy}
\end{equation}
where $\bar{H}_c = \frac{1}{N_c}\sum_{i=1}^{N_c} H(S_c^i)$ is the average entropy across subbands for channel $c$, with adaptive binning (bins $= \sqrt{|S|}$).

\begin{table}[t]
\centering
\footnotesize
\caption{SubbandEntropyMetric Parameters}
\label{tab:subband_params}
\begin{tabular}{lccp{3.2cm}}
\toprule
\textbf{Param} & \textbf{Sym} & \textbf{Val} & \textbf{Rationale} \\
\midrule
Lum. Weight & $w_L$ & 0.84 & From~\cite{rosenholtz2007measuring}; luminance dominates clutter perception \\
Chrom. Weight & $w_{ab}$ & 0.08 & Equal weight per channel; $\sum w = 1.0$ \\
Pyramid Levels & $L$ & 3 & Multi-scale without excess computation \\
Orientations & $K$ & 4 & Standard config (0°, 45°, 90°, 135°) \\
Zero Thresh. & $\tau_0$ & 0.008 & Low-variation channels ignored \\
\bottomrule
\end{tabular}
\end{table}

\begin{figure}[t]
\centering
\includegraphics[width=\columnwidth]{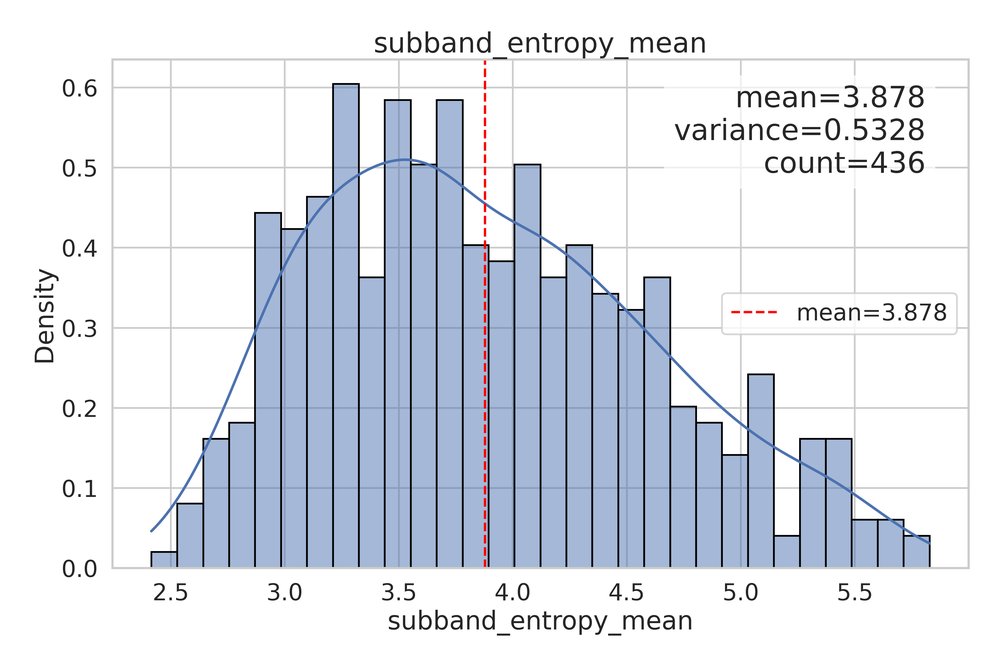}
\caption{Distribution of Subband Entropy scores ($\mu = 3.878$).}
\label{fig:subband_entropy_dist}
\end{figure}

Figure~\ref{fig:subband_entropy_dist} shows the distribution with a mean of 3.878 bits, aligning with structured presentation content.

%----------------------------------------------------------
\subsection{ColorfulnessMetric Parameters}
\label{subsec:colorfulness_params}

The Colorfulness metric quantifies chromatic variety using the Hasler \& Süsstrunk method~\cite{hasler2003measuring}:
\begin{equation}
C = \sqrt{\sigma_{rg}^2 + \sigma_{yb}^2} + 0.3 \sqrt{\mu_{rg}^2 + \mu_{yb}^2}
\label{eq:colorfulness}
\end{equation}
where $rg = R - G$ and $yb = 0.5(R + G) - B$.

\begin{figure}[t]
\centering
\includegraphics[width=\columnwidth]{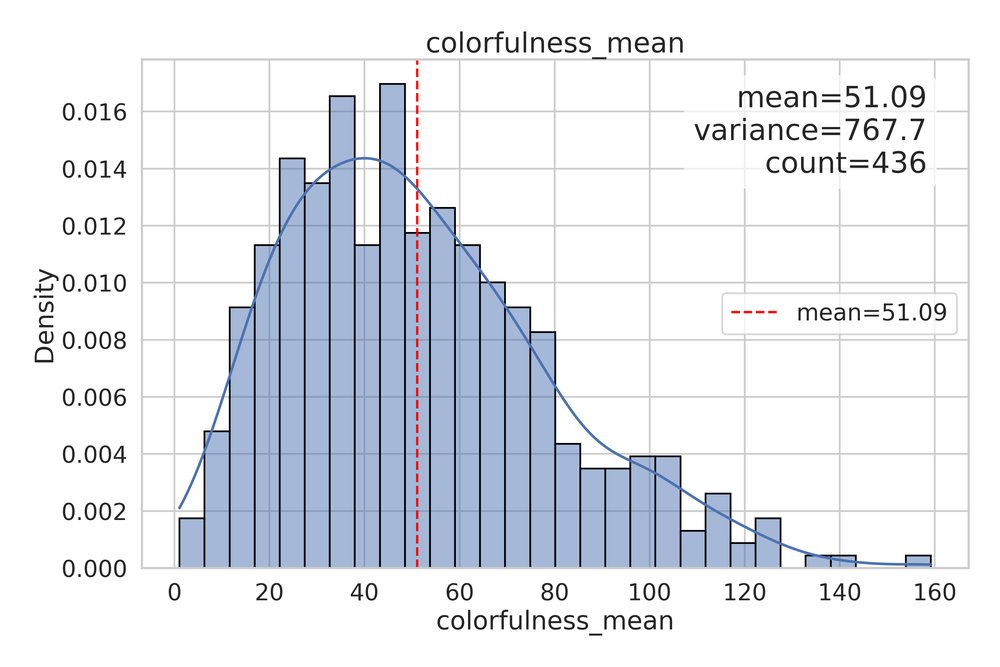}
\caption{Colorfulness distribution ($\mu \approx 50$, $\sigma = 27.7$).}
\label{fig:colorfulness_dist}
\end{figure}

The coefficient $\alpha = 0.3$ was empirically derived through psychophysical experiments, balancing chromatic spread ($\sigma$) and bias ($\mu$). Figure~\ref{fig:colorfulness_dist} shows professional slides exhibit moderate colorfulness ($\mu = 51.09$).

%----------------------------------------------------------
\subsection{ColorHarmonyMetric Parameters}
\label{subsec:color_harmony_params}

The Color Harmony metric evaluates adherence to harmonic color templates~\cite{cohen2006color}. For each pixel $p$ with hue $H_p$ and saturation $S_p$:
\begin{equation}
D_{\text{template}}(\alpha) = \frac{\sum_p S_p \cdot d(H_p, T_\alpha)}{\sum_p S_p}
\label{eq:harmony_distance}
\end{equation}
where $d(H_p, T_\alpha)$ is the angular distance to template $T$ rotated by $\alpha$. The slide score uses Gaussian decay:
\begin{equation}
S_{\text{slide}} = \exp\left(-\bar{D}^2 / 2\sigma^2\right)
\label{eq:harmony_score}
\end{equation}

\begin{table}[t]
\centering
\footnotesize
\caption{Harmonic Templates (center, width as fractions of 360°)}
\label{tab:harmonic_templates}
\begin{tabular}{lll}
\toprule
\textbf{Type} & \textbf{Sectors} & \textbf{Desc.} \\
\midrule
i & (0, 0.05) & Monochromatic \\
V & (0, 0.26) & Analogous \\
L & (0, 0.05), (0.25, 0.22) & Split-comp. \\
I & (0, 0.05), (0.50, 0.05) & Complementary \\
T & (0.25, 0.50) & Triadic \\
Y & (0, 0.26), (0.50, 0.05) & Split-comp. var. \\
X & (0, 0.26), (0.50, 0.26) & Double comp. \\
\bottomrule
\end{tabular}
\end{table}

\begin{table}[t]
\centering
\footnotesize
\caption{ColorHarmonyMetric Parameters}
\label{tab:harmony_params}
\begin{tabular}{lccp{3cm}}
\toprule
\textbf{Param} & \textbf{Sym} & \textbf{Val} & \textbf{Rationale} \\
\midrule
Gaussian $\sigma$ & $\sigma$ & 0.0005 & Strictness; optimized via human pref. \\
Angular Res. & $N$ & 360 & Full-degree granularity \\
Sat. Thresh. & $S_{\min}$ & 0.1 & Exclude achromatic pixels \\
\bottomrule
\end{tabular}
\end{table}

\noindent\textbf{Deck-Level Scoring:}
\begin{equation}
\text{Score} = 5 \mu_{\text{deck}} - 30 \sigma_{\text{deck}}
\label{eq:deck_score}
\end{equation}
rewarding high average harmony while penalizing inconsistency.

%----------------------------------------------------------
\subsection{VisualHRVMetric Parameters}
\label{subsec:visual_hrv_params}

The Visual HRV metric assesses presentation pacing by measuring variability across consecutive slides, inspired by physiological HRV analysis~\cite{van1993heart}. For a raw entropy value $x$, calculate the normalized Score $S$ (0 to 1):

$$S(x) = \frac{1}{1 + e^{-k \cdot (x - \mu)}}$$,
where
$\mu$ (Midpoint): The value of a "standard" slide,
$k$ (Slope): Sensitivity,
$x$: The raw Subband Entropy of the current slide.

Given normalized scores $\{S_1, \ldots, S_n\} \in [0, 1]$:
\begin{equation}
\text{RMSSD} = \sqrt{\frac{1}{n-1}\sum_{i=1}^{n-1} |S_{i+1} - S_i|^2}
\label{eq:rmssd}
\end{equation}

Overload detection uses moving average $\bar{S}_j^{(w)} = \frac{1}{w}\sum_{k=0}^{w-1} S_{j+k}$. The final score:
\begin{equation}
\text{Score} = 100 \left(1 - \frac{|\text{RMSSD} - \tau|}{\tau_w}\right) - p \cdot N_{\text{overload}}
\label{eq:hrv_score}
\end{equation}

\begin{table}[t]
\centering
\footnotesize
\caption{VisualHRVMetric Parameters}
\label{tab:hrv_params}
\begin{tabular}{lccp{3cm}}
\toprule
\textbf{Param} & \textbf{Sym} & \textbf{Val} & \textbf{Rationale} \\
\midrule
Target RMSSD & $\tau$ & 0.03 & Ideal variability center \\
Half-width & $\tau_w$ & 0.2 & Range $[0, 0.5]$ \\
Overload Win. & $w$ & 3 & Cognitive processing unit \\
Overload Thr. & $\theta$ & 0.75 & High-complexity flag \\
Penalty & $p$ & 10.0 & Per overload event \\
\bottomrule
\end{tabular}
\end{table}

\begin{table}[t]
\centering
\footnotesize
\caption{VisualHRV Interpretation Bands}
\label{tab:hrv_interpretation}
\begin{tabular}{lcp{3.5cm}}
\toprule
\textbf{Category} & \textbf{RMSSD} & \textbf{Interpretation} \\
\midrule
Flatline & $< 0.01$ & Monotonous; disengaging \\
Healthy & $[0.01, 0.1]$ & Optimal visual rhythm \\
Transitional & other & Borderline pacing \\
Strobe Light & $> 0.30$ & Jarring transitions \\
\bottomrule
\end{tabular}
\end{table}

%----------------------------------------------------------
\subsection{Parameter Optimization Results}
\label{subsec:optimization_results}

Using 1000 human-ranked slide pairs, we performed grid search to maximize Spearman $\rho$ (Table~\ref{tab:optimization_results}).

\begin{table}[t]
\centering
\footnotesize
\caption{Parameter Optimization Results}
\label{tab:optimization_results}
\begin{tabular}{llccc}
\toprule
\textbf{Metric} & \textbf{Param} & \textbf{Init} & \textbf{Opt} & \textbf{$\Delta\rho$} \\
\midrule
ColorHarmony & $\sigma$ & 1 & 0.01 & +0.08 \\
VisualHRV & $\tau$ & 0.1 & 0.03 & +0.05 \\
SubbandEntropy & $w_L$ & 0.84 & 0.84 & -- \\
Colorfulness & $\alpha$ & 0.30 & 0.30 & -- \\
\bottomrule
\end{tabular}
\end{table}

SubbandEntropy and Colorfulness parameters retained literature values ($p > 0.05$). ColorHarmony $\sigma$ showed the largest sensitivity, with a stricter threshold better distinguishing professional designs.

% Furthermore, to demonstrate the inherent robustness of our metric design without any preference fitting, we evaluated an ``untuned'' version using default heuristic weights directly from prior literature (e.g., \citet{hasler2003measuring,caldwell2008web,rosenholtz2007measuring,cohen2006color}). Remarkably, even without any parameter tuning, these default computational metrics achieve a Spearman correlation of $\sim$0.62, still substantially outperforming the LLM-as-a-Judge baseline (0.57). Tuning on the disjoint 1000-pair set further pushes this alignment to 0.710.

\subsection{Summary}
Our tuning combines principled estimates from empirical distributions with human judgment refinement:
\begin{itemize}
\item \textbf{SubbandEntropy}: Literature weights ($w_L\!=\!0.84$, $w_{ab}\!=\!0.08$) optimal
\item \textbf{Colorfulness}: Hasler-Süsstrunk $\alpha\!=\!0.3$ generalizes well
\item \textbf{ColorHarmony}: Stricter $\sigma\!=\!0.01$ captures professional standards
\item \textbf{VisualHRV}: $\tau\!=\!0.03$, $\theta\!=\!0.75$ balances variety vs.\ cognitive load
\end{itemize}

\section{Layout Detection Examples}
\label{app: ocr}

We employ the PP-DocLayout\_plus-L model~\cite{cui2025paddleocr} to detect and localize textual elements within presentation slides. This deep learning-based layout analysis model provides precise bounding box coordinates for various element types, including document titles, body text, images, and footers. Each detected element is assigned a confidence score indicating the model's certainty in the detection.

\subsection{Detection Output Format}

The model outputs a structured JSON format containing the following information for each detected element:
\begin{itemize}
    \item \textbf{label}: The semantic category of the detected region (e.g., \texttt{text}, \texttt{image}, \texttt{doc\_title}, \texttt{footer})
    \item \textbf{score}: Confidence score ranging from 0 to 1, representing the model's certainty
    \item \textbf{coordinate}: Bounding box coordinates in the format $[x_{\min}, y_{\min}, x_{\max}, y_{\max}]$
\end{itemize}

\subsection{Example for OCR Detections}

\begin{figure}[t]
    \centering

    \subfigure[Title slide with a full-width background image, document title, subtitle text, and footer.]{
        \includegraphics[width=0.8\columnwidth]{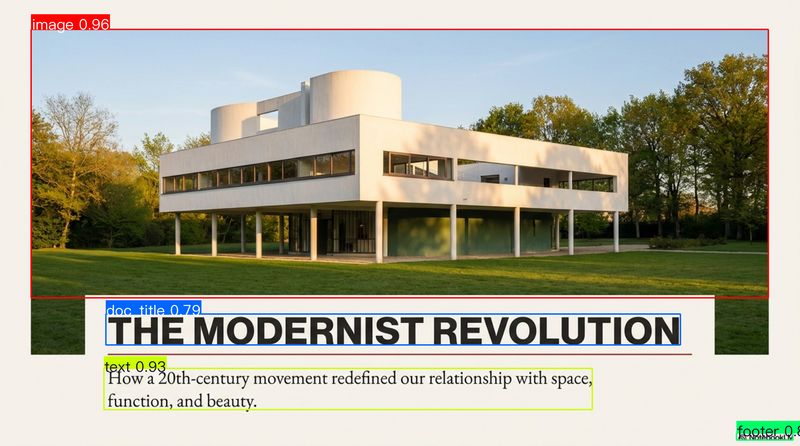}
        \label{fig:detection_slide1}
    }
    \hfill
    \subfigure[Content slide with side-by-side images, title, dual text blocks, and footer.]{
        \includegraphics[width=0.8\columnwidth]{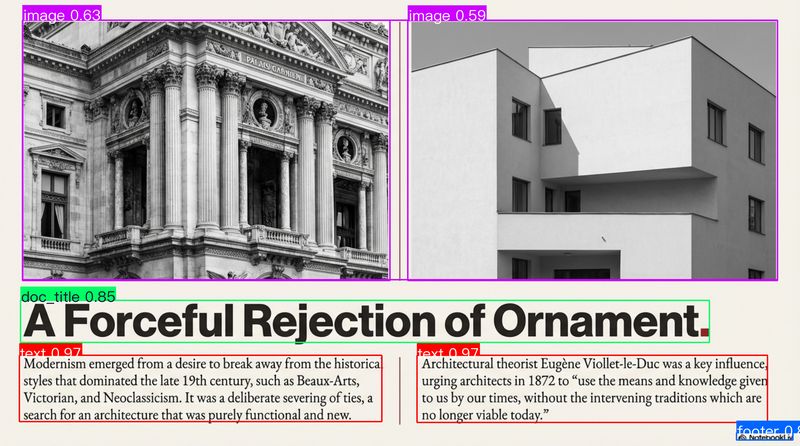}
        \label{fig:detection_slide2}
    }

    \vspace{0.5em}

    \subfigure[Two-column layout with title, multiple text regions, and a large image on the right.]{
        \includegraphics[width=0.8\columnwidth]{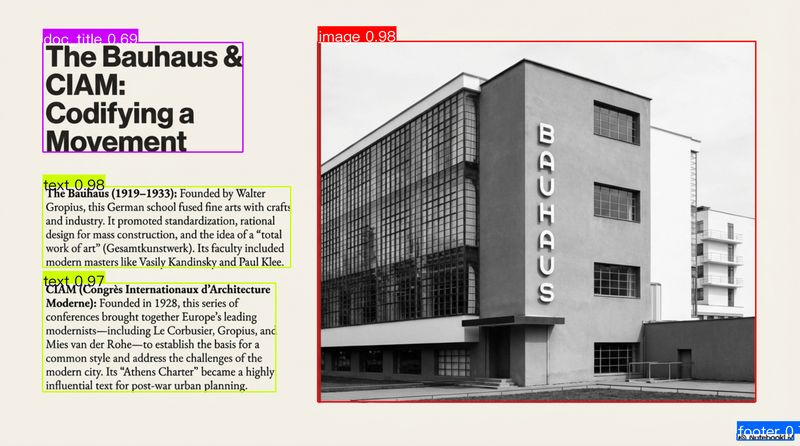}
        \label{fig:detection_slide5}
    }

    \caption{Layout detection results using PP-DocLayout\_plus-L on presentation slides. Bounding boxes are color-coded by element type: images, document titles, text, and footers \etc. Confidence scores are displayed alongside each detection.}
    \label{fig:layout_detection_examples}
\end{figure}

Figure~\ref{fig:layout_detection_examples} illustrates the layout detection results on representative slides from the \textit{Modern Architecture} presentation. The detected bounding boxes are visualized with different colors corresponding to element types: titles (green), body text (blue), images (red), and footers (gray).

% \begin{figure}[t]
%     \centering
%     \begin{minipage}[b]{0.5\textwidth}
%         \includegraphics[width=\textwidth]{detection_slide_0001_res.jpg}
%         \caption{Title slide with a full-width background image, document title, subtitle text, and footer.}
%         \label{fig:detection_slide1}
%     \end{minipage}
%     \hfill
%     \begin{minipage}[b]{0.5\textwidth}
%         \includegraphics[width=\textwidth]{detection_slide_0002_res.jpg}
%         \caption{Content slide with side-by-side images, title, dual text blocks, and footer.}
%         \label{fig:detection_slide2}
%     \end{minipage}
%     \hfill
%     \begin{minipage}[b]{0.5\textwidth}
%         \includegraphics[width=\textwidth]{detection_slide_0005_res.jpg}
%         \caption{Two-column layout with title, multiple text regions, and a large image on the right.}
%         \label{fig:detection_slide5}
%     \end{minipage}
%     \caption{Layout detection results using PP-DocLayout\_plus-L on presentation slides. Bounding boxes are color-coded by element type:images, document titles, text, and footers \etc. Confidence scores are displayed alongside each detection.}
%     \label{fig:layout_detection_examples}
% \end{figure}

\subsection{Detection Statistics}

Table~\ref{tab:detection_stats} summarizes the detection results across the example presentation slides. The model demonstrates high accuracy in identifying textual elements, with average confidence scores exceeding 0.90 for text regions and 0.80 for document titles.

\begin{table}[htbp]\small
    \centering
    \caption{Summary of layout detection results on example slides.}
    \label{tab:detection_stats}
    \begin{tabular}{lccc}
        \toprule
        \textbf{Element Type} & \textbf{Count} & \textbf{Avg. Score} & \textbf{Min. Score} \\
        \midrule
        Document Title & 15 & 0.78 & 0.69 \\
        Text & 28 & 0.95 & 0.87 \\
        Image & 22 & 0.82 & 0.59 \\
        Footer & 15 & 0.79 & 0.78 \\
        \bottomrule
    \end{tabular}
\end{table}

By leveraging these precise bounding boxes, we can accurately extract text content from presentation slides while avoiding interference from complex background images, decorative elements, and other non-textual components. This approach ensures that the text extraction process focuses exclusively on genuine textual content, improving the quality of downstream processing tasks.

\section{Visual Rhythm Score Formulation Details}

This appendix provides the detailed mathematical steps for calculating the Visual Rhythm Score described in the main text.

\subsection{Subband Entropy ($E_{SE}$)}\label{app: subband}
The calculation of Subband Entropy for a single slide follows the methodology of Rosenholtz et al.~\cite{rosenholtz2007measuring}.

\paragraph{1. Color Space Normalization.}
To decouple luminance from chromaticity, we operate in the CIELAB color space. The channels are first normalized to a unit range:
\begin{align}
    L' &= L/100 \\
    \{a', b'\} &= (\{a, b\} + 128)/255
\end{align}

\paragraph{2. Channel Decomposition and Entropy.}
We decompose each normalized channel $C \in \{L', a', b'\}$ using a Steerable Pyramid with 3 scales and 4 orientations. This simulates the receptive fields of the primary visual cortex (V1). For the resulting set of subbands $\{S_1, \dots, S_n\}$, we calculate the mean Shannon entropy for each channel:
\begin{equation}
    E_C = \frac{1}{n}\sum_{i=1}^{n} \left( -\sum_{k} p_k \log_2(p_k) \right)_{S_i}
\end{equation}
where $p_k$ is the probability of a given intensity level $k$ within a subband $S_i$.

\paragraph{3. Composite Score.}
The composite Subband Entropy $E_{SE}$ is a weighted sum emphasizing luminance, as the human visual system is most sensitive to structural contrast. The weight for the chrominance channels is set to $w_c = 0.0625$. The final raw entropy is mapped to a scalar score $S_{entropy}$ using a Gaussian function centered on an optimal complexity mean $\mu_{opt}$.

\subsection{Temporal Pacing (RMSSD)}
The RMSSD calculation quantifies the variability in visual complexity across the slide sequence.

\paragraph{1. Calculating Successive Differences.}
First, we compute the flux $\Delta_i$ as the absolute change in the complexity score $S_{entropy}$ between adjacent slides $i$ and $i+1$:
\begin{equation}
    \Delta_i = |S^{(i+1)}_{entropy} - S^{(i)}_{entropy}|
\end{equation}
This yields a sequence of differences $[\Delta_1, \dots, \Delta_{N-1}]$.

\paragraph{2. Root Mean Square.}
The RMSSD is then calculated as the root mean square of this sequence of successive differences, as shown in the main text.

\section{Aesthetic Metrics: Interpretation and Insights}
\label{app:aesthetic_insights}
\subsection{Color Harmony: Interpretation and Insights}
\label{app:color_harmony_insights}
This metric is designed to detect “color clutter” by measuring how well a slide’s hue distribution fits established harmonic templates~\cite{cohen2006color}. Intuitively, professional decks maintain a coherent hue structure across slides (low deviation) even when layouts and content types change, which preserves brand identity and reduces perceptual noise.

\paragraph{Good Pattern (Low Deviation / Stable Profile).}
A well-designed deck typically stays within a small deviation band (often near $0^\circ$ to $10^\circ$ in practice~\cite{luo2008photo}), consistent with a chosen template (e.g., split-complementary). Even as the deck transitions from title to data or diagrams, hues remain constrained to the same template sectors, creating a unified palette.

\paragraph{Bad Pattern (Spikes / Palette Violations).}
Sudden spikes in harmonic deviation often arise from inserted assets (stock photos, copied charts, screenshots) whose palettes conflict with the established template. These deviations are perceived as clutter and can increase the viewer’s cognitive load by breaking the deck’s visual “contract.”

\subsection{Engagement: Interpretation and Insights}
\label{app:engagement_insights}
Harmony alone can yield visually correct but emotionally flat slides. The adapted colorfulness metric $M$ captures perceived vibrancy~\cite{hasler2003measuring}, while the pacing term targets temporal coherence in that vibrancy across slides, aligning with the idea that presentations should not feel static or erratic~\cite{post2017unity}.

\paragraph{Volatile Pacing (“Jumpscare” Effect).}
Large slide-to-slide swings in $M$ (high $\sigma_{pacing}$) create visual strobing: the audience repeatedly re-adapts to different intensity regimes, which feels jarring and amateurish.

\paragraph{Near-Zero Pacing (“Ghostly” Effect).}
If $M$ is consistently low and $\sigma_{pacing}\approx 0$, the deck may be formally tidy yet perceptually monotonous. This pattern often corresponds to over-reliance on grayscale text-on-white with minimal accent colors, which fails to sustain attention.

\paragraph{Ideal State (Controlled Band).}
High-quality decks maintain a stable “plateau” of vibrancy with moderate, purposeful variation—enough to mark emphasis and section boundaries without causing volatility.

\subsection{Usability (Figure-Ground Contrast): Interpretation and Insights}
\label{app:usability_insights}
Aesthetic choices must preserve legibility~\cite{kohler1967gestalt}. By computing contrast within detected text regions, the metric targets the local perceptual condition that determines readability, rather than relying on global image statistics.

\paragraph{The “Fog” Pattern (Low Local Contrast).}
Low contrast ratios in text regions (e.g., light text on light background) make information inaccessible, especially for visually impaired users or when projected in bright rooms. Such failures may remain hidden if only global contrast is considered.

\paragraph{Why Layout-Aware Contrast Matters (Local vs.\ Global).}
A slide can have high global contrast (e.g., dark background with a bright shape) yet still place text in a locally low-contrast region (e.g., dark gray on black). Region-level evaluation correctly flags the usability failure that histogram-based or global metrics can miss.

\subsection{Visual Rhythm (VisualHRV): Interpretation and Insights}
\label{app:visualhrv_insights}
Presentations are temporal media; beyond per-slide complexity, the sequence structure matters for sustained comprehension~\cite{duarte2010resonate}. VisualHRV connects computational cues (entropy-derived clutter and its temporal change) with pacing principles in cognitive psychology.

\paragraph{Static vs.\ Dynamic Cost.}
Subband Entropy proxies “feature congestion” and the cost of visual search~\cite{rosenholtz2007measuring}. However, engagement and comprehension depend on how this cost evolves over time, not only on a single slide’s burden.

\paragraph{Effective Rhythm (Non-Flat, Intentional Variation).}
Motivated by narrative “sparkline” framing~\cite{duarte2010resonate} and cognitive load theory~\cite{sweller1988cognitive}, effective decks often alternate between dense and sparse slides, enabling consolidation and preventing working-memory overload. This manifests as meaningful temporal variability (higher RMSSD) rather than a uniform sequence.

\paragraph{Failure Mode (High Complexity + Low Variation).}
A persistently demanding sequence with low fluctuation (low RMSSD at high mean complexity) can induce sustained cognitive strain and audience disengagement over time (a “flatline” difficulty profile), consistent with attention fatigue under continuous high perceptual load.

\section{Style-Transfer Validation of Aesthetic Metrics}
\label{app:style_transfer}

To examine whether the aesthetic metrics transfer across visual styles, we ran a style-transfer validation over 1,435 Slides-Align1.5k aesthetics records spanning six style buckets. Table~\ref{tab:style_transfer} summarizes the key diagnostics.

\begin{table*}[htb]
\centering
\small
\caption{Style-transfer diagnostics of the aesthetic metrics over 1,435 Slides-Align1.5k records.}
\label{tab:style_transfer}
\begin{NiceTabular}{p{3.4cm}p{6.6cm}p{4.8cm}}
\toprule
\textbf{Diagnostic} & \textbf{Result} & \textbf{Interpretation} \\
\midrule
Style coverage & 1,435 records across six buckets: standard corporate/mixed 959 (66.8\%), minimalist-light 212 (14.8\%), image-rich marketing 124 (8.6\%), diagrammatic/scientific 91 (6.3\%), data-heavy report 37 (2.6\%), dark-mode minimalist 12 (0.8\%) & The validation covers multiple visual styles, while standard corporate/mixed remains the dominant bucket and is disclosed. \\
Dark-mode check & Dark decks have slightly better human mean rank than light decks (3.84 vs.\ 4.27), higher figure-ground contrast (0.764 vs.\ 0.734), and similar corrected colorfulness (0.570 vs.\ 0.585) & We do not find a systematic dark-mode penalty. \\
Leave-one-style-out & Removing any single style bucket changes the composite alignment score by less than 0.05 & The robustness result is not driven by one style bucket alone. \\
Boundary cases & Image-rich marketing decks have larger rank errors; harmony can penalize creative/non-standard palettes; visual rhythm is lower on dark images because of the entropy computation & The metrics should be interpreted as style-tested proxies, not universal aesthetics measures. \\
\bottomrule
\end{NiceTabular}
\end{table*}

These results are reassuring but also help us state the limitations precisely. We present the aesthetic metrics as human-aligned proxies within the tested style distribution, disclose the style composition above, and note that domain- or style-specific recalibration may still be needed for styles outside this distribution.

\section{Detailed Specifications of the PEI Taxonomy}\label{app: PEI}

This appendix provides the technical criteria and definitions for the Presentation Editability Intelligence (PEI) levels introduced in Section \ref{subsec: edit}. Table~\ref{tab:pei_levels} shows the Technical Hallmark and Critical Failure Condition for each level.

\subsection{Executive Summary}
The \textbf{Presentation Editability Intelligence (PEI)} framework is a hierarchical standard for evaluating the structural integrity, semantic logic, and editability of AI-generated presentations.
Unlike traditional metrics that measure visual similarity (e.g., FID scores), PEI measures \textbf{Editability Depth}. It asserts that a professional presentation is not merely a static image but a complex database of relationships defined by the Office Open XML standard.

\subsubsection{The Knockout Rule (Dependency Logic)}
The framework operates on a strict \textbf{dependency-based knockout mechanism}.
\begin{itemize}
    \item Higher levels (e.g., L4 Data) rely on the existence of lower levels (e.g., L3 Structure).
    \item  {Evaluation Protocol:} If a file fails a specific criterion, evaluation ceases immediately. The file is assigned the highest level it successfully completes.
    \item \textit{Example:} A file with perfect animations (L5) but broken charts (L4 failure) is classified as  {Level 3}.
\end{itemize}

\subsection{Input Triage \& Routing}
The evaluation process begins with  {Input Format Analysis}, which determines the evaluation pipeline and the  {Maximum Attainable Level (MAL)}.
\subsubsection{Scenario A: The Static Input}
\begin{itemize}
    \item  {Supported Formats:} \texttt{.pdf}, \texttt{.png}, \texttt{.jpg}, \texttt{.jpeg}
    \item  {Protocol:}  {Immediate Termination}.
    \item  {Maximum Attainable Level:}  {L0}
    \item  {Technical Rationale:} These formats are flattened raster containers. They do not support object separation, text reflow, or XML data binding.
\end{itemize}
\subsubsection{Scenario B: The Web Input}
\begin{itemize}
    \item  {Supported Formats:} \texttt{URL} (Web Viewers, HTML5 Decks, Online Canvas links)
    \item  {Protocol:}  {Visual 
    \& Interactive Inspection}.
    \item  {Maximum Attainable Level:}  {L2 (Vector)}
    \item  {Technical Rationale:} Web viewers render the Document Object Model (DOM) visually but obscure the underlying file structure. It is technically impossible to verify deep editability features—such as Master Slide inheritance (\texttt{p:sldMaster}) or embedded Excel binary binding (\texttt{c:chart})—through a standard web interface. Therefore, the score is capped at the limit of visual verification (L2).
\end{itemize}
\subsubsection{Scenario C: The Native Input}
\begin{itemize}
    \item  {Supported Formats:} \texttt{.pptx}, \texttt{.potx} (Office Open XML formats)
    \item  {Protocol:}  {Full Deep-Scan Evaluation}.
    \item  {Maximum Attainable Level:}  {L5 (Cinematic)}
    \item  {Technical Rationale:} These files provide full access to the XML schema, allowing verification of Master Slides, Data Relationships, and Animation Timings.
\end{itemize}
 
\subsection{The PEI Hierarchy: Detailed Definitions}
This section defines the technical criteria for each level.
\begin{table*}[t]
\centering
\caption{The PEI Hierarchical Framework. This table utilizes vertical color coding to contrast capabilities with limitations. The \colorbox{peiGreen}{Technical Hallmark} column identifies the core value proposition, while the \colorbox{peiRed}{Critical Failure Condition} column highlights the ``knockout'' factors that disqualify a system.}
\label{tab:pei_levels}

\resizebox{\textwidth}{!}{%
\begin{tabular}{c l l >{\columncolor{peiGreen}}l >{\columncolor{peiRed}}p{6cm}} 
\toprule

% 表头
\rowcolor{headerGray} 
\textbf{Level} & \textbf{Class Name} & \textbf{Operational Status} & \textbf{Technical Hallmark} & \textbf{Critical Failure Condition} \\ \midrule

% --- L5 ---
\textbf{L5} & \textit{Cinematic} & \textbf{Dynamic Experience} & Animation Logic, Media Embeds & Static slides only; Videos treated as static images; No temporal transitions. \\ 
\addlinespace[1.5ex] %稍微增加间距

% --- L4 ---
\textbf{L4} & \textit{Parametric} & \textbf{Enterprise Tool} & Native Data (\texttt{<c:chart>}), SmartArt & Charts drawn as vector shapes; Data uneditable (broken Excel link). \\ 
\addlinespace[1.5ex]

% --- L3 ---
\textbf{L3} & \textit{Structural} & Functional Tool & Global Masters (\texttt{<p:sldMaster>}) & Hardcoded layouts (no master inheritance); Logical groups ungrouped. \\ 
\addlinespace[1.5ex]

% --- L2 ---
% 修正了 be- come 和末尾多余的 &
\textbf{L2} & \textit{Vector} & \textit{Visual Toy} & SVG Paths, Scalable Primitives & Icons or diagrams become blurry when zoomed in; Text fragmentation. \\ 
\addlinespace[1.5ex]

% --- L1 ---
% 修正了标点空格，优化了语义
\textbf{L1} & \textit{Patchwork} & \textit{Text-Editable Toy} & Raster Backgrounds + OCR Text & Rasterized Text (OCR failed); Non-selectable background elements. \\ 
\addlinespace[1.5ex]

% --- L0 ---
% --- L0 ---
% 解释：
% 1. \multicolumn{1}{...c}：强制单元格居中，并保留红色背景
% 2. \tikz{...}：画一条线，(0,0.4)是左上，(6,0)是右下
\textbf{L0} & \textit{Static} & \textit{Flat Image} & Text is rasterized pixels & \multicolumn{1}{>{\columncolor{peiRed}}c}{\tikz[baseline=-0.3ex]{\draw[line width=0.5pt] (0,0.4) -- (6,0);}} \\ \bottomrule

\end{tabular}%
}
\end{table*}

\subsubsection{Phase 0: The Flat Phase}
\paragraph{Level 0: Static (The Flat Image)} % Note: Emojis removed for compatibility
\begin{itemize}
    \item  {Definition:} The content is indistinguishable from a static bitmap.
    \item  {Technical Hallmark:} Content is flattened. Text is rasterized pixels, not character strings.
    \item Critical Failure Condition (Knockout):N/A (Bottom of the hierarchy)
\end{itemize}
\begin{figure}[H]
    \centering
    % TODO: Replace with absolute path
    \includegraphics[width=0.4\textwidth]{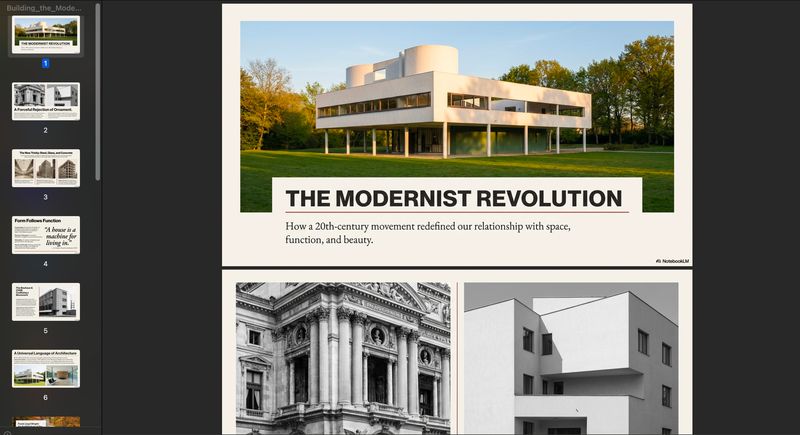}
    \caption{Level 0: Static (Uneditable PDF)}
\end{figure}
\subsubsection{Phase 1: The Visual Phase (Surface Fidelity)}
\paragraph{Level 1: Patchwork (The Text-Editable Toy)}
\begin{itemize}
    \item  {Definition:} The file contains editable elements, but they are fragmented and structurally broken.
    \item  {Technical Hallmark:} "OCR-style" reconstruction. Paragraphs are split into multiple single-line text boxes. Layouts rely on absolute positioning coordinates rather than flow.
    \item  {Critical Failure Condition (Knockout):}
\begin{itemize}
    \item \textbf{Rasterized Text:} The "text" looks like letters but is actually an image (bitmap). Users cannot select, copy, or edit the characters.
    \item \textbf{No Selectability:} Clicking on content selects the entire slide background instead of individual elements.
\end{itemize}

\end{itemize}
\begin{figure}[t]
    \centering
    \subfigure[Fragmented OCR text blocks]{
        \includegraphics[width=0.8\columnwidth]{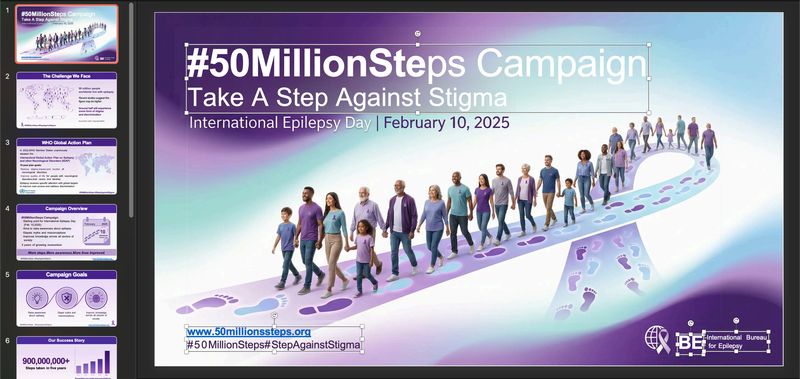}
        \label{fig:L1-1}
    }
    \hfill
    \subfigure[Broken background elements ]{
        \includegraphics[width=0.8\columnwidth]{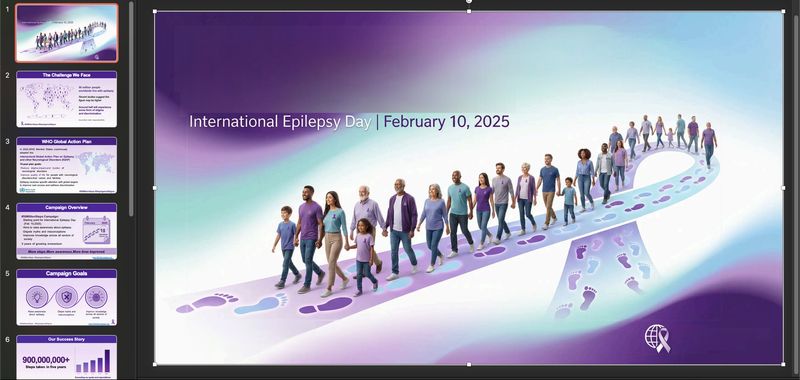}
        \label{L1-2}
    }
    \caption{Level 1 Examples }
    \label{fig:L1figures}
\end{figure}

\paragraph{Level 2: Vector (The Visual Toy)}
\begin{itemize}
    \item  {Definition:} Visual clarity is achieved via vector graphics, but elements lack logical grouping.
    \item  {Technical Hallmark:} Usage of SVG paths and Scalable Primitives. Graphics remain sharp at 400\% zoom.
    \item  {Critical Failure Condition (Knockout):}
\begin{itemize}
    \item \textbf{Pixelated Graphics:} Icons or diagrams become blurry when zoomed in (indicating they are Screenshots, not Vectors).
    \item \textbf{Text Fragmentation:} Paragraphs are broken into separate text boxes for each line (failing the "Reflow" requirement of a true Vector container).
\end{itemize}

\end{itemize}

\begin{figure}[t]
    \centering
    \subfigure[Scalable primitives]{
        \includegraphics[width=0.8\columnwidth]{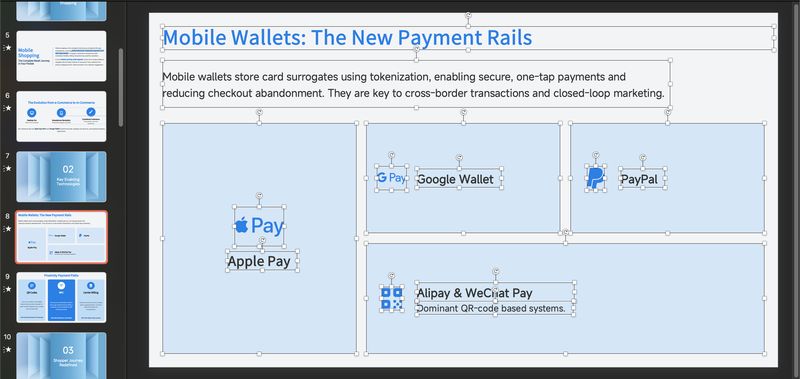}
        \label{fig:L2-1}
    }
    \hfill
    \subfigure[SVG icons and paths]{
        \includegraphics[width=0.8\columnwidth]{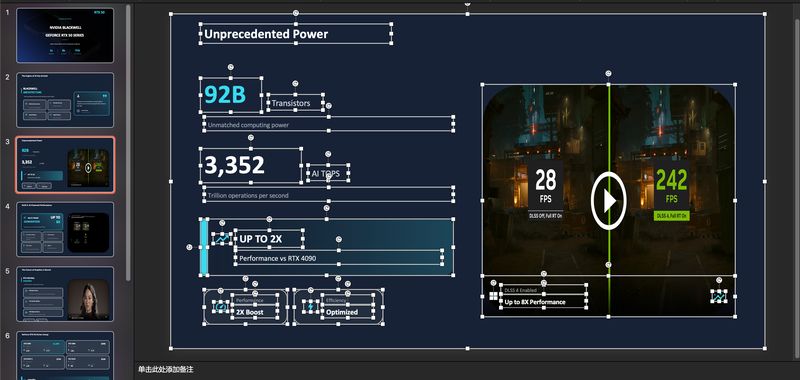}
        \label{L2-2}
    }
    \caption{Level 2 Examples }
    \label{fig:L1figures}
\end{figure}

\subsubsection{Phase 2: The Structural Phase (Logic \& Data)}
\paragraph{Level 3: Structural (The Functional Tool)}
\begin{itemize}
    \item  {Definition:} The system adheres to presentation software logic (Masters and Grouping).
    \item  {Technical Hallmark:}
    \begin{itemize}
        \item  {Logical Grouping:} Related vector elements are bound using the Group function.
        \item  {Master Inheritance:} The file utilizes the \texttt{<p:sldMaster>} schema. Layout changes in the Master View propagate globally to all slides.
    \end{itemize}
    \item  {Critical Failure Condition (Knockout):}
\begin{itemize}
    \item \textbf{Atomic Isolation:} Complex graphics consist of hundreds of loose shapes requiring individual selection (No Grouping).
    \item \textbf{Hardcoded Backgrounds:} Background elements are pasted onto every individual slide rather than inherited from the Master Slide.
\end{itemize}

\end{itemize}

\begin{figure}[t]
    \centering
    \subfigure[The use of Grouping]{
        \includegraphics[width=0.8\columnwidth]{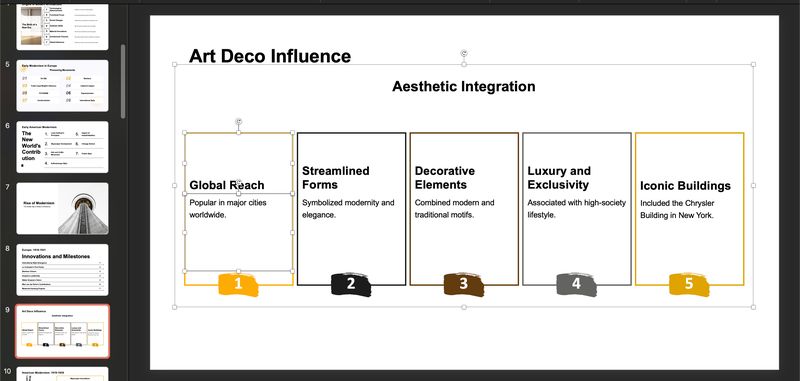}
        \label{fig:L2-1}
    }
    \hfill
    \subfigure[Master of the origin slides]{
        \includegraphics[width=0.8\columnwidth]{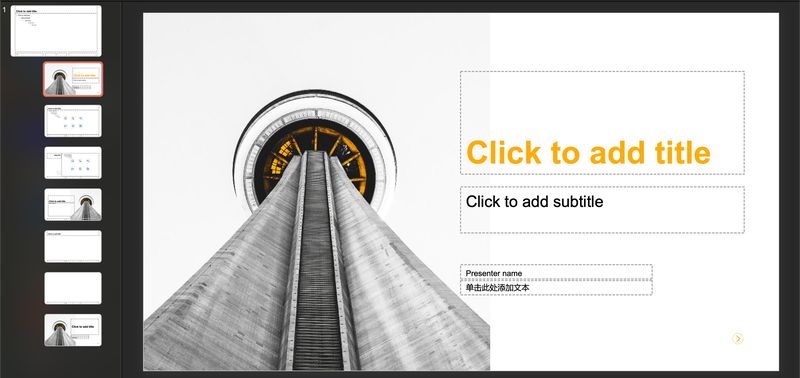}
        \label{L2-2}
    }
    \caption{Level 3 Examples }
    \label{fig:L1figures}
\end{figure}

\paragraph{Level 4: Parametric (The Enterprise Tool)}
\begin{itemize}
    \item  {Definition:} Visuals are driven by native data parameters.
    \item  {Technical Hallmark:}
    \begin{itemize}
        \item  {Native Data Binding:} Charts are instantiated as \texttt{\textless c:chart\textgreater} objects linked to an embedded Excel binary (\texttt{.xlsx}).
        \item  {SmartArt/Diagrams:} Process flows use semantic connectors, not just distinct lines.
    \end{itemize}
    \item  {Critical Failure Condition (Knockout):}
    \begin{itemize}
        \item \textbf{"Dead" Vector Charts:} A chart looks perfect but is constructed from static rectangles and text boxes. Right-clicking shows "Ungroup" instead of "Edit Data".
        \item \textbf{Broken Data Link:} The chart object exists, but the underlying Excel data is missing or corrupt.
    \end{itemize}
    
\end{itemize}

\begin{figure}[H]
    \centering
    % TODO: Replace with absolute path
    \includegraphics[width=0.4\textwidth]{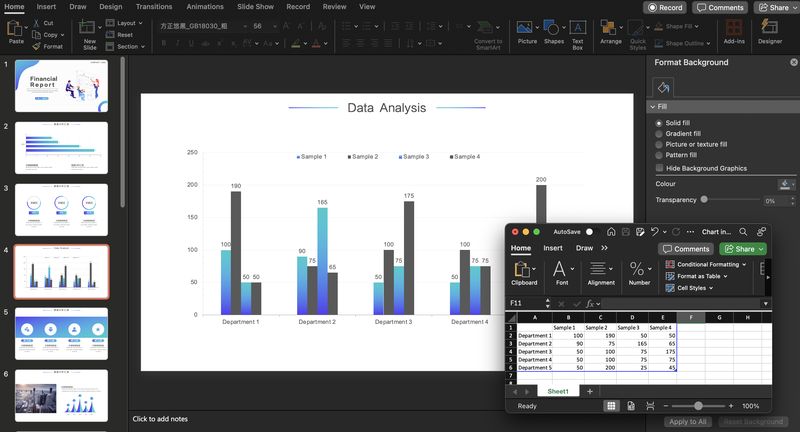}
    \caption{Level 4: Parametric (Native data in the chart)(Human-made)}
\end{figure}

\subsubsection{Phase 3: The Experience Phase (Time \& Narrative)}

\paragraph{Level 5: Cinematic (The Dynamic Experience)}
\begin{itemize}
    \item  {Definition:} The presentation functions as a directed, temporal narrative.
    \item  {Technical Hallmark:}
    \begin{itemize}
        \item  {Animation Logic:} Elements utilize Build-In/Build-Out effects sequences that match the reading order.
        \item  {Native Media:} Video/Audio is embedded in the DOM with playback controls.
    \end{itemize}
    \item  {Critical Failure Condition (Knockout):}
\begin{itemize}
    \item \textbf{Static State:} The presentation is functionally perfect (Data \& Structure are correct), but lacks time-dimension attributes (No animations, no transitions).
    \item \textbf{External Dependency:} Media files are linked to a local path on the creator's machine rather than embedded, causing playback failure.
\end{itemize}

\end{itemize}
\subsection{ Evaluation Protocols}
Select the protocol below matching your input type.
\begin{figure}[H]
    \centering
    % TODO: Replace with absolute path
    \includegraphics[width=0.45\textwidth]{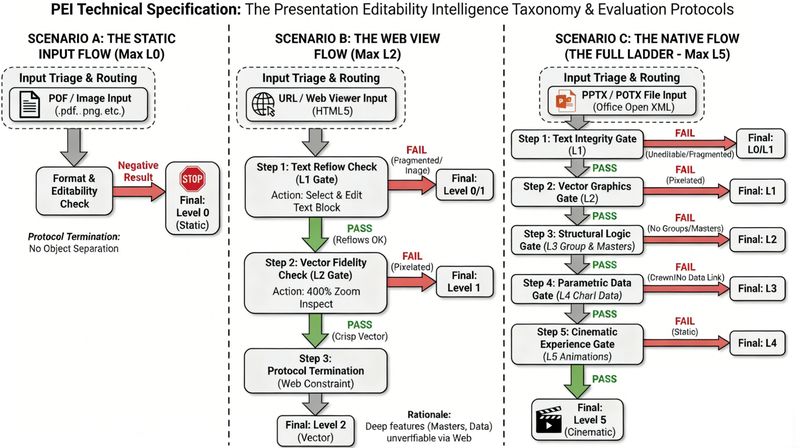}
    \caption{Evaluation Protocols Flow}
\end{figure}
\subsubsection{Protocol A: The Static Flow}
 {Input:} PDF / Image \\
 {Procedure:}
\begin{enumerate}
    \item  {Format Check:} Identify file extension (\texttt{.pdf}, \texttt{.png}, etc.).
    \item  {Editability Check:} Attempt to select text or move an object.
    \begin{itemize}
        \item \textit{Result:} Negative.
    \end{itemize}
    \item  {Final Classification:}  {Level 0 (Static).}
\end{enumerate}
\subsubsection{Protocol B: The Web Flow}
 {Input:} URL / Web Viewer \\
 {Constraint:} Max Rating = L2.
\paragraph{Step 1: Text Reflow Validation (L1 Check)}
\begin{itemize}
    \item \textit{Action:} Click on a text block in the web view.
    \item \textit{Check:} Is it selectable text? If you delete words, does the text box resize or reflow naturally?
    \item \textit{Decision:}
    \begin{itemize}
        \item If Text is Image or Unselectable:  {Classify as L0.}
        \item If Text is fragmented/does not reflow:  {Classify as L1.}
        \item If Text behaves correctly:  {Proceed to Step 2.}
    \end{itemize}
\end{itemize}
\paragraph{Step 2: Vector Fidelity Validation (L2 Check)}
\begin{itemize}
    \item \textit{Action:} Zoom browser to 400\%. Inspect icons and diagrams.
    \item \textit{Check:} Are edges crisp (Vector/SVG) or pixelated (Raster)?
    \item \textit{Decision:}
    \begin{itemize}
        \item If Pixelated:  {Classify as L1.}
        \item If Crisp/Vector:  {Classify as L2.}
    \end{itemize}
\end{itemize}
\paragraph{Step 3: Protocol Termination}
\begin{itemize}
    \item \textit{Reasoning:} Web views cannot reliably prove the existence of Master Slides or editable Excel data.
    \item \textit{Final Classification:}  {Level 2 (Vector).}
\end{itemize}
 
\subsubsection{Protocol C: The PPTx Flow}
 {Input:} PPTX File \\
 {Procedure:} Perform checks sequentially. Stop immediately upon failure.
\paragraph{Step 1: The Text Integrity Check (L1 Gate)}
\begin{itemize}
    \item \textit{Action:} Select a paragraph. Edit the text to double its length.
    \item \textit{Criteria:} The text must stay within its container and wrap automatically. The paragraph must be a single object, not multiple lines.
    \item \textit{Result:}
    \begin{itemize}
        \item  {Fail:} Content is uneditable (L0) or fragmented (L1).  {STOP.}
        \item  {Pass:}  {Proceed to Step 2.}
    \end{itemize}
\end{itemize}
\paragraph{Step 2: The Vector Graphics Check (L2 Gate)}
\begin{itemize}
    \item \textit{Action:} Zoom to 400\%. Inspect non-text elements (icons, shapes).
    \item \textit{Criteria:} Elements must be vector shapes (Shapes/SVG), not raster screenshots.
    \item \textit{Result:}
    \begin{itemize}
        \item  {Fail (Pixelated):}  {Downgrade to Level 1. STOP.}
        \item  {Pass:}  {Proceed to Step 3.}
    \end{itemize}
\end{itemize}
\paragraph{Step 3: The Structural Logic Check (L3 Gate)}
\begin{itemize}
    \item \textit{Action A (Grouping):} Click a complex icon. Does it move as one unit (Group) or scatter into pieces?
    \item \textit{Action B (Masters):} View $\rightarrow$ Slide Master. Add a distinct shape to the layout. Close Master View. Does the shape appear on the slides?
    \item \textit{Criteria:} Complex elements must be grouped; Layouts must inherit from Master.
    \item \textit{Result:}
    \begin{itemize}
        \item  {Fail:}  {Classify as Level 2. STOP.}
        \item  {Pass:}  {Proceed to Step 4.}
    \end{itemize}
\end{itemize}
\paragraph{Step 4: The Data Native Check (L4 Gate)}
\begin{itemize}
    \item \textit{Action:} Identify a chart. Right-click the chart area. Look for "Edit Data."
    \item \textit{Criteria:} The "Edit Data" option must exist and successfully open an embedded Excel sheet. Changing a value in Excel must instantly update the chart visual.
    \item \textit{Result:}
    \begin{itemize}
        \item  {Fail (No option/Broken link):}  {Classify as Level 3. STOP.}
        \item  {Pass:}  {Proceed to Step 5.}
    \end{itemize}
\end{itemize}
\paragraph{Step 5: The Cinematic Check (L5 Gate)}
\begin{itemize}
    \item \textit{Action:} Run "Slide Show" mode from the beginning.
    \item \textit{Criteria:} Slides must transition automatically or smoothly. Elements should animate in (Build-ins). Embedded video must play natively.
    \item \textit{Result:}
    \begin{itemize}
        \item  {Fail (Static Show):}  {Classify as Level 4. STOP.}
        \item  {Pass:}  {Classify as Level 5 (Cinematic).}
    \end{itemize}
\end{itemize}

Note: The protocols described in I.4 serve as the conceptual and human-facing operational criteria to illustrate the rationale behind each PEI level. In the actual SlidesGen-Bench pipeline, these checks are executed fully automatically.

\subsection{Automated Parsing Pipeline for PEI}
For native .pptx outputs, our evaluation script programmatically unzips the archive and scans the underlying Office Open XML tree. For instance:
Level 3 (Structural): The script checks for the existence of customized layout mapping in slideMaster.xml and validates <p:grpSp> (Group Shape) tags.
Level 4 (Parametric): The script parses /charts/chart*.xml to verify if <c:externalData> nodes are valid and linked to an intact embedded Excel binary (.xlsx).
Level 5 (Cinematic): The script inspects the <p:timing> node within the slide XML to confirm the presence of temporal animation tags (e.g., <p:bldLst>).
For Web/Image outputs, the triage script automatically limits the max attainable level to L2 based on file extensions and DOM node extraction (e.g., counting <svg> vs <canvas> tags).

\subsection{Disclaimer:} It is crucial to note that PEI is inherently \textit{format-conditional}. Because it measures secondary structural editing capacity within the traditional Office Open XML paradigm, systems outputting flattened images or web-based HTML canvases face a natural architectural ceiling (e.g., typically capped at Level 2). This matrix does not penalize image-centric models for visual quality, but rather objectively delineates their current limitations in enterprise-level structural workflows.

\section{PEI Parser-Derived Subcriteria}
\label{app:pei_subcriteria}

To support finer-grained comparison of systems that cluster at the same PEI level, we additionally extract parser-derived subcriteria directly from the generated file structures. Table~\ref{tab:pei_subcriteria} lists the six subcriteria and how each is checked automatically; the main PEI levels remain unchanged.

\begin{table*}[htbp]
\centering
\footnotesize
\caption{PEI parser-derived subcriteria.}
\label{tab:pei_subcriteria}
\begin{NiceTabular}{p{3.6cm}p{11.2cm}}
\toprule
\textbf{Subcriterion} & \textbf{Parser check} \\
\midrule
Text object editability & Text is stored as selectable characters rather than rasterized pixels (e.g., \texttt{<a:t>} nodes in PPTX, text nodes in HTML DOM). \\
Object separability & Elements are individually addressable objects (shape IDs / DOM nodes) rather than a single flattened canvas. \\
Grouping & Related elements are bound with \texttt{<p:grpSp>} or equivalent group constructs. \\
Native chart/table editability & Charts/tables are instantiated as native objects (\texttt{<c:chart>} with embedded data) rather than drawn vector approximations. \\
Narrative structure & A document-wide hierarchy exists (e.g., \texttt{<p:sldMaster>} inheritance and layout mapping). \\
Animation/multimedia support & Temporal attributes exist (\texttt{<p:timing>}, build lists) and media is embedded natively in the DOM. \\
\bottomrule
\end{NiceTabular}
\end{table*}

\section{Evaluation Setup for Open-Source Frameworks} \label{app:opensource_eval}

To ensure a fair and rigorous comparison between commercial platforms and open-source frameworks under our unified benchmark setting, we made necessary adaptations to the input pipelines of AutoPresent~\cite{ge2025autopresent} and PPT-Agent~\cite{zheng2025pptagent}. The detailed configurations are as follows:

\paragraph{Adaptations for AutoPresent.}
AutoPresent is designed to synthesize slides from scratch but does not natively support the direct ingestion of raw PDF documents or ultra-long texts. To circumvent this limitation, we introduced a text pre-processing step: we utilized an LLM to extract the core information from the source PDFs, summarize the content, and format it into structured bullet points that perfectly match AutoPresent's required input format. Furthermore, while AutoPresent defaults to DALL-E for its image generation module, we replaced this backend with Nano-Banana-Pro. This modification was made to ensure optimal visual synthesis performance and to align the model's visual capabilities with the state-of-the-art multi-modal standards evaluated in our benchmark.

\paragraph{Adaptations for PPT-Agent.}
Evaluating PPT-Agent presented a unique challenge, as its core methodology relies on extracting and modifying layouts from a user-provided reference PPT template, alongside the source document. Because our benchmark evaluates ``from-scratch'' slide generation, supplying a bespoke reference template for each test case would introduce an unfair advantage. To establish an equitable comparison, we simulated the template-routing mechanisms utilized by commercial systems (e.g., Gamma.com.ai and Kimi standard mode). Specifically, we curated a localized template library comprising 50 high-quality, human-designed PPTs encompassing a wide range of usage scenarios and visual themes. Each template was meticulously annotated with semantic tags. During inference, we deployed an LLM to analyze the input document and dynamically retrieve the most contextually appropriate template from this library. The selected template and the document were then seamlessly fed into PPT-Agent to render the final slides.

\section{System Snapshot and Versioning}
\label{app:snapshot}

Commercial slide-generation systems evolve quickly, so we present the reported results as a dated \method\ v1.0 snapshot: all commercial systems were accessed between November and December 2025, and the exact model snapshots are archived with the open-source release to support reruns under the same protocol. Table~\ref{tab:system_snapshot} summarizes the evaluated systems, their generation paradigm, and the access window.

\begin{table*}[htbp]
\centering
\footnotesize
\caption{Evaluated systems and access window.}
\label{tab:system_snapshot}
\begin{NiceTabular}{p{3.4cm}p{3.4cm}p{8.2cm}}
\toprule
\textbf{System} & \textbf{Paradigm} & \textbf{Access window / notes} \\
\midrule
Gamma.com.ai & Source file (template) & Nov--Dec 2025 \\
Quark PPT & Source file (template) & Nov--Dec 2025 \\
Kimi-PPT (Standard / Smart) & Source file (template) & Nov--Dec 2025 \\
Zhipu PPT & HTML code generation & Nov--Dec 2025 \\
Skywork & HTML code generation & Nov--Dec 2025 \\
NotebookLM & Image generation & Nov--Dec 2025 \\
Skywork-Banana Mode & Image generation & Nov--Dec 2025 \\
Kimi-Banana Mode & Image generation & Nov--Dec 2025 \\
AutoPresent* & Source file (code) & Academic framework; adapted pipeline \\
PPTAgent* & Source file (template) & Academic framework; adapted pipeline \\
\bottomrule
\end{NiceTabular}
\end{table*}

The released dataset, evaluation code, prompts, scoring scripts, and generated artifacts support independent verification and versioned benchmark snapshots as systems change. The fixed v1.0 results serve as a reproducible reference point, while the open-source release provides the mechanism for continued updates.

\section{Aesthetics results for different purpose}\label{app: visual}
We report the aesthetics results breakdown by purpose in Table \ref{tab:domain_breakdown}. The results reveal distinct visual signatures for different presentation types. Product Launch slides demonstrate exceptional performance, achieving the highest Usability (5.53), Engagement (7.75), and Harmony (-1.01), reflecting a comprehensive design priority on capturing audience attention, ensuring clarity, and maintaining visual balance for marketing impact. Meanwhile, Work Report slides exhibit the highest Rhythm (11.31), suggesting that these domains often utilize structured, repeating visual patterns to effectively convey professional updates and data.

\begin{table*}[t]
    \small
    \centering
    % Adjusted spacing for a cleaner look
    \setlength{\tabcolsep}{10.0pt}
    \begin{NiceTabular}{l|c|cccc}
         \toprule
         {\bf Purpose} & {\bf Count} & {\bf Usability} & {\bf Engagement} & {\bf Harmony} & {\bf Rhythm} \\
         \midrule
         Brand Promote & 94 & 5.17 & 7.39 & -1.08 & 11.03 \\
         Business Plan & 56 & 5.08 & 6.61 & -1.24 & 10.86 \\
         Personal Statement & 160 & 5.26 & 6.94 & -1.35 & 10.13 \\
         Product Launch & 150 & \textbf{5.53} & \textbf{7.75} & \textbf{-1.01} & 10.53 \\
         Topic Introduction & 789 & 4.74 & 7.22 & -1.28 & 10.13 \\
         Work Report & 97 & 5.32 & 7.38 & -1.14 & \textbf{11.31} \\
         \bottomrule
    \end{NiceTabular}
    \caption{Aesthetics results breakdown across different presentation domains. The table reports the number of samples (Count) and average aesthetic scores for each category. Best values for each metric are highlighted in \textbf{bold}.}
    \label{tab:domain_breakdown}
\end{table*}

\section{Details of Human Annotation} \label{app:annotator_details}

To rigorously construct the Slides-Align1.5k dataset and evaluate human alignment, we implemented a strict annotation protocol. The details regarding annotator selection, training, and agreement are outlined below.

\paragraph{Annotator Qualifications and Diversity.}
Our annotator pool was carefully curated to represent a balanced mix of technical, aesthetic, and professional perspectives, ensuring that the ground truth reflects a holistic view of presentation quality. The group consisted of:
\begin{itemize}[leftmargin=5mm, itemsep=0mm]
    \item \textbf{Ph.D. students in Computer Science}, who excel at evaluating the structural logic, content fidelity, and information hierarchy of the slides.
    \item \textbf{Undergraduate students from the College of Art}, who provided expert judgment on visual harmony, typography layout, and color palettes.
    \item \textbf{Office clerks}, representing the ``end-user'' perspective to assess the practical utility, readability, and professional suitability of the generated decks in real-world scenarios.
\end{itemize}

\paragraph{Standardized Training and Requirements.}
To minimize subjective variance across diverse backgrounds, all annotators were required to participate in unified training sessions before starting the ranking tasks. During these sessions, we provided explicit definitions of the four targeted aesthetic dimensions: Harmony, Engagement, Usability, and Visual Rhythm. We systematically showcased representative ``good'' and ``bad'' patterns for each dimension and established clear operational guidelines for the ranking interface. A set of canonical examples was also provided as reference. This standardization ensured that all annotators shared a common, rigorous understanding of what constitutes a ``professional'' presentation.

\paragraph{Inter-annotator Agreement and Compensation.}
The reliability of human judgments is demonstrated by the inter-annotator agreement results (see the ``Humans'' entry in Table~\ref{tab:SlidesGen_results}). The annotation consensus achieved an Average Spearman correlation of 0.85 with a remarkably low Standard Deviation of 0.12. This high correlation verifies that our protocol successfully elicited stable and consistent preferences, serving as a robust ``gold standard'' for calibrating our computational metrics. Furthermore, all participants were fairly compensated according to the local institutional standards for professional data labeling.

\begin{figure*}[htp]
    \centering
    \includegraphics[width=\linewidth]{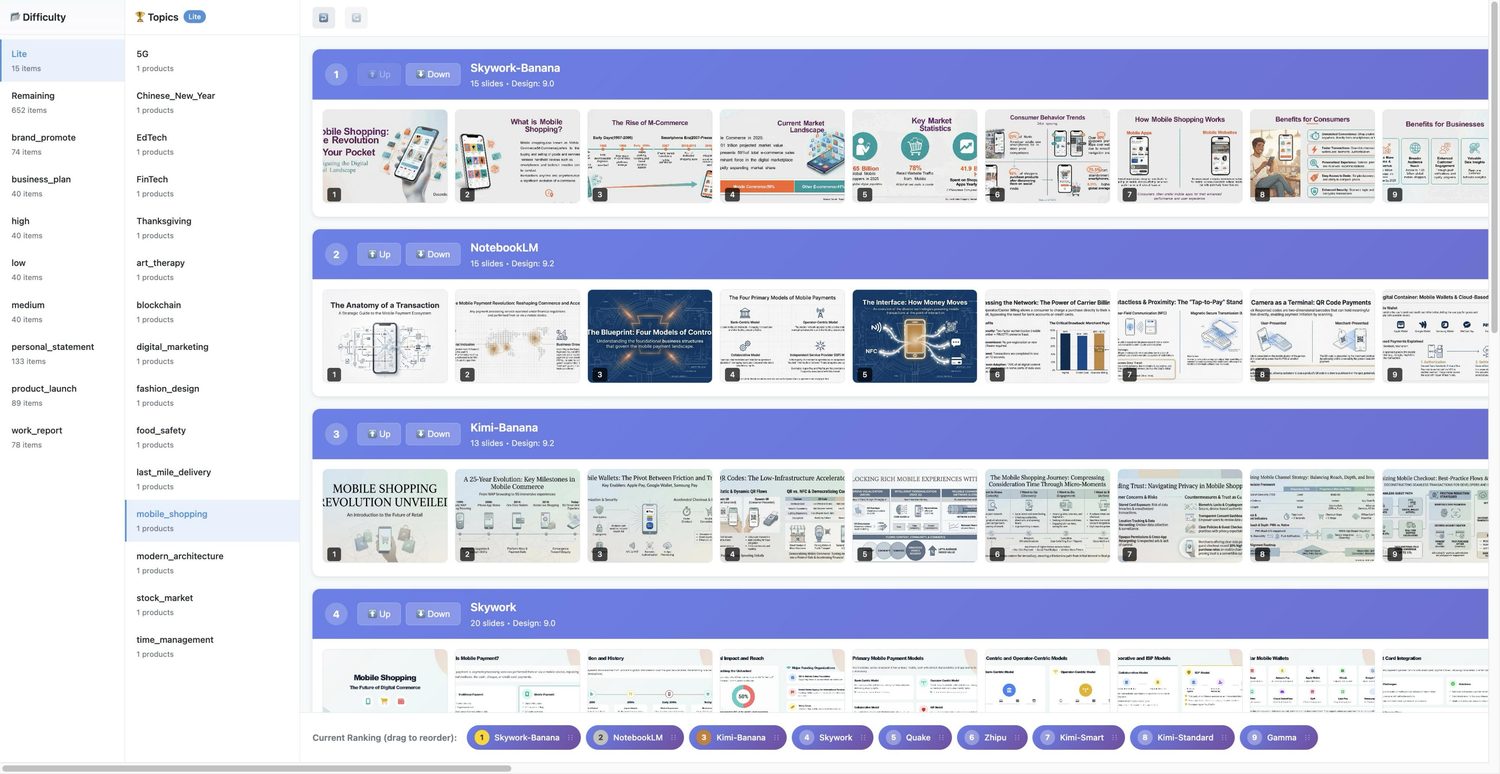}
    \caption{The web-based annotation interface for ranking.}
    \label{fig:rank}
\end{figure*}

\paragraph{Annotation Platforms and Process.}
To obtain reliable human judgments on slide aesthetics, we developed a web-based annotation interface for ranking~\ref{fig:rank}.Given the same input prompt and source documents, multiple AI systems generated slide presentations. 

Human annotators were presented with high-resolution rendered slide images from the competing systems simultaneously. Rather than assigning absolute scores—which are highly susceptible to subjective calibration drift—we adopted a \textit{relative ranking granularity}. For each prompt-matched set, annotators were required to rank the generated decks from best to worst. 

\section{QuizBank Error Analysis Details}
\label{app:quizbank_details}

\subsection{Error Taxonomy and Statistics}
We conducted a human audit on all 2499 instances. Human annotators directly inspected the rendered slide images against the ground truth. We categorize errors into four types. Table~\ref{tab:error_distribution} presents the overall distribution, showing that missing information is the most prevalent issue.

Decoupling Generation Errors from Extraction Bottlenecks: A natural concern is whether "Missing Content" errors stem from the generation model's omission or the extraction pipeline's VLM failure to parse dense layouts. Table~\ref{tab:error_distribution} shows that only 1.2\% of the errors were falsely penalized due to VLM extraction failures on extremely dense or overlapping layouts. This confirms that our automated QuizBank evaluation faithfully reflects the generation models' true capabilities with a highly acceptable error margin.

\begin{table*}[htp]
    \centering
    \begin{tabular}{lrrp{3cm}}
        \toprule
        \textbf{Error Type} & \textbf{Count} & \textbf{Perc.} & \textbf{Definition} \\
        \midrule
        Type 1: Missing Content & 1658 & 66.3\% & Slide omits required facts/figures. \\
        Type 2: Value Mismatch & 791 & 31.7\% & Content exists but values differ from GT. \\
        Type 3: VLM Extract Failure & 31 & 1.2\% & VLM fails to extract visible info. \\
        Type 4: Other & 19 & 0.8\% & Unclassified/Formatting issues. \\
        \bottomrule
    \end{tabular}
    \caption{\textbf{Distribution of Error Types.} Based on 2499 incorrect answers.}
    \label{tab:error_distribution}
\end{table*}

\subsection{Model-Specific Breakdown}
Table~\ref{tab:model_breakdown} details error distributions by product. We observe that data-heavy topics (e.g., \textit{internet\_of\_things}, \textit{cryptocurrency}) suffer the highest error rates due to the demand for precise numerical reasoning.

\begin{table*}[htp]
    \centering
    \begin{tabular}{lcccc}
    \toprule
    \textbf{Product} & \textbf{Total Errors} & \textbf{Missing Content} & \textbf{Value Mismatch} & \textbf{Other} \\
    \midrule
    Gamma & 490 & 66\% & 32\% & 2\% \\
    NotebookLM & 363 & 58\% & 41\% & 1\% \\
    Skywork & 310 & 61\% & 36\% & 3\% \\
    Quark & 278 & 70\% & 28\% & 2\% \\
    Skywork-Banana & 262 & 67\% & 31\% & 2\% \\
    Kimi-Standard & 222 & 69\% & 29\% & 2\% \\
    Kimi-Smart & 197 & 63\% & 35\% & 2\% \\
    Kimi-Banana & 191 & 44\% & 52\% & 4\% \\
    Zhipu & 186 & 49\% & 49\% & 2\% \\
    \bottomrule
\end{tabular}
    \caption{\textbf{Error Distribution by Product.} Sorted by total error count. Top-performing models (bottom rows) show a shift from "Missing Content" to "Value Mismatch."}
    \label{tab:model_breakdown}
\end{table*}

\subsection{Qualitative Examples}
Table~\ref{tab:qualitative_errors} provides real-world examples of the three primary error categories.

\begin{table*}[htp]
    \centering
    \small
    \begin{tabular}{p{0.15\linewidth} p{0.15\linewidth} p{0.6\linewidth}}
        \toprule
        \textbf{Type} & \textbf{Topic} & \textbf{Description \& Root Cause} \\
        \midrule
        \textbf{Missing Content} & 5G Deployment & \textbf{Q:} Which country achieved the first full commercial deployment? \newline \textbf{Issue:} The slide discussed 5G concepts but omitted country-specific milestones. \\
        \midrule
        \textbf{Value Mismatch} & Digital Marketing & \textbf{Q:} Cost per lead comparison? \newline \textbf{Issue:} Slide stated "44\% less expensive," but Ground Truth expected a different statistic. \\
        \midrule
        \textbf{VLM Failure} & Film Industry & \textbf{Q:} Nollywood's global ranking? \newline \textbf{Issue:} Slide text contained "fourth," but VLM failed to extract it due to dense layout. \\
        \bottomrule
    \end{tabular}
    \caption{\textbf{Qualitative Examples of Common Errors.}}
    \label{tab:qualitative_errors}
\end{table*}

\section{QuizBank Evaluation Robustness}
\label{app:quizbank_robust}

\paragraph{Oracle-Context QA.}
To isolate VLM answering noise from generated-slide content loss, we ran an oracle-context QA sanity check: the evaluator answers the same QuizBank MCQs from ground-truth source evidence instead of the slide-derived context. Across all questions, oracle-context QA achieves \textbf{100\%} accuracy, showing that the questions are answerable from their source documents and that the VLM answer stage succeeds whenever the required evidence is present.

\paragraph{Backend Sensitivity.}
We ran an all-question backend-sensitivity comparison between DeepSeek-v4-Pro and MiniMax-M2.7 under otherwise identical settings (Table~\ref{tab:backend_sensitivity}).

\begin{table}[htb]
\centering
\footnotesize
\caption{Backend-sensitivity comparison on the full QuizBank question set.}
\label{tab:backend_sensitivity}
\begin{tabular}{lcc}
\toprule
\textbf{Backend} & \textbf{Accuracy} & \textbf{Questions answered} \\
\midrule
DeepSeek-v4-Pro & 60.8\% & All questions \\
MiniMax-M2.7 & 61.1\% & All questions \\
\bottomrule
\end{tabular}
\end{table}

The nearly identical accuracies and symmetric disagreement cases suggest no clear systematic backend bias in the all-question comparison. The $\sim$60\% slide-context accuracy reflects generated-slide content fidelity; under oracle source evidence across all questions, the same QA stage reaches 100\%.

\paragraph{Bootstrap Confidence Intervals.}
To quantify the stability of the main quantitative results, we computed 95\% bootstrap confidence intervals by resampling the evaluated decks with replacement (1,000 iterations) and taking the 2.5th and 97.5th percentiles. Tables~\ref{tab:quizbank_ci} and \ref{tab:total_ci} report the per-system QuizBank accuracy and the composite total score produced by our scoring pipeline, together with the evaluated $n$ (number of decks) per system. The intervals are tight, typically within $\pm$2--5\%, and the top-ranked system's lower confidence bound remains at or above the runner-up's upper bound, indicating that the main benchmark conclusions are reasonably stable. For example, Kimi-Banana's QuizBank lower confidence bound (85.4\%) is at the same level as Skywork-Banana's upper bound, and the composite total-score ranking is clearly separated at the top: Kimi-Banana's interval [83.1, 84.0] lies above Skywork's [80.6, 81.9].

\begin{table*}[htb]
\centering
\small
\caption{QuizBank accuracy by product with 95\% bootstrap confidence intervals and evaluated $n$.}
\label{tab:quizbank_ci}
\begin{NiceTabular}{lccc}
\toprule
\textbf{Product} & \textbf{Mean Accuracy} & \textbf{95\% CI} & $n$ \\
\midrule
Kimi-Banana & 87.1\% & [85.4\%, 88.7\%] & 175 \\
Zhipu & 86.8\% & [83.8\%, 89.5\%] & 180 \\
Skywork-Banana & 82.9\% & [80.2\%, 85.4\%] & 182 \\
Kimi-Smart & 82.1\% & [79.1\%, 84.8\%] & 116 \\
Skywork & 81.3\% & [78.8\%, 83.7\%] & 186 \\
Quark & 80.9\% & [77.5\%, 83.9\%] & 158 \\
Kimi-Standard & 79.9\% & [76.4\%, 83.1\%] & 118 \\
Gamma & 69.9\% & [66.3\%, 73.5\%] & 183 \\
NotebookLM & 69.0\% & [62.6\%, 75.2\%] & 137 \\
\bottomrule
\end{NiceTabular}
\end{table*}

\begin{table*}[htb]
\centering
\small
\caption{Composite total score by product with 95\% bootstrap confidence intervals.}
\label{tab:total_ci}
\begin{NiceTabular}{lcc}
\toprule
\textbf{Product} & \textbf{Score} & \textbf{95\% CI} \\
\midrule
Kimi-Banana & 83.5 & [83.1, 84.0] \\
Skywork & 81.2 & [80.6, 81.9] \\
Skywork-Banana & 79.2 & [78.4, 80.0] \\
Zhipu & 78.4 & [77.0, 79.6] \\
Kimi-Smart & 70.0 & [69.3, 70.7] \\
Quark & 68.9 & [67.9, 69.7] \\
NotebookLM & 67.6 & [63.8, 71.4] \\
Kimi-Standard & 67.6 & [66.9, 68.2] \\
Gamma & 65.1 & [63.9, 66.2] \\
\bottomrule
\end{NiceTabular}
\end{table*}

\section{QuizBank Coverage and Bias Audit}
\label{app:quizbank_audit}

To characterize the coverage and potential generation bias of the QuizBank, we conducted a full-bank audit over all available MCQs. Table~\ref{tab:quizbank_audit} summarizes the main results.

\begin{table*}[htb]
\centering
\small
\caption{Full-bank QuizBank coverage and bias audit.}
\label{tab:quizbank_audit}
\begin{NiceTabular}{p{3.8cm}p{5.8cm}p{5.4cm}}
\toprule
\textbf{Audit item} & \textbf{Result} & \textbf{Interpretation} \\
\midrule
Available MCQs & 1,879 over 188 document records & Full-bank audit rather than a small manual-only subset \\
Declared type balance & Data 49.7\%, Concept 50.3\% & The intended 5-data/5-concept design is preserved \\
Data share by domain & 47.0\%--57.7\% & No single domain is dominated by one declared type \\
Data share by difficulty & Low 47.7\%, Medium 50.0\%, High 51.3\% & Type balance is stable across difficulty levels \\
Evidence anchors before sampling & 14.24 per document on average & The final 10-question probe is sampled from a broader verified pool \\
Question-demand tendency & Numeric/date/statistic 70.3\%; entity/framework 9.2\%; relation/comparison 7.5\%; procedure/action 5.1\%; definition/concept 3.4\% & QuizBank primarily measures source-grounded factual retention \\
Answer-position prior (after fixed-seed shuffling) & A 25.3\%, B 25.0\%, C 24.3\%, D 25.4\% & The answer-position prior is removed without changing question semantics \\
Wrong selected labels in stored evaluator outputs & A 36.7\%, B 15.7\%, C 14.5\%, D 33.2\% & Observed errors are not driven by exploiting the original B/C prior \\
\bottomrule
\end{NiceTabular}
\end{table*}

The audit clarifies both the strength and the boundary of QuizBank. It is strongest for checking whether generated slides preserve concrete facts, quantities, entities, and relations from the source document. It is less direct for long-form narrative synthesis, rhetorical framing, and argument coherence, which we treat as complementary dimensions rather than claims made by the QuizBank score alone. The released QuizBank uses the shuffled option order.

\section{QuizBank Construction}\label{app: quiz_construct}

We adopt a three-step LLM-based pipeline for quiz generation: (1) extracting quantitative and qualitative anchor points with verbatim citations (Figure~\ref{fig:step1_extraction_prompt}), (2) verifying and refining extracted points against the source document (Figure~\ref{fig:step2_verification_prompt}), and (3) generating multiple-choice questions with source-grounded explanations (Figure~\ref{fig:step3_quiz_generation_prompt}). This multi-stage approach ensures high factual accuracy and traceability in the generated quiz bank.

\begin{figure*}[htbp]
\begin{tcolorbox}[colback=blue!5!white,colframe=blue!75!black]
\begin{small}
\textbf{Step 1: Source of Truth Extraction Prompt:}

System Role: Forensic Analyst.

\vspace{2mm}

Context:
\begin{itemize}[noitemsep,topsep=0pt]
    \item Domain: \{domain\} (\{focus\})
    \item Purpose: \{one\_sentence\}
    \item Text: \{document\_text\}
\end{itemize}

\vspace{2mm}

Task: Extract "Source of Truth" points.

CRITICAL: Do not use keywords. Write "Detailed Contextual Statements".

\vspace{2mm}

\textbf{Instructions:}
\begin{enumerate}[noitemsep,topsep=0pt]
    \item \textbf{Quantitative (Data):} Find 4-6 hard data points (Numbers, Dates, Specs).
    \begin{itemize}[noitemsep,topsep=0pt]
        \item \textit{Bad:} "Revenue grew."
        \item \textit{Good:} "Q3 Revenue grew by 25\% YoY to \$10M."
    \end{itemize}
    \item \textbf{Qualitative (Concepts):} Find 6-8 core concepts.
    \begin{itemize}[noitemsep,topsep=0pt]
        \item \textit{Bad:} "AI Strategy."
        \item \textit{Good:} "Pivoting to 'AI-First' to reduce costs by 40\%."
    \end{itemize}
    \item \textbf{Citation:} Quote the \textit{original verbatim text} and Cite the Page Number using "[=== PAGE X START ===]".
\end{enumerate}

\vspace{2mm}

\textbf{Output JSON:}
\begin{verbatim}
{
  "quantitative_anchors": [{"statement": "...", 
    "source_quote": "...", "location": "Page X"}],
  "qualitative_key_points": [{"statement": "...", 
    "source_quote": "...", "location": "Page X"}]
}
\end{verbatim}

\end{small}
\end{tcolorbox}
\caption{Step 1: Source of Truth Extraction Prompt for extracting quantitative and qualitative anchor points from source documents with verbatim citations.}
\label{fig:step1_extraction_prompt}
\end{figure*}

\begin{figure*}[htbp]
\begin{tcolorbox}[colback=green!5!white,colframe=green!75!black]
\begin{small}
\textbf{Step 2: Verification and Refinement Prompt:}

System Role: Strict Editor.

\vspace{2mm}

Context:
\begin{itemize}[noitemsep,topsep=0pt]
    \item Text: \{document\_text\}
    \item Draft JSON: \{draft\_json\}
\end{itemize}

\vspace{2mm}

\textbf{Task: Audit, Refine, and Expand.}
\begin{enumerate}[noitemsep,topsep=0pt]
    \item \textbf{Verify:} Check 'source\_quote' exists in Text.
    \item \textbf{Expand:} If a massive point is missing, ADD IT.
    \item \textbf{Filter:} Remove trivial points.
\end{enumerate}

\vspace{2mm}

Output: Return Polished JSON.

\end{small}
\end{tcolorbox}
\caption{Step 2: Verification Prompt for auditing extracted anchor points against source documents, expanding coverage, and filtering trivial information.}
\label{fig:step2_verification_prompt}
\end{figure*}

\begin{figure*}[htbp]
\begin{tcolorbox}[colback=orange!5!white,colframe=orange!75!black]
\begin{small}
\textbf{Step 3: Quiz Generation Prompt:}

System Role: Professional Exam Setter.

\vspace{2mm}

Input Context:
\begin{itemize}[noitemsep,topsep=0pt]
    \item Verified Data: \{quantitative\_anchors\}
    \item Verified Concepts: \{qualitative\_key\_points\}
\end{itemize}

\vspace{2mm}

Task: Generate exactly \{target\_count\} Multiple Choice Questions (MCQs).

\vspace{2mm}

\textbf{STRICT FORMATTING RULES:}
\begin{enumerate}[noitemsep,topsep=0pt]
    \item \textbf{Options:} Must be a list of 4 strings, explicitly starting with "A. ", "B. ", "C. ", "D. ".
    \item \textbf{Correct Answer:} Must be a SINGLE LETTER ("A", "B", "C", or "D"). Do not write the full text.
    \item \textbf{Explanation:} Must quote the source page.
\end{enumerate}

\vspace{2mm}

\textbf{EXAMPLE JSON OUTPUT (Follow this format exactly):}
\begin{verbatim}
{
  "quiz_bank": [{
    "id": 1,
    "type": "Data",
    "question": "What is the active duration 
                 for the 2025 campaign?",
    "options": [
      "A. January 1 through January 31, 2025",
      "B. January 6 through February 10, 2025",
      "C. February 1 through February 28, 2025",
      "D. March 1 through March 30, 2025"
    ],
    "correct_answer": "B",
    "explanation": "Based on Page 3: 'From January 6 
                    through February 10, 2025...'"
  }]
}
\end{verbatim}

\end{small}
\end{tcolorbox}
\caption{Step 3: Quiz Generation Prompt for creating multiple-choice questions from verified anchor points with strict formatting constraints and one-shot example.}
\label{fig:step3_quiz_generation_prompt}
\end{figure*}

\section{QuizBank Evaluation Prompt}\label{app: quiz_eval_prompt}

Figure~\ref{fig:single_slide_extraction_prompt} shows the prompt we used to extract single slide contents for content evaluation. We extract the exact texts and the important information in the charts or images. Figure~\ref{fig:quiz_evaluation_prompt} shows the prompt to do the "open-book" QuizBank test.

% -----------------------------------------------------------------------------
% 2. SINGLE SLIDE EXTRACTION PROMPT
% -----------------------------------------------------------------------------
\begin{figure*}[htbp]
\begin{tcolorbox}[colback=blue!5!white,colframe=blue!75!black]
\begin{small}
\textbf{Single Slide Extraction Prompt:}

Analyze the attached slide image for the specific purpose of a Content Quality \& Accuracy Audit. I need to compare this slide against source documentation, so precision is paramount.

\vspace{2mm}

Please extract the content into the following structured concise Markdown file:

\vspace{2mm}

\textbf{1. Textual Claims (Verbatim):}
\begin{itemize}[noitemsep,topsep=0pt]
    \item \textbf{Headlines:} Extract the exact Title and Subtitle.
    \item \textbf{Core Statements:} List every distinct claim or bullet point found in the body text exactly as written. Do not summarize.
    \item \textbf{Callouts:} Extract text from any bubbles, arrows, or highlight boxes.
\end{itemize}

\vspace{2mm}

\textbf{2. Quantitative Data Extraction:}
\begin{itemize}[noitemsep,topsep=0pt]
    \item \textbf{Chart/Table Data:} For every chart or table, list the specific data points visible. (e.g., `Q1 Revenue: \$10m', `Year-over-Year growth: 15\%').
    \item \textbf{In-Text metrics:} List any standalone numbers found in the text (e.g., `300+ employees', `50\% reduction').
\end{itemize}

\vspace{2mm}

\textbf{3. Visual Interpretation:}

Describe important images or icons that contain information and explain if they convey a specific sentiment or data point (e.g., `A green up-arrow indicating positive trend'). Ignore all decorative elements or irrelevant information.

\vspace{2mm}

Return the extracted content in a structured concise Markdown file.

\end{small}
\end{tcolorbox}
\caption{Single Slide Extraction Prompt for content extraction from individual slides. This prompt is used to extract verbatim textual claims, quantitative data, and visual interpretations for downstream content evaluation.}
\label{fig:single_slide_extraction_prompt}
\end{figure*}

% -----------------------------------------------------------------------------
% 4. QUIZ EVALUATION PROMPT
% -----------------------------------------------------------------------------
\begin{figure*}[htbp]
\begin{tcolorbox}[colback=blue!5!white,colframe=blue!75!black]
\begin{small}
\textbf{Quiz Evaluation Prompt:}

You are an expert quiz evaluator. Answer the following multiple-choice questions strictly based on the provided slide context and cite the source.

\vspace{2mm}

\textbf{Presentation Topic:} \{topic\}

\textbf{Extracted Slide Contents:} \{slide\_contents\}

\textbf{Quiz Questions:} \{quiz\_questions\}

\vspace{2mm}

\textbf{INSTRUCTIONS}

\vspace{1mm}

1. Read all extracted slide contents carefully.

2. For each question, select the best answer based ONLY on what's presented in the slides.

3. If the information is not covered in the slides, make your best inference or select ``insufficient information''.

4. Provide brief reasoning for each answer.

\vspace{2mm}

\textbf{OUTPUT FORMAT (JSON)}

\begin{verbatim}
{
    "answers": [
        {
            "question_id": <number>,
            "selected_answer": "<A|B|C|D>",
            "reasoning": "<brief explanation>"
        },
        ...
    ]
}
\end{verbatim}

\end{small}
\end{tcolorbox}
\caption{Quiz Evaluation Prompt for assessing content coverage through multiple-choice questions. The \{topic\}, \{slide\_contents\}, and \{quiz\_questions\} are replaced with the corresponding presentation data and generated quiz.}
\label{fig:quiz_evaluation_prompt}
\end{figure*}

\section{Prompt For LLM-as-Judge Aesthetics Evaluation}\label{app: Judge}
While we follow the standard nomenclature of "LLM-as-a-Judge", our implementation is fully multimodal. To establish a rigorous baseline, we utilized the gemini-3-flash-preview API. Rather than relying on textual intermediate representations (e.g., Markdown or code), we fed the actual rendered slide images (in high resolution) directly to the model alongside the evaluation prompts.
For the LLM-as-Judge Rating method, the model receives all slide images of a presentation and scores them based on the definitions of Harmony, Engagement, Usability, and Rhythm. For the LLM-as-Judge Arena, it takes two sets of slide images (Model A vs. Model B) and predicts human preference.

Figure~\ref{fig:arena_comparison_prompt} shows the head-to-head comparison prompt between slides. The criteria is the same as the proposed aesthetics metrics.
Figure~\ref{fig:visual_evaluation_prompt} shows the VLM rating prompt. The criteria is the same as the proposed aesthetics metrics.

% -----------------------------------------------------------------------------
% 1. ARENA COMPARISON PROMPT
% -----------------------------------------------------------------------------
\begin{figure*}[htbp]
\begin{tcolorbox}[colback=blue!5!white,colframe=blue!75!black]
\begin{small}
\textbf{Arena Comparison Prompt:}

You are an expert presentation judge. Compare two presentations (PPT A and PPT B) on the same topic based on their slide images.

\vspace{2mm}

\textbf{Topic:} \{topic\}

\textbf{PPT A:} \{num\_slides\_a\} slides \quad \textbf{PPT B:} \{num\_slides\_b\} slides

\vspace{2mm}

\textbf{CRITERIA}

\vspace{1mm}

1. \textbf{VISUAL DESIGN}: Color scheme, typography, consistency, image quality, theme

2. \textbf{LAYOUT}: Spatial balance, alignment, no overlapping, professional structure

\vspace{2mm}

\textbf{OUTPUT (JSON only)}

\vspace{1mm}

\begin{verbatim}
{
    "Visual_Design": {
        "winner": "A"|"B"|"Tie", 
        "score_difference": 1-5, 
        "reason": "<brief reason>"
    },
    "Layout": {
        "winner": "A"|"B"|"Tie", 
        "score_difference": 1-5, 
        "reason": "<brief reason>"
    },
    "Overall_Winner": "A"|"B"|"Tie",
    "Overall_Reason": "<brief overall comparison>",
    "Confidence": 1-5
}
\end{verbatim}

\end{small}
\end{tcolorbox}
\caption{Arena Comparison Prompt for head-to-head evaluation of two presentations. The \{topic\}, \{num\_slides\_a\}, and \{num\_slides\_b\} are replaced with the corresponding presentation metadata.}
\label{fig:arena_comparison_prompt}
\end{figure*}

% -----------------------------------------------------------------------------
% 5. VISUAL EVALUATION PROMPT
% -----------------------------------------------------------------------------
\begin{figure*}[htbp]
\begin{tcolorbox}[colback=blue!5!white,colframe=blue!75!black]
\begin{small}
\textbf{Visual Evaluation Prompt:}

You are an expert presentation designer evaluating the VISUAL DESIGN, LAYOUT, and COMPLEXITY of a PowerPoint presentation. Evaluate based ONLY on the provided slide images.

\vspace{2mm}

\textbf{Presentation Topic:} \{topic\} \quad \textbf{Number of Slides:} \{num\_slides\}

\vspace{2mm}

\textbf{VISUAL DESIGN CRITERIA (Weight: 40\%)}
\begin{itemize}[noitemsep,topsep=0pt,leftmargin=*]
    \item \textbf{Color\_Scheme} (20\%): Harmonic balance, contrast ratios, unified aesthetic.
    \item \textbf{Typography} (20\%): Readable fonts, consistent sizes, clear hierarchy.
    \item \textbf{Visual\_Consistency} (20\%): Color coherence, recurring motifs, layout stability.
    \item \textbf{Image\_Quality} (20\%): High quality, relevant, properly integrated images.
    \item \textbf{Theme\_Appropriateness} (20\%): Visual theme matches content and audience.
\end{itemize}

\vspace{1mm}

\textbf{LAYOUT CRITERIA (Weight: 40\%)}
\begin{itemize}[noitemsep,topsep=0pt,leftmargin=*]
    \item \textbf{Spatial\_Balance} (40\%): Effective whitespace, balanced elements.
    \item \textbf{Element\_Alignment} (30\%): Proper alignment of text, images, elements.
    \item \textbf{No\_Overlapping} (30\%): No obscured or cut-off elements.
\end{itemize}

\vspace{1mm}

\textbf{COMPLEXITY CRITERIA (Weight: 20\%)}
\begin{itemize}[noitemsep,topsep=0pt,leftmargin=*]
    \item \textbf{Charts\_and\_Data} (25\%): Charts, graphs, data visualizations where appropriate.
    \item \textbf{Visual\_Elements} (25\%): Icons, illustrations, diagrams, infographics.
    \item \textbf{Advanced\_Design} (25\%): Gradients, shadows, animations, depth effects.
    \item \textbf{Layout\_Variety} (25\%): Varied layouts appropriate for content types.
\end{itemize}

\vspace{2mm}

\textbf{INSTRUCTIONS}

1. Examine ALL slide images carefully in sequence.

2. Assign an EXACT INTEGER score from 0-10 for each sub-criterion (no decimals).

3. Provide constructive feedback on strengths and areas for improvement.

\vspace{2mm}

\textbf{OUTPUT FORMAT (JSON)}

\begin{verbatim}
{
    "Visual_Design": {
        "sub_scores": {
            "Color_Scheme": <0-10>, "Typography": <0-10>,
            "Visual_Consistency": <0-10>, "Image_Quality": <0-10>,
            "Theme_Appropriateness": <0-10>
        },
        "reason": "<detailed reasoning>"
    },
    "Layout": {
        "sub_scores": {
            "Spatial_Balance": <0-10>, "Element_Alignment": <0-10>,
            "No_Overlapping": <0-10>
        },
        "reason": "<detailed reasoning>"
    },
    "Complexity": {
        "sub_scores": {
            "Charts_and_Data": <0-10>, "Visual_Elements": <0-10>,
            "Advanced_Design": <0-10>, "Layout_Variety": <0-10>
        },
        "reason": "<detailed reasoning>"
    },
    "Overall_Feedback": "<brief summary>",
    "Top_Strengths": ["<strength 1>", "<strength 2>"],
    "Areas_for_Improvement": ["<area 1>", "<area 2>"]
}
\end{verbatim}

\end{small}
\end{tcolorbox}
\caption{Visual Evaluation Prompt for assessing presentation aesthetics. The \{topic\} and \{num\_slides\} are replaced with the corresponding presentation metadata. Each sub-criterion is scored on a 0-10 integer scale.}
\label{fig:visual_evaluation_prompt}
\end{figure*}

\section{Generated Slides Examples}

\begin{figure*}[htbp]
    \centering
    
    % Gamma
    \begin{minipage}{\textwidth}
    \centering
    \includegraphics[width=0.32\textwidth]{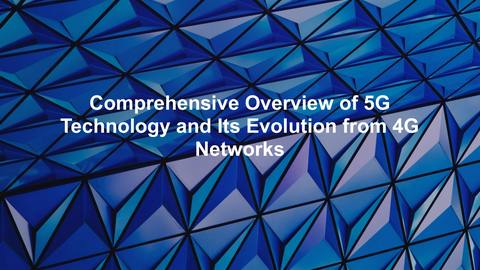}
    \includegraphics[width=0.32\textwidth]{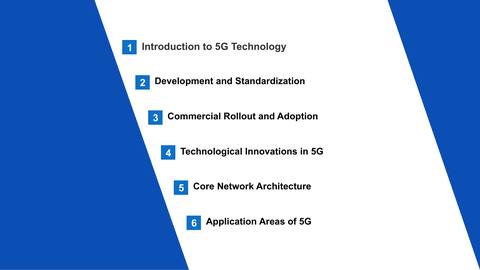}
    \includegraphics[width=0.32\textwidth]{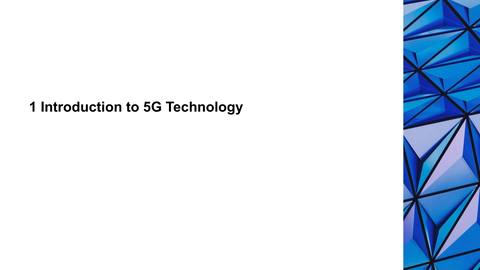}\\[3pt]
    \includegraphics[width=0.32\textwidth]{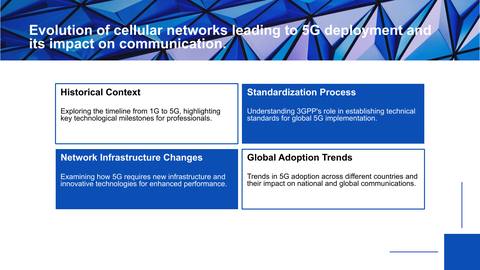}
    \includegraphics[width=0.32\textwidth]{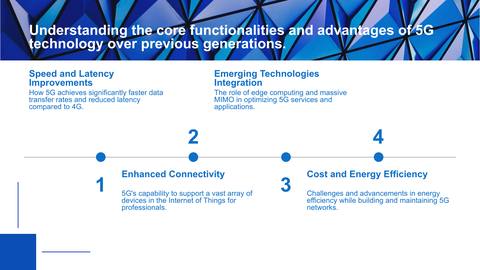}
    \includegraphics[width=0.32\textwidth]{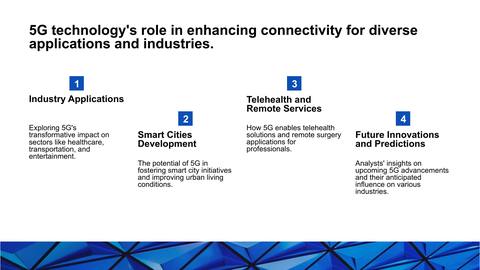}
    
    \textbf{(a) Gamma}
    \end{minipage}
    
    \vspace{6pt}
    
    % Kimi-Banana
    \begin{minipage}{\textwidth}
    \centering
    \includegraphics[width=0.32\textwidth]{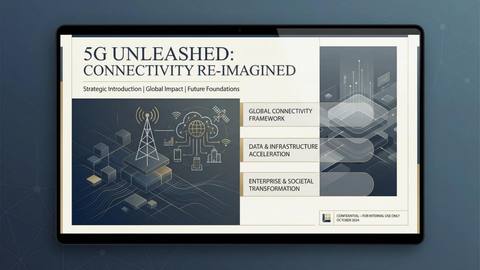}
    \includegraphics[width=0.32\textwidth]{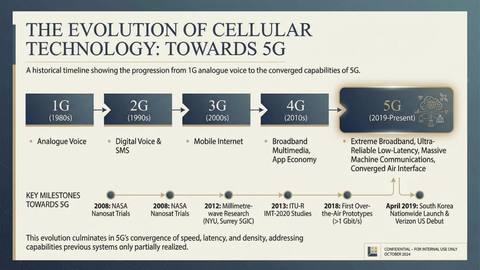}
    \includegraphics[width=0.32\textwidth]{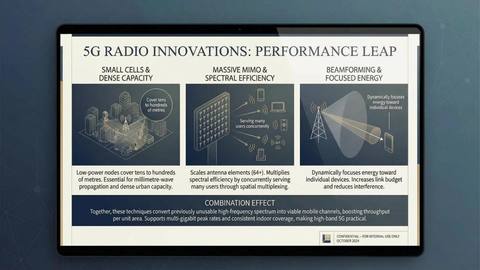}\\[3pt]
    \includegraphics[width=0.32\textwidth]{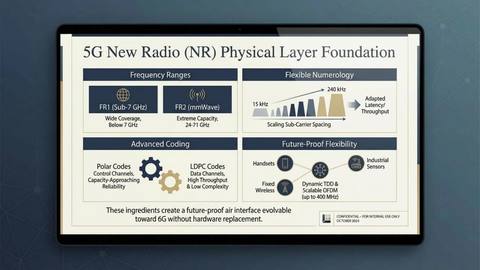}
    \includegraphics[width=0.32\textwidth]{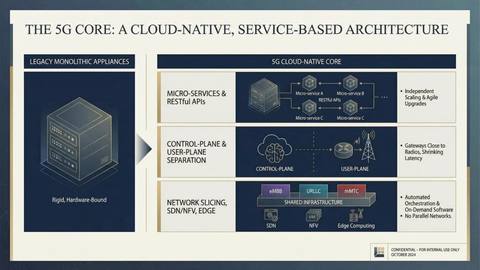}
    \includegraphics[width=0.32\textwidth]{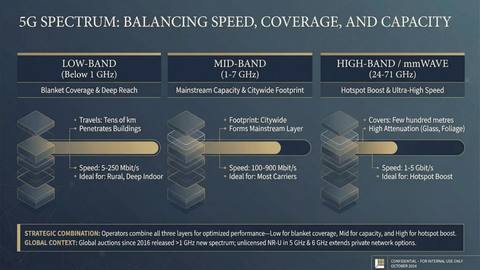}
    
    \textbf{(b) Kimi-Banana}
    \end{minipage}
    
    \vspace{6pt}
    
    % Kimi-Smart
    \begin{minipage}{\textwidth}
    \centering
    \includegraphics[width=0.32\textwidth]{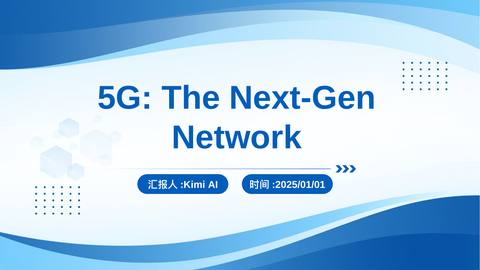}
    \includegraphics[width=0.32\textwidth]{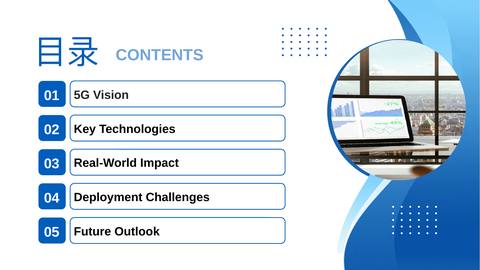}
    \includegraphics[width=0.32\textwidth]{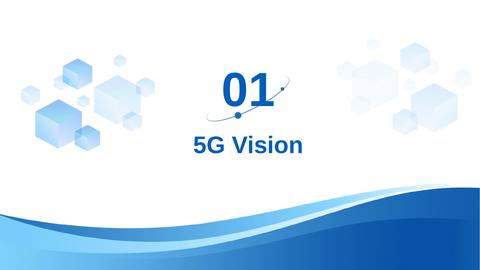}\\[3pt]
    \includegraphics[width=0.32\textwidth]{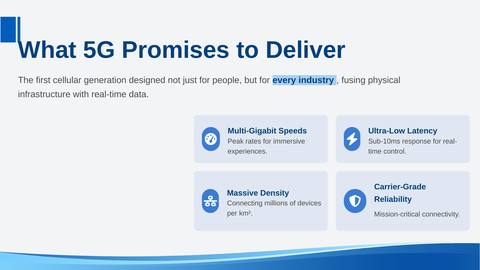}
    \includegraphics[width=0.32\textwidth]{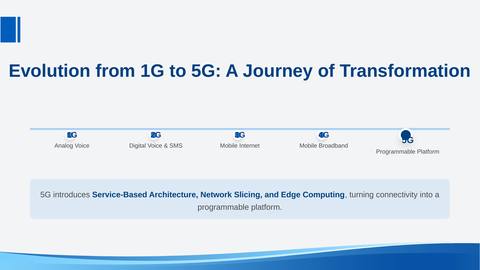}
    \includegraphics[width=0.32\textwidth]{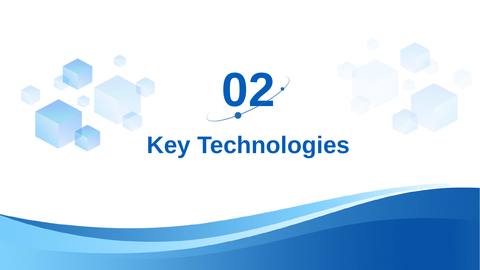}
    
    \textbf{(c) Kimi-Smart}
    \end{minipage}
    
    \caption{Comparison of slides generated on ``5G Technology'' (Part 1 of 3). Each subfigure shows 6 consecutive slides (2 rows $\times$ 3 columns) from a single product.}
    \label{fig:5g-comparison-1}
\end{figure*}

\begin{figure*}[htbp]
    \centering
    
    % Kimi-Standard
    \begin{minipage}{\textwidth}
    \centering
    \includegraphics[width=0.32\textwidth]{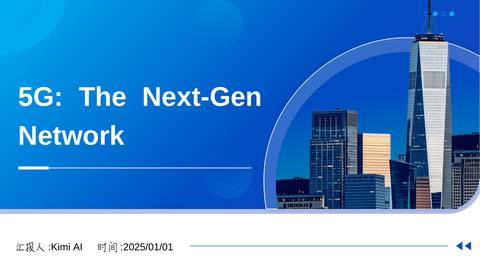}
    \includegraphics[width=0.32\textwidth]{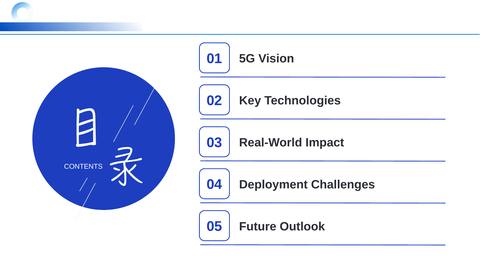}
    \includegraphics[width=0.32\textwidth]{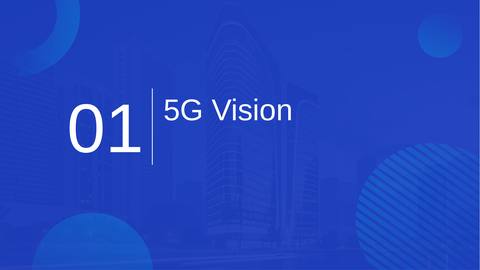}\\[3pt]
    \includegraphics[width=0.32\textwidth]{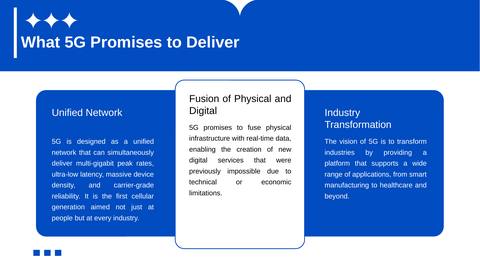}
    \includegraphics[width=0.32\textwidth]{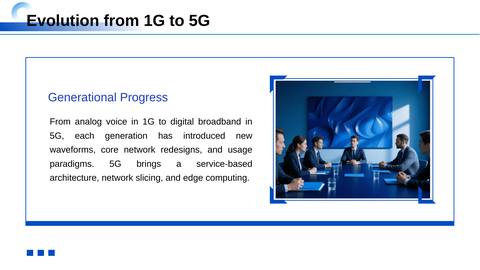}
    \includegraphics[width=0.32\textwidth]{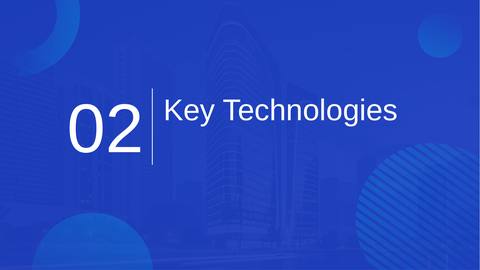}
    
    \textbf{(d) Kimi-Standard}
    \end{minipage}
    
    \vspace{6pt}
    
    % NotebookLM
    \begin{minipage}{\textwidth}
    \centering
    \includegraphics[width=0.32\textwidth]{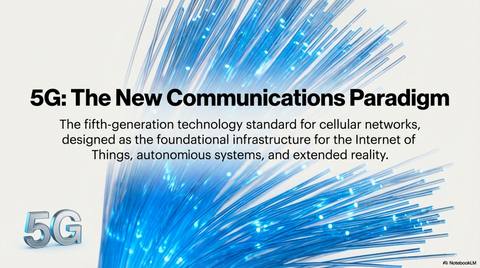}
    \includegraphics[width=0.32\textwidth]{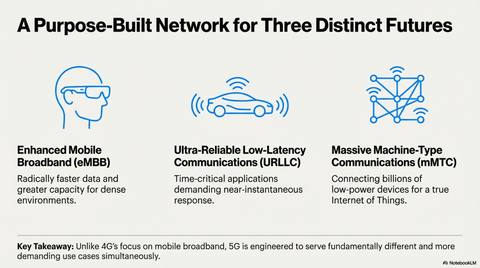}
    \includegraphics[width=0.32\textwidth]{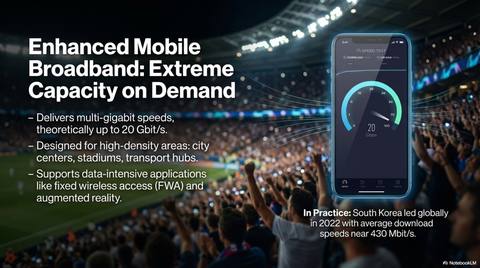}\\[3pt]
    \includegraphics[width=0.32\textwidth]{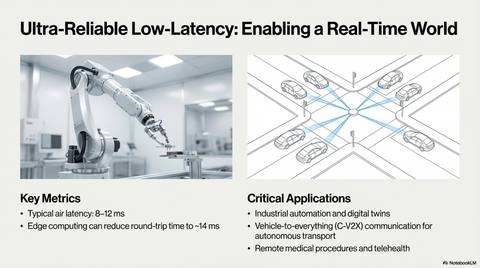}
    \includegraphics[width=0.32\textwidth]{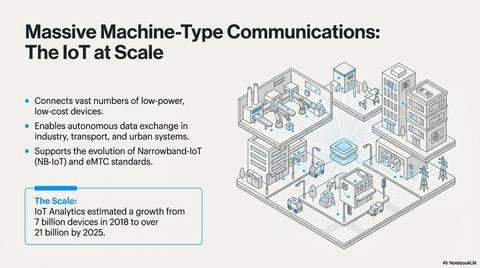}
    \includegraphics[width=0.32\textwidth]{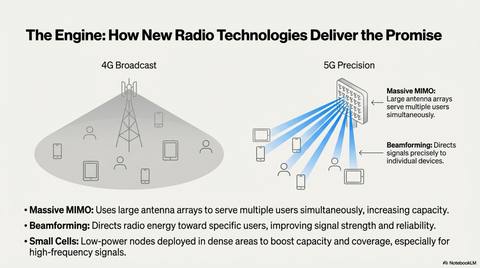}
    
    \textbf{(e) NotebookLM}
    \end{minipage}
    
    \vspace{6pt}
    
    % Quake
    \begin{minipage}{\textwidth}
    \centering
    \includegraphics[width=0.32\textwidth]{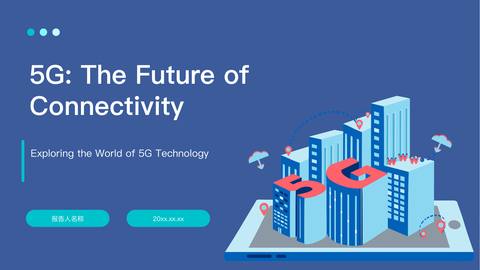}
    \includegraphics[width=0.32\textwidth]{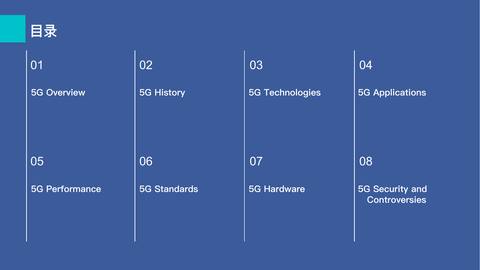}
    \includegraphics[width=0.32\textwidth]{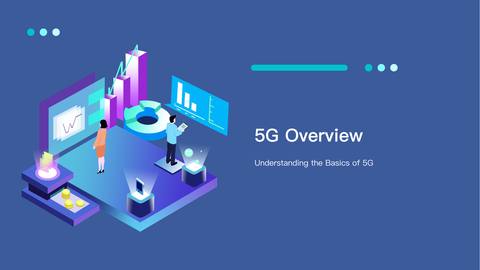}\\[3pt]
    \includegraphics[width=0.32\textwidth]{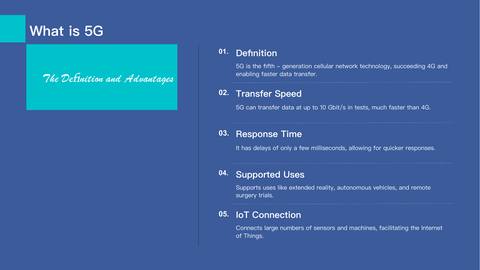}
    \includegraphics[width=0.32\textwidth]{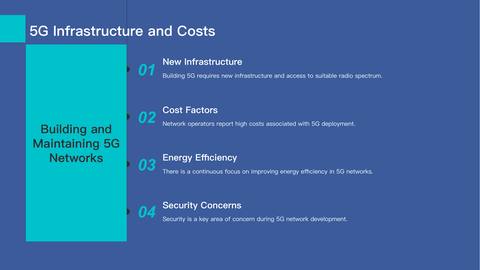}
    \includegraphics[width=0.32\textwidth]{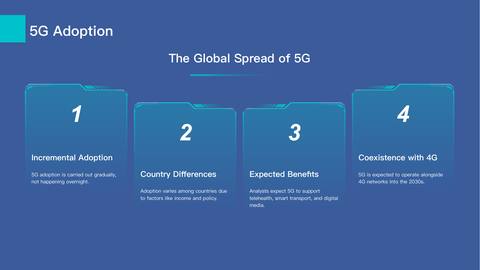}
    
    \textbf{(f) Quake}
    \end{minipage}
    
    \caption{Comparison of slides generated on ``5G Technology'' (Part 2 of 3). Each subfigure shows 6 consecutive slides (2 rows $\times$ 3 columns) from a single product.}
    \label{fig:5g-comparison-2}
\end{figure*}

\begin{figure*}[htbp]
    \centering
    
    % Skywork
    \begin{minipage}{\textwidth}
    \centering
    \includegraphics[width=0.32\textwidth]{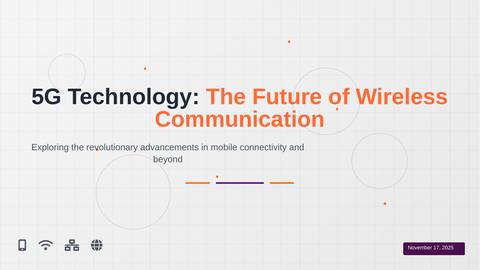}
    \includegraphics[width=0.32\textwidth]{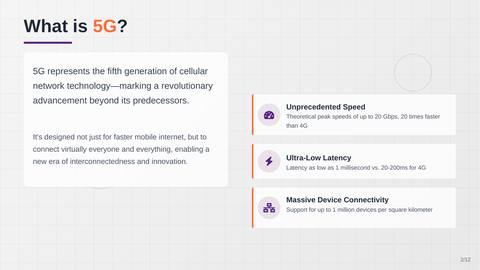}
    \includegraphics[width=0.32\textwidth]{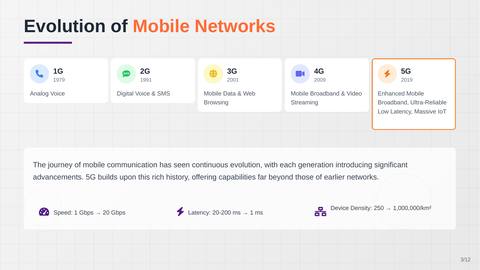}\\[3pt]
    \includegraphics[width=0.32\textwidth]{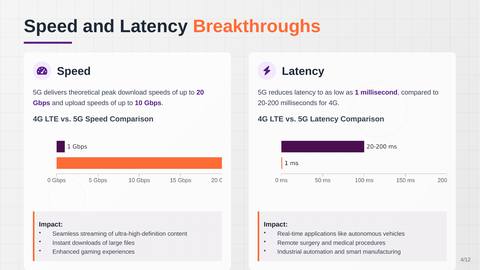}
    \includegraphics[width=0.32\textwidth]{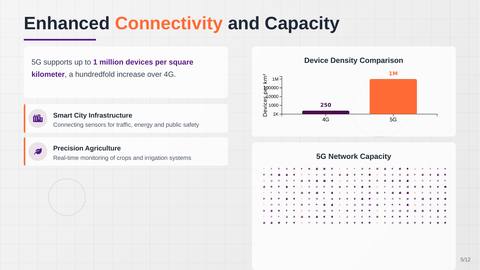}
    \includegraphics[width=0.32\textwidth]{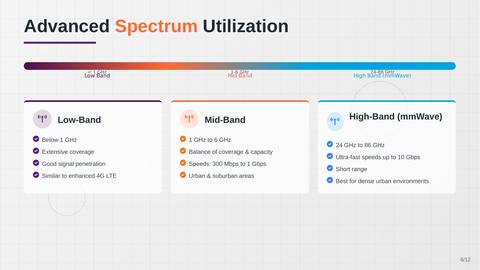}
    
    \textbf{(g) Skywork}
    \end{minipage}
    
    \vspace{6pt}
    
    % Skyworks-Banana
    \begin{minipage}{\textwidth}
    \centering
    \includegraphics[width=0.32\textwidth]{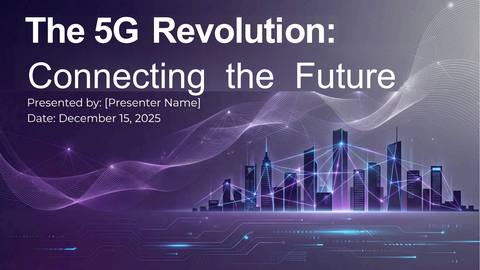}
    \includegraphics[width=0.32\textwidth]{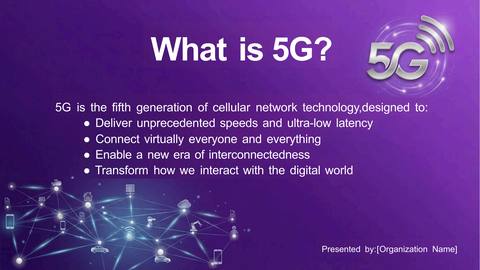}
    \includegraphics[width=0.32\textwidth]{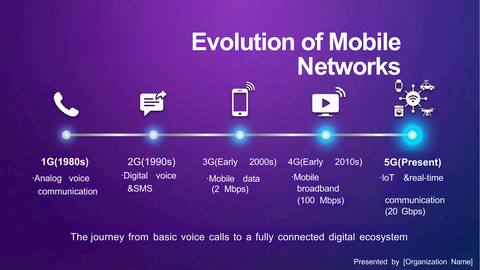}\\[3pt]
    \includegraphics[width=0.32\textwidth]{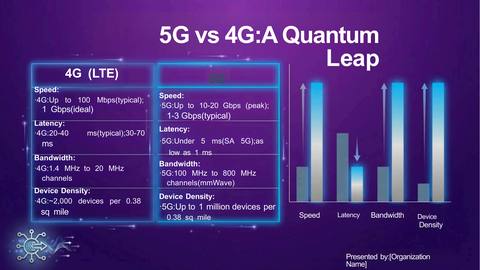}
    \includegraphics[width=0.32\textwidth]{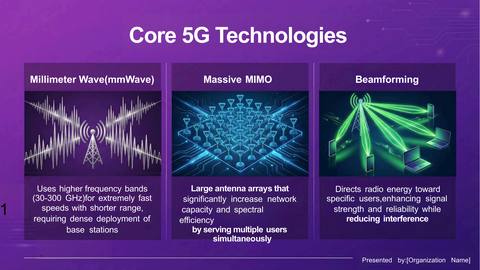}
    \includegraphics[width=0.32\textwidth]{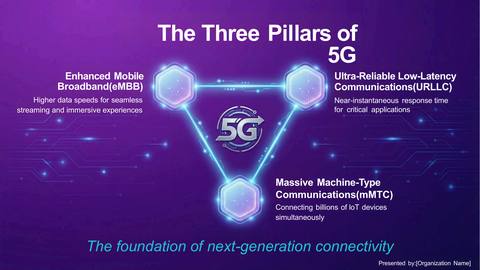}
    
    \textbf{(h) Skyworks-Banana}
    \end{minipage}
    
    \vspace{6pt}
    
    % Zhipu
    \begin{minipage}{\textwidth}
    \centering
    \includegraphics[width=0.32\textwidth]{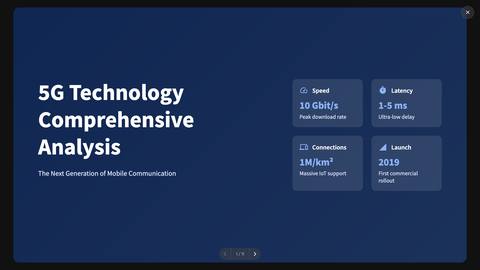}
    \includegraphics[width=0.32\textwidth]{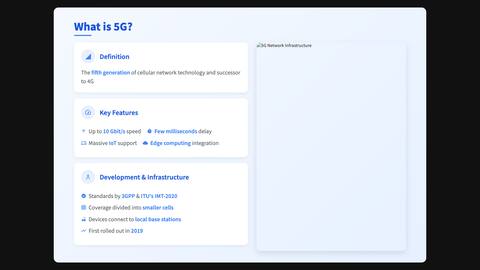}
    \includegraphics[width=0.32\textwidth]{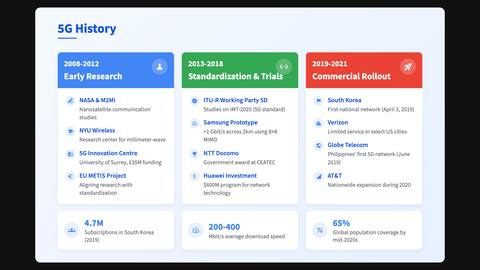}\\[3pt]
    \includegraphics[width=0.32\textwidth]{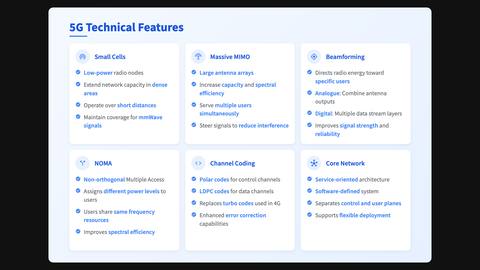}
    \includegraphics[width=0.32\textwidth]{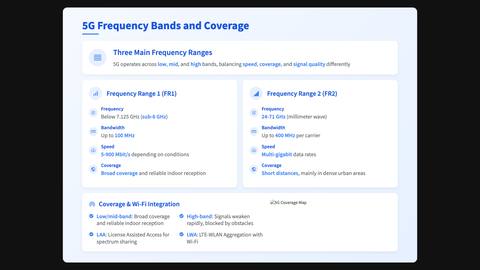}
    \includegraphics[width=0.32\textwidth]{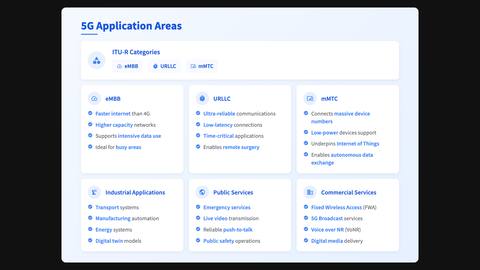}
    
    \textbf{(i) Zhipu}
    \end{minipage}
    
    \caption{Comparison of slides generated on ``5G Technology'' (Part 3 of 3). Each subfigure shows 6 consecutive slides (2 rows $\times$ 3 columns) from a single product.}
    \label{fig:5g-comparison-3}
\end{figure*}

\begin{figure*}[htbp]
    \centering
    
    % Gamma
    \begin{minipage}{\textwidth}
    \centering
    \includegraphics[width=0.32\textwidth]{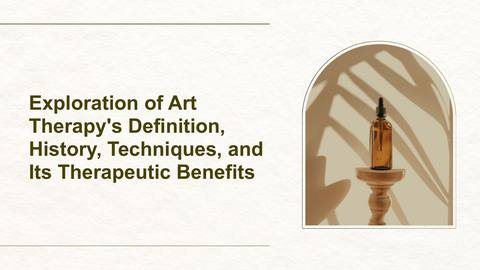}
    \includegraphics[width=0.32\textwidth]{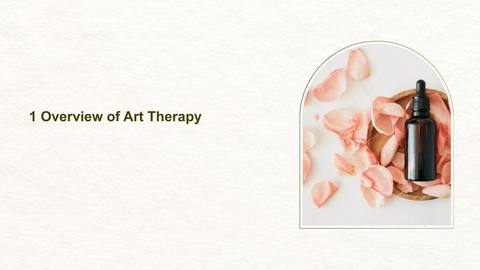}
    \includegraphics[width=0.32\textwidth]{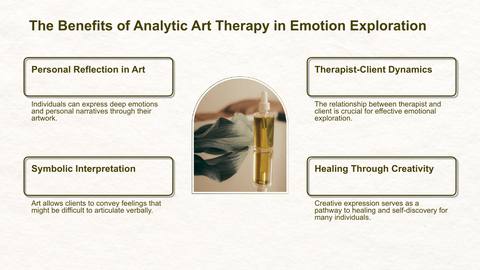}\\[3pt]
    \includegraphics[width=0.32\textwidth]{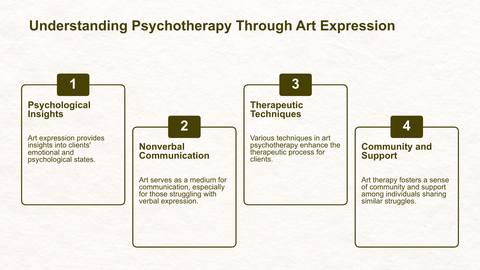}
    \includegraphics[width=0.32\textwidth]{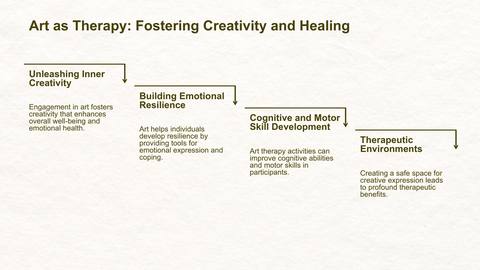}
    \includegraphics[width=0.32\textwidth]{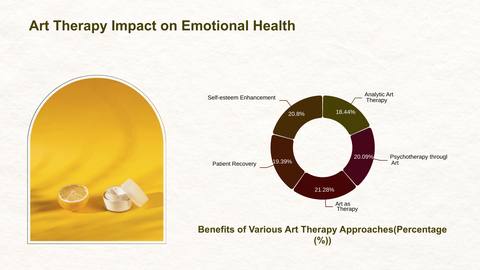}
    
    \textbf{(a) Gamma}
    \end{minipage}
    
    \vspace{6pt}
    
    % Kimi-Banana
    \begin{minipage}{\textwidth}
    \centering
    \includegraphics[width=0.32\textwidth]{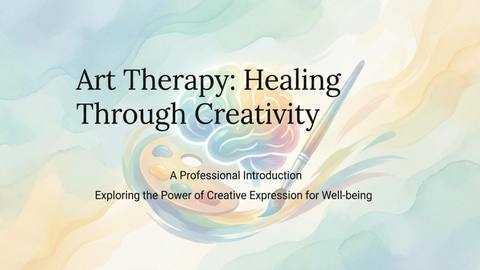}
    \includegraphics[width=0.32\textwidth]{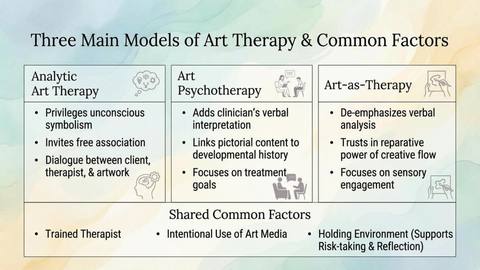}
    \includegraphics[width=0.32\textwidth]{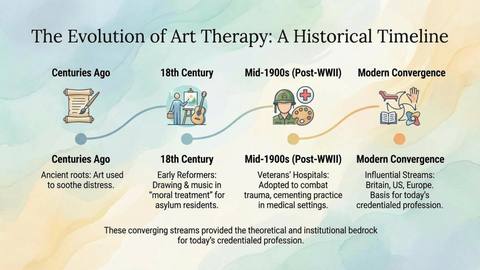}\\[3pt]
    \includegraphics[width=0.32\textwidth]{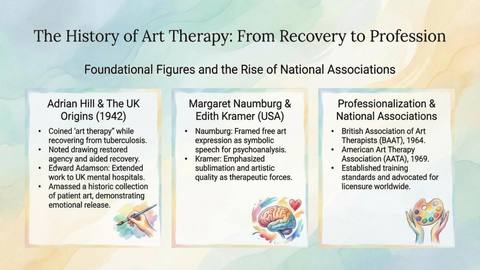}
    \includegraphics[width=0.32\textwidth]{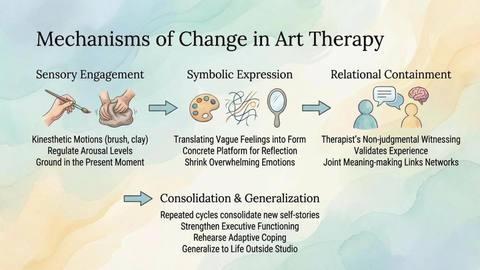}
    \includegraphics[width=0.32\textwidth]{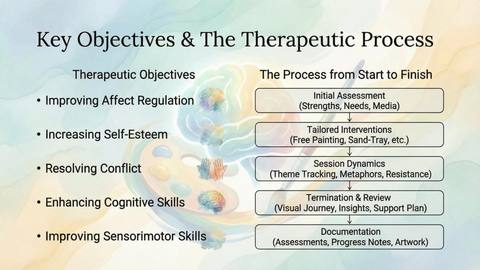}
    
    \textbf{(b) Kimi-Banana}
    \end{minipage}
    
    \vspace{6pt}
    
    % Kimi-Smart
    \begin{minipage}{\textwidth}
    \centering
    \includegraphics[width=0.32\textwidth]{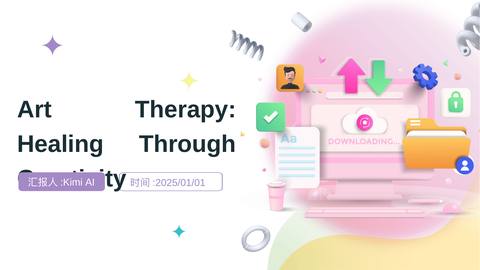}
    \includegraphics[width=0.32\textwidth]{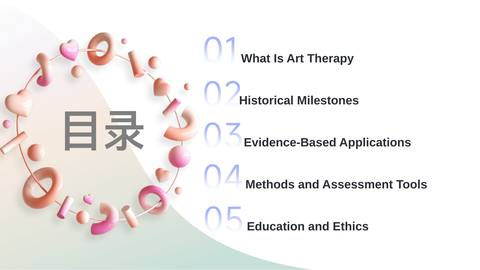}
    \includegraphics[width=0.32\textwidth]{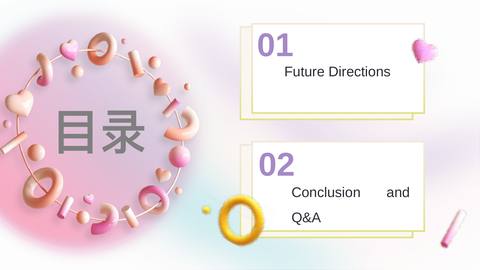}\\[3pt]
    \includegraphics[width=0.32\textwidth]{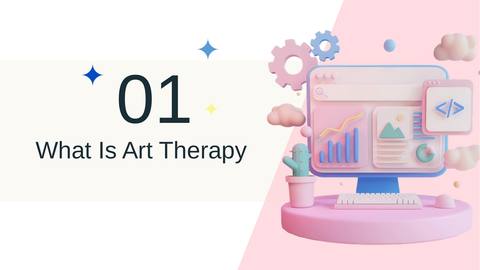}
    \includegraphics[width=0.32\textwidth]{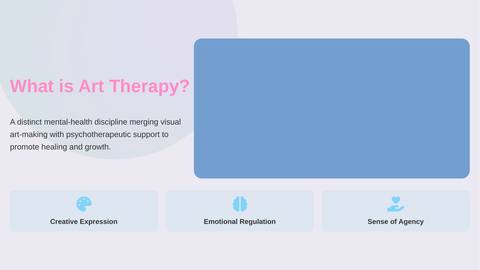}
    \includegraphics[width=0.32\textwidth]{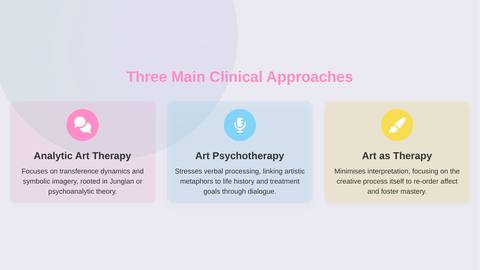}
    
    \textbf{(c) Kimi-Smart}
    \end{minipage}
    
    \caption{Comparison of slides generated on ``Art Therapy'' (Part 1 of 3). Each subfigure shows 6 consecutive slides (2 rows $\times$ 3 columns) from a single product.}
    \label{fig:art-therapy-comparison-1}
\end{figure*}

\begin{figure*}[htbp]
    \centering
    
    % Kimi-Standard
    \begin{minipage}{\textwidth}
    \centering
    \includegraphics[width=0.32\textwidth]{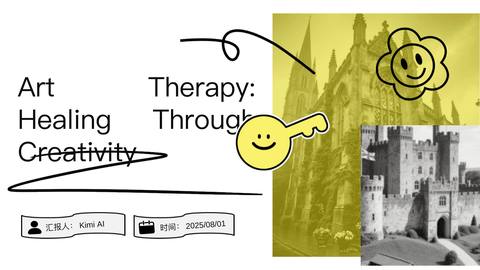}
    \includegraphics[width=0.32\textwidth]{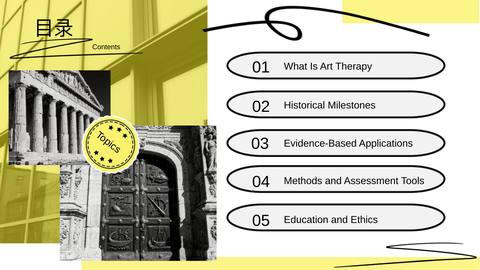}
    \includegraphics[width=0.32\textwidth]{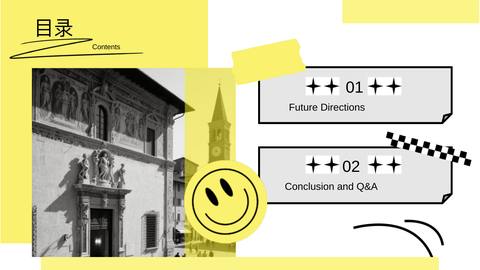}\\[3pt]
    \includegraphics[width=0.32\textwidth]{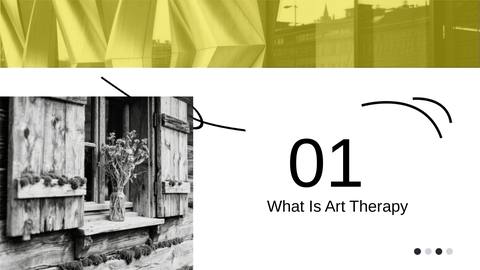}
    \includegraphics[width=0.32\textwidth]{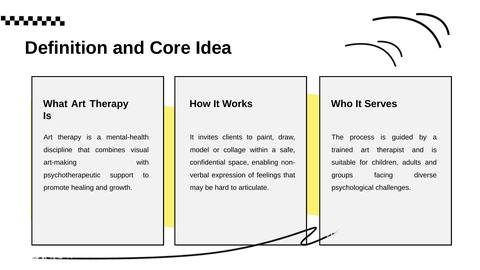}
    \includegraphics[width=0.32\textwidth]{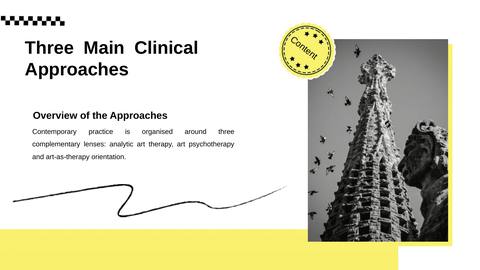}
    
    \textbf{(d) Kimi-Standard}
    \end{minipage}
    
    \vspace{6pt}
    
    % NotebookLM
    \begin{minipage}{\textwidth}
    \centering
    \includegraphics[width=0.32\textwidth]{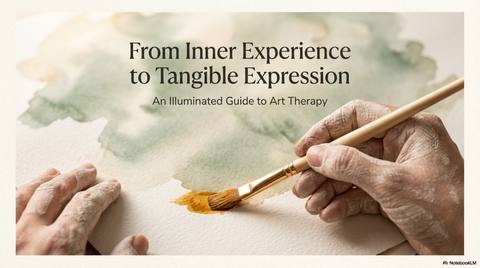}
    \includegraphics[width=0.32\textwidth]{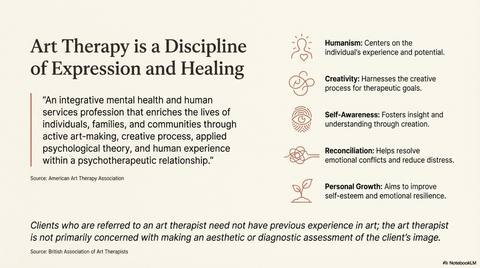}
    \includegraphics[width=0.32\textwidth]{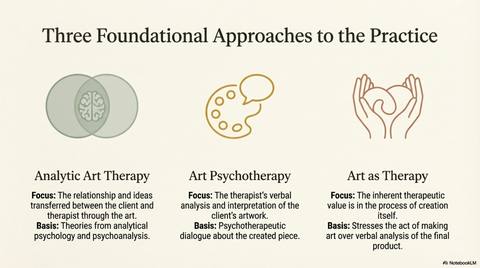}\\[3pt]
    \includegraphics[width=0.32\textwidth]{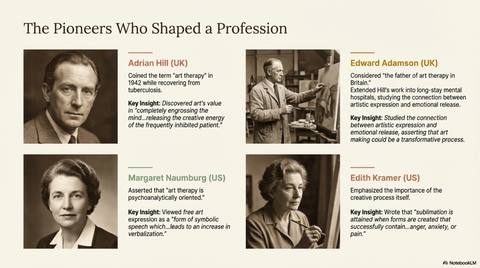}
    \includegraphics[width=0.32\textwidth]{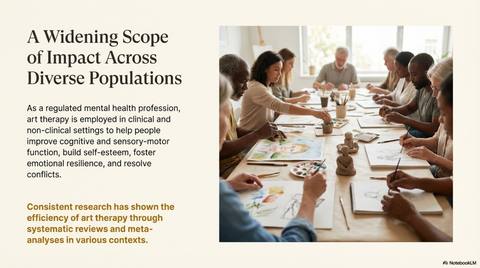}
    \includegraphics[width=0.32\textwidth]{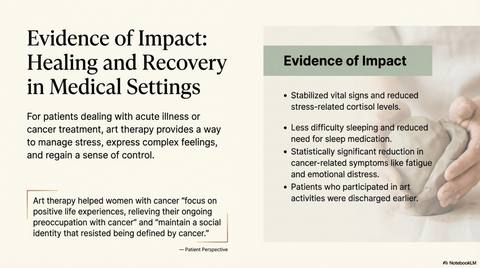}
    
    \textbf{(e) NotebookLM}
    \end{minipage}
    
    \vspace{6pt}
    
    % Quake
    \begin{minipage}{\textwidth}
    \centering
    \includegraphics[width=0.32\textwidth]{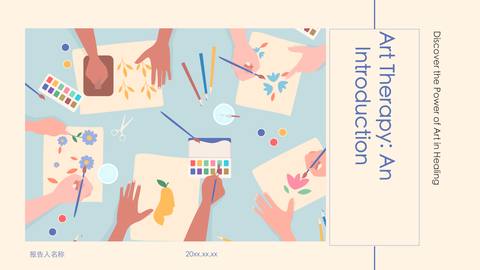}
    \includegraphics[width=0.32\textwidth]{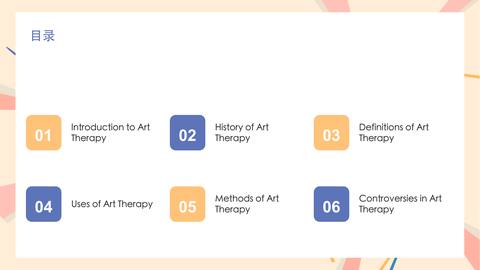}
    \includegraphics[width=0.32\textwidth]{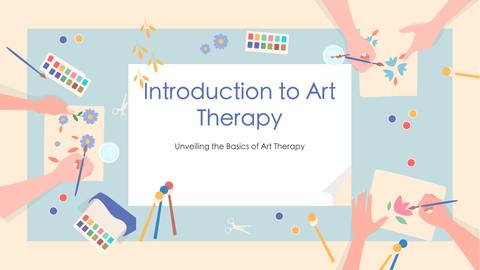}\\[3pt]
    \includegraphics[width=0.32\textwidth]{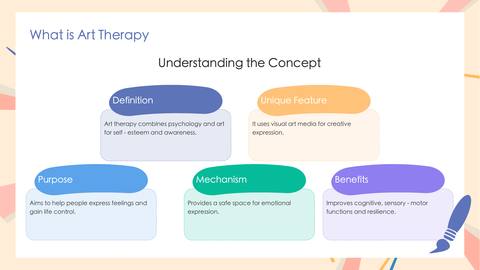}
    \includegraphics[width=0.32\textwidth]{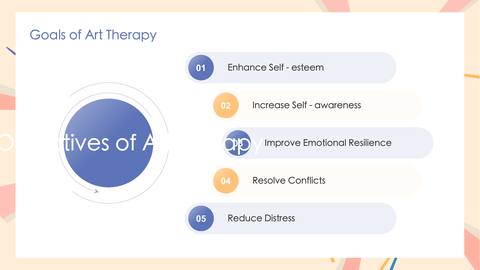}
    \includegraphics[width=0.32\textwidth]{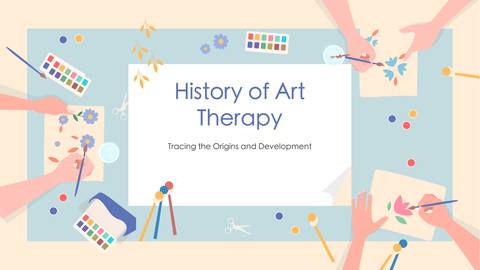}
    
    \textbf{(f) Quake}
    \end{minipage}
    
    \caption{Comparison of slides generated on ``Art Therapy'' (Part 2 of 3). Each subfigure shows 6 consecutive slides (2 rows $\times$ 3 columns) from a single product.}
    \label{fig:art-therapy-comparison-2}
\end{figure*}

\begin{figure*}[htbp]
    \centering
    
    % Skywork
    \begin{minipage}{\textwidth}
    \centering
    \includegraphics[width=0.32\textwidth]{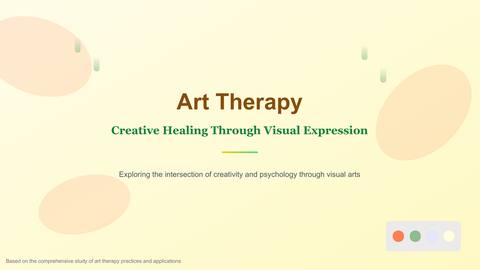}
    \includegraphics[width=0.32\textwidth]{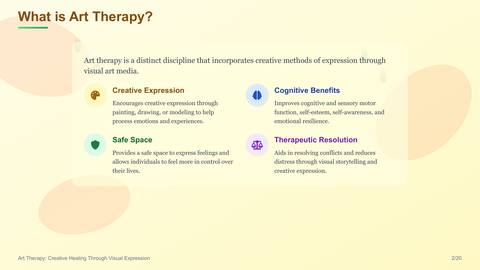}
    \includegraphics[width=0.32\textwidth]{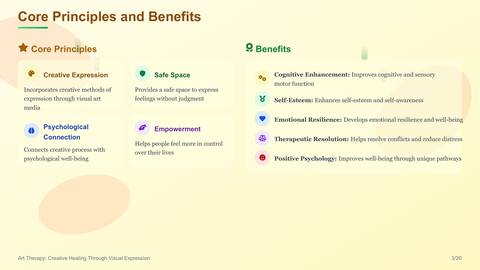}\\[3pt]
    \includegraphics[width=0.32\textwidth]{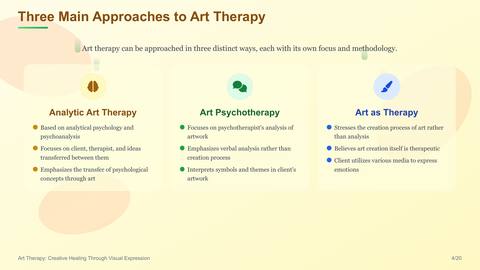}
    \includegraphics[width=0.32\textwidth]{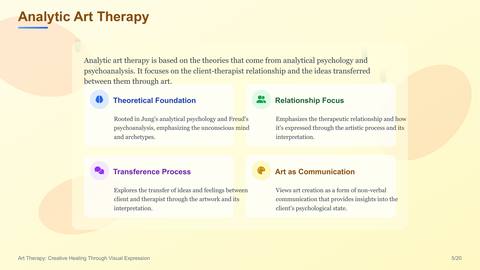}
    \includegraphics[width=0.32\textwidth]{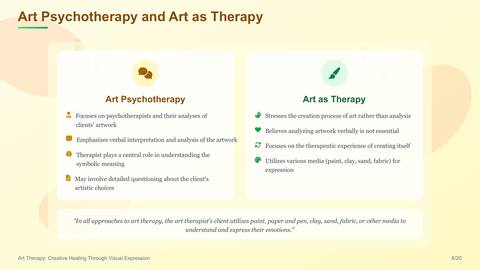}
    
    \textbf{(g) Skywork}
    \end{minipage}
    
    \vspace{6pt}
    
    % Skyworks-Banana
    \begin{minipage}{\textwidth}
    \centering
    \includegraphics[width=0.32\textwidth]{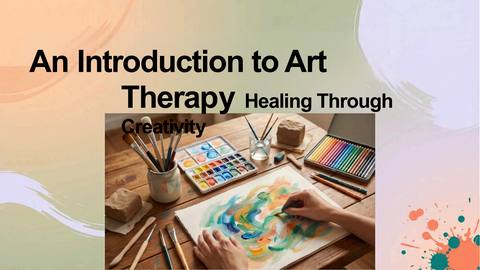}
    \includegraphics[width=0.32\textwidth]{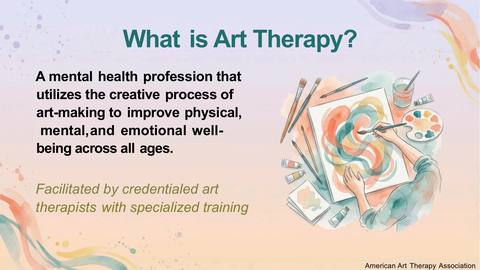}
    \includegraphics[width=0.32\textwidth]{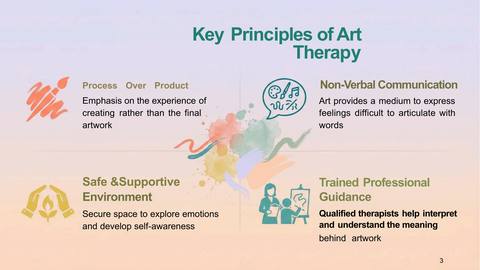}\\[3pt]
    \includegraphics[width=0.32\textwidth]{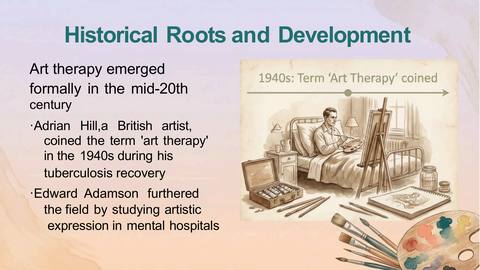}
    \includegraphics[width=0.32\textwidth]{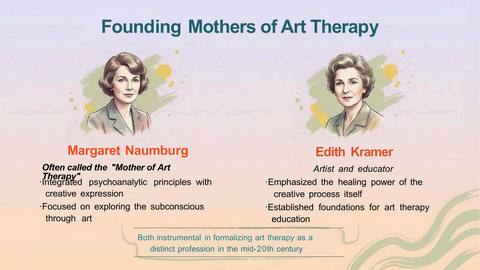}
    \includegraphics[width=0.32\textwidth]{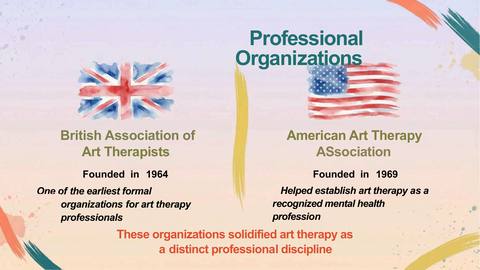}
    
    \textbf{(h) Skyworks-Banana}
    \end{minipage}
    
    \vspace{6pt}
    
    % Zhipu
    \begin{minipage}{\textwidth}
    \centering
    \includegraphics[width=0.32\textwidth]{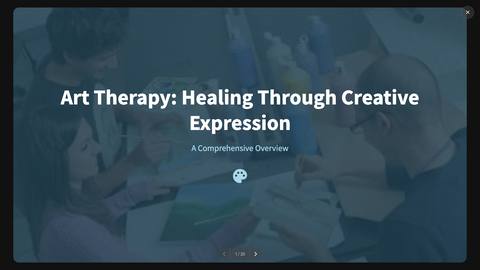}
    \includegraphics[width=0.32\textwidth]{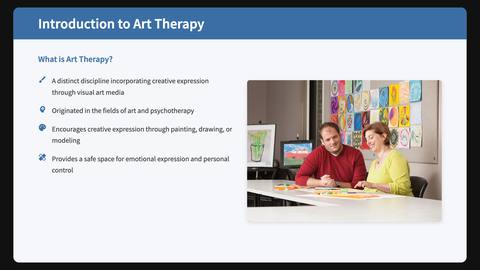}
    \includegraphics[width=0.32\textwidth]{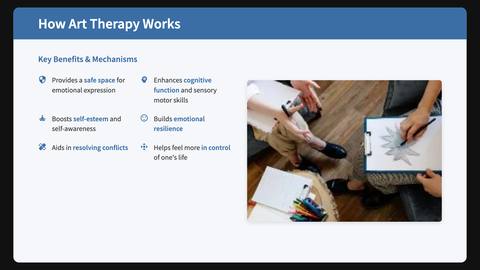}\\[3pt]
    \includegraphics[width=0.32\textwidth]{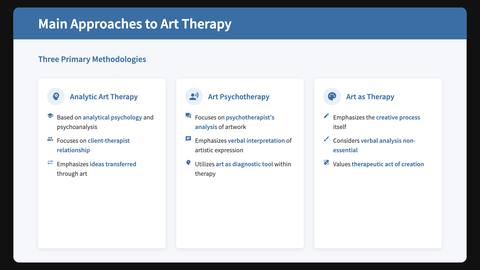}
    \includegraphics[width=0.32\textwidth]{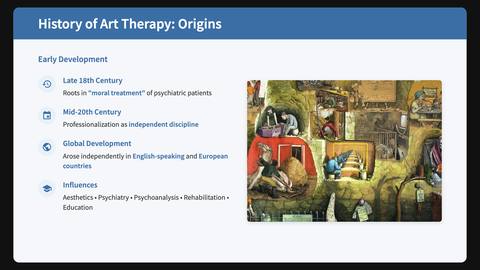}
    \includegraphics[width=0.32\textwidth]{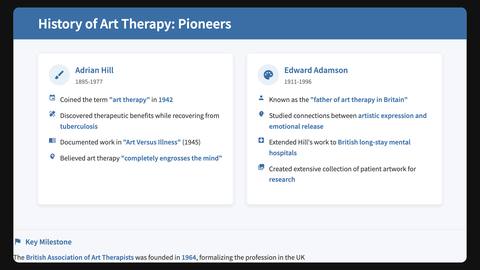}
    
    \textbf{(i) Zhipu}
    \end{minipage}
    
    \caption{Comparison of slides generated on ``Art Therapy'' (Part 3 of 3). Each subfigure shows 6 consecutive slides (2 rows $\times$ 3 columns) from a single product.}
    \label{fig:art-therapy-comparison-3}
\end{figure*}

\begin{figure*}[htbp]
    \centering
    
    % Gamma
    \begin{minipage}{\textwidth}
    \centering
    \includegraphics[width=0.32\textwidth]{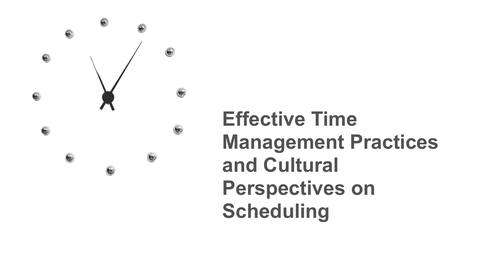}
    \includegraphics[width=0.32\textwidth]{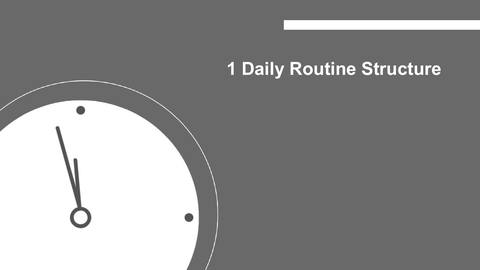}
    \includegraphics[width=0.32\textwidth]{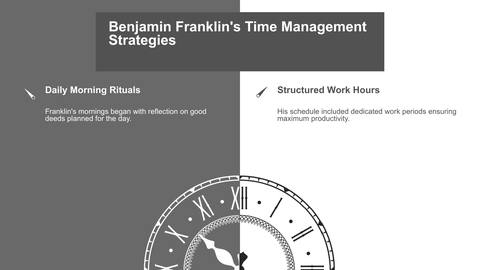}\\[3pt]
    \includegraphics[width=0.32\textwidth]{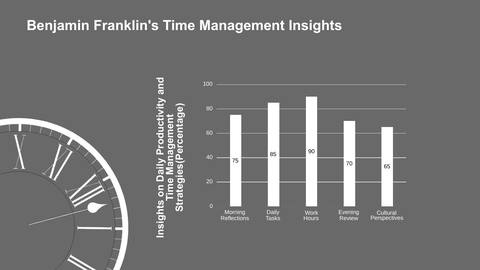}
    \includegraphics[width=0.32\textwidth]{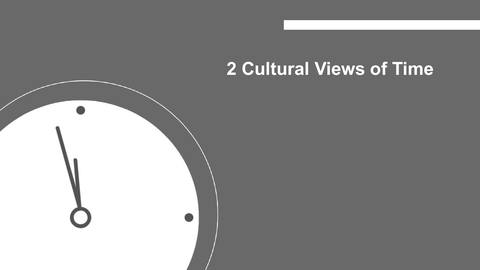}
    \includegraphics[width=0.32\textwidth]{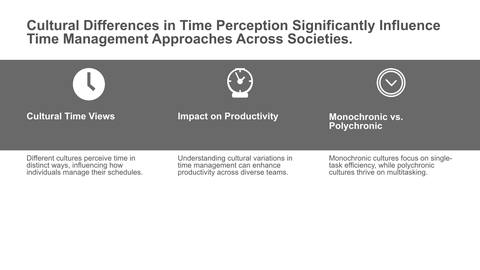}
    
    \textbf{(a) Gamma}
    \end{minipage}
    
    \vspace{6pt}
    
    % Kimi-Banana
    \begin{minipage}{\textwidth}
    \centering
    \includegraphics[width=0.32\textwidth]{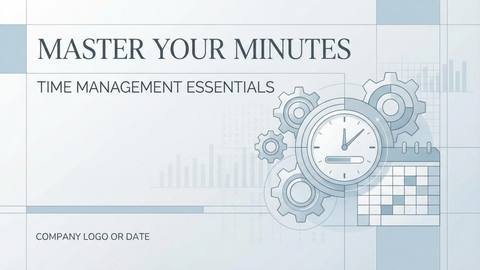}
    \includegraphics[width=0.32\textwidth]{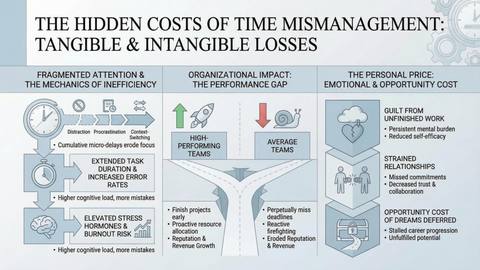}
    \includegraphics[width=0.32\textwidth]{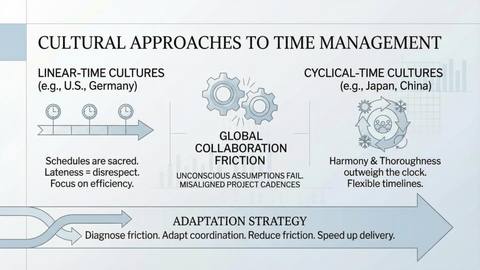}\\[3pt]
    \includegraphics[width=0.32\textwidth]{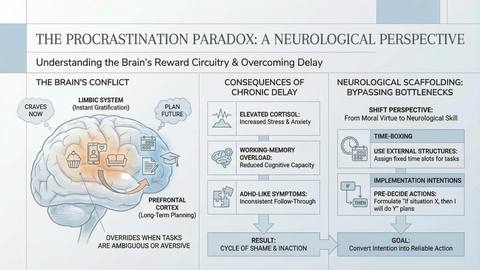}
    \includegraphics[width=0.32\textwidth]{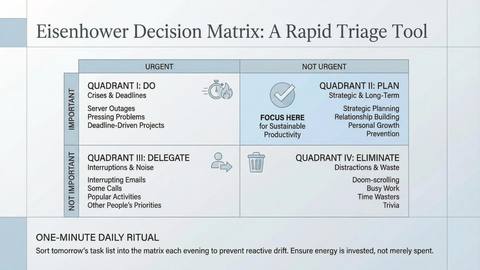}
    \includegraphics[width=0.32\textwidth]{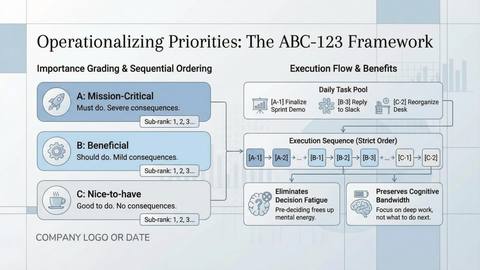}
    
    \textbf{(b) Kimi-Banana}
    \end{minipage}
    
    \vspace{6pt}
    
    % Kimi-Smart
    \begin{minipage}{\textwidth}
    \centering
    \includegraphics[width=0.32\textwidth]{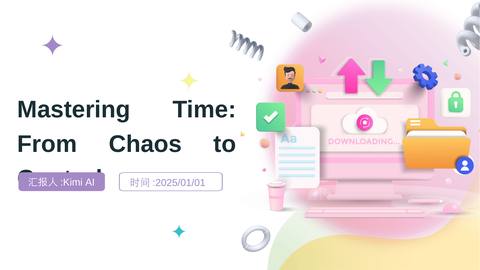}
    \includegraphics[width=0.32\textwidth]{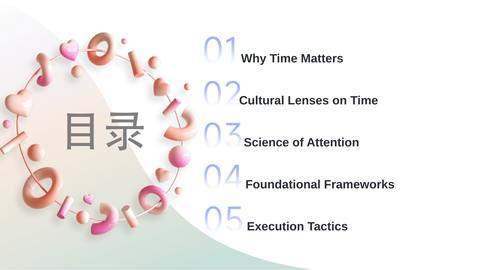}
    \includegraphics[width=0.32\textwidth]{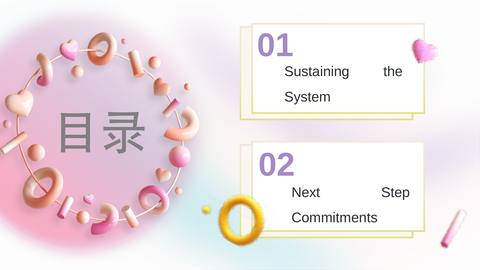}\\[3pt]
    \includegraphics[width=0.32\textwidth]{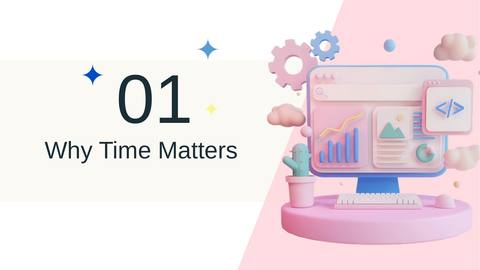}
    \includegraphics[width=0.32\textwidth]{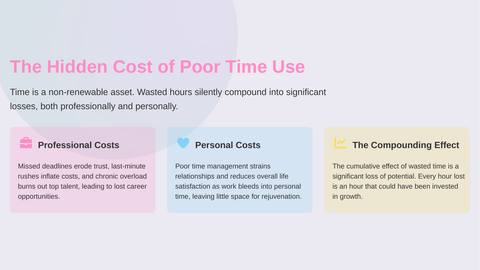}
    \includegraphics[width=0.32\textwidth]{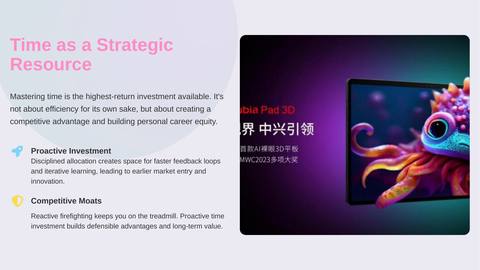}
    
    \textbf{(c) Kimi-Smart}
    \end{minipage}
    
    \caption{Comparison of slides generated on ``Time Management'' (Part 1 of 3). Each subfigure shows 6 consecutive slides (2 rows $\times$ 3 columns) from a single product.}
    \label{fig:time-management-comparison-1}
\end{figure*}

\begin{figure*}[htbp]
    \centering
    
    % Kimi-Standard
    \begin{minipage}{\textwidth}
    \centering
    \includegraphics[width=0.32\textwidth]{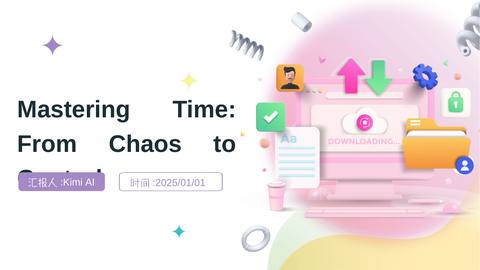}
    \includegraphics[width=0.32\textwidth]{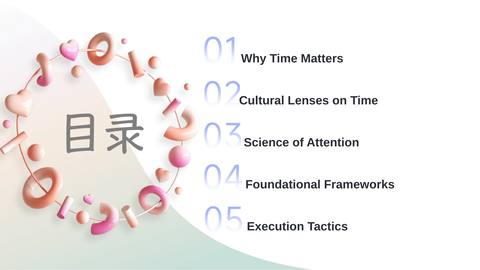}
    \includegraphics[width=0.32\textwidth]{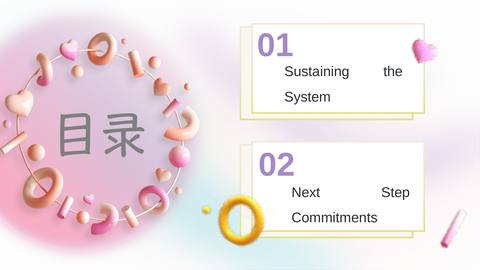}\\[3pt]
    \includegraphics[width=0.32\textwidth]{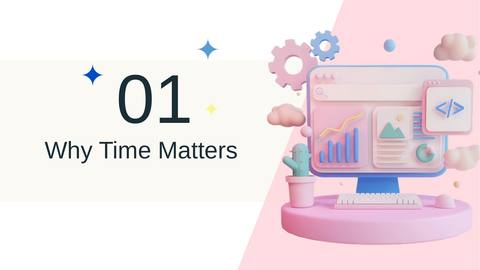}
    \includegraphics[width=0.32\textwidth]{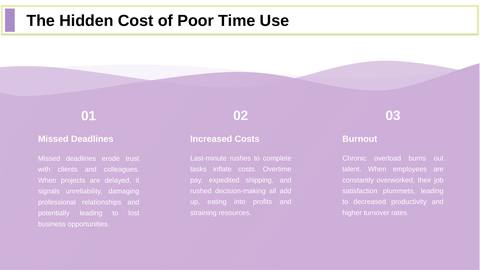}
    \includegraphics[width=0.32\textwidth]{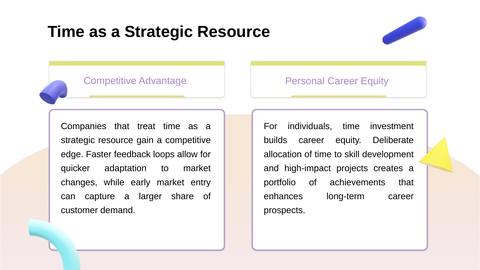}
    
    \textbf{(d) Kimi-Standard}
    \end{minipage}
    
    \vspace{6pt}
    
    % NotebookLM
    \begin{minipage}{\textwidth}
    \centering
    \includegraphics[width=0.32\textwidth]{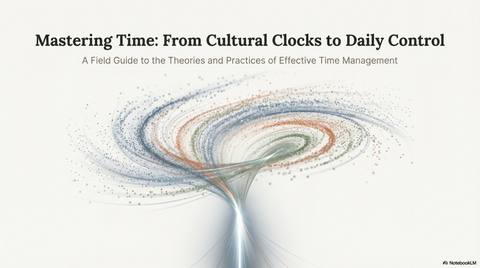}
    \includegraphics[width=0.32\textwidth]{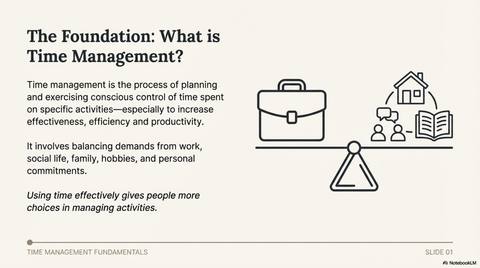}
    \includegraphics[width=0.32\textwidth]{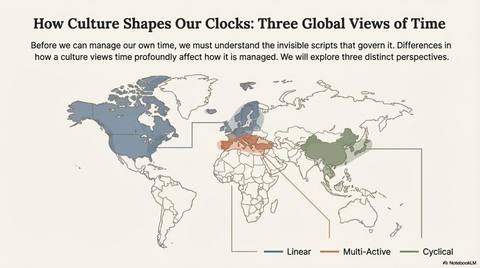}\\[3pt]
    \includegraphics[width=0.32\textwidth]{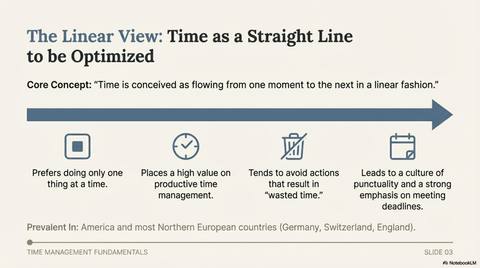}
    \includegraphics[width=0.32\textwidth]{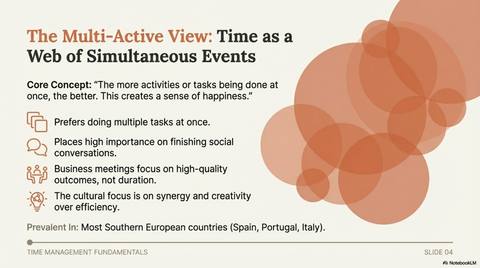}
    \includegraphics[width=0.32\textwidth]{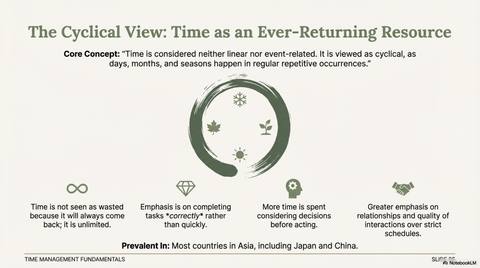}
    
    \textbf{(e) NotebookLM}
    \end{minipage}
    
    \vspace{6pt}
    
    % Quake
    \begin{minipage}{\textwidth}
    \centering
    \includegraphics[width=0.32\textwidth]{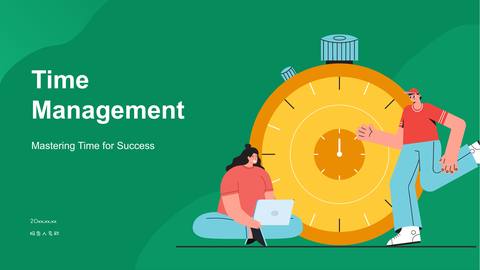}
    \includegraphics[width=0.32\textwidth]{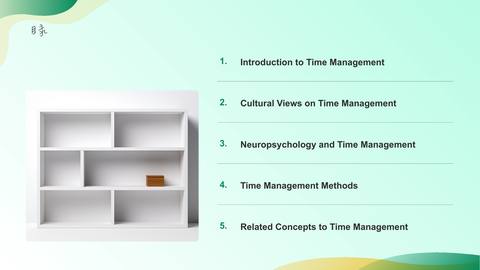}
    \includegraphics[width=0.32\textwidth]{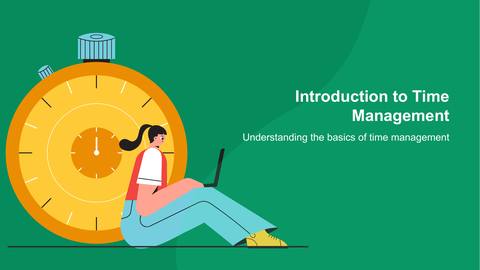}\\[3pt]
    \includegraphics[width=0.32\textwidth]{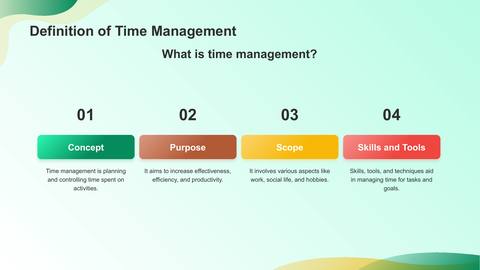}
    \includegraphics[width=0.32\textwidth]{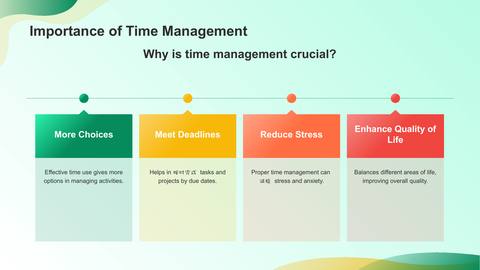}
    \includegraphics[width=0.32\textwidth]{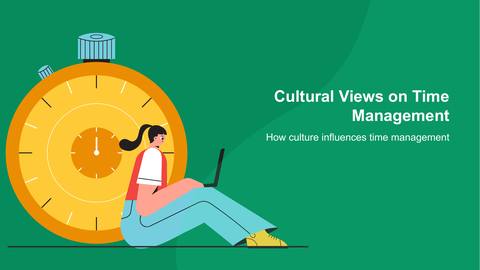}
    
    \textbf{(f) Quake}
    \end{minipage}
    
    \caption{Comparison of slides generated on ``Time Management'' (Part 2 of 3). Each subfigure shows 6 consecutive slides (2 rows $\times$ 3 columns) from a single product.}
    \label{fig:time-management-comparison-2}
\end{figure*}

\begin{figure*}[htbp]
    \centering
    
    % Skywork
    \begin{minipage}{\textwidth}
    \centering
    \includegraphics[width=0.32\textwidth]{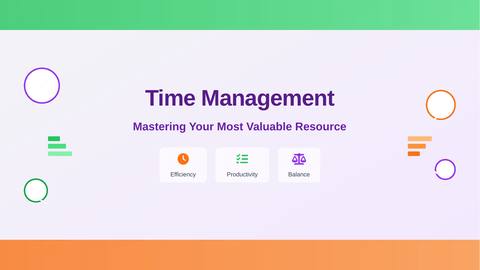}
    \includegraphics[width=0.32\textwidth]{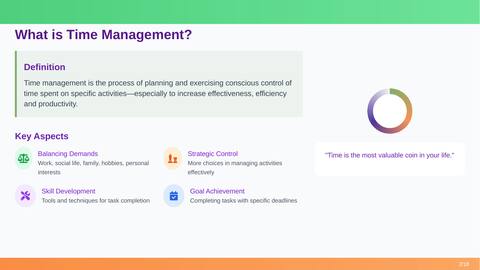}
    \includegraphics[width=0.32\textwidth]{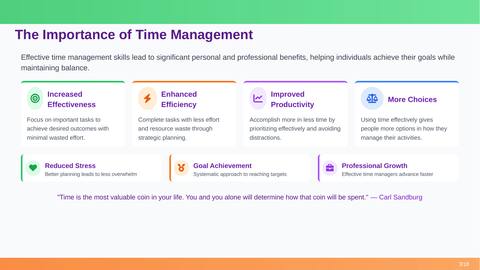}\\[3pt]
    \includegraphics[width=0.32\textwidth]{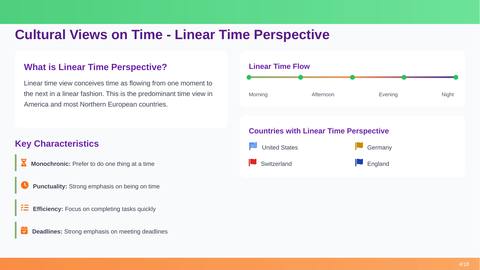}
    \includegraphics[width=0.32\textwidth]{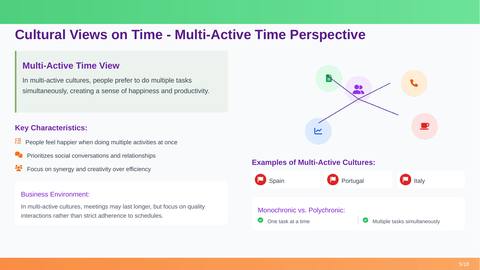}
    \includegraphics[width=0.32\textwidth]{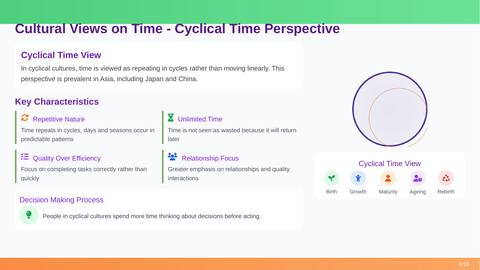}
    
    \textbf{(g) Skywork}
    \end{minipage}
    
    \vspace{6pt}
    
    % Skyworks-Banana
    \begin{minipage}{\textwidth}
    \centering
    \includegraphics[width=0.32\textwidth]{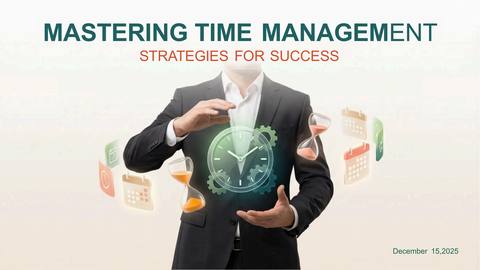}
    \includegraphics[width=0.32\textwidth]{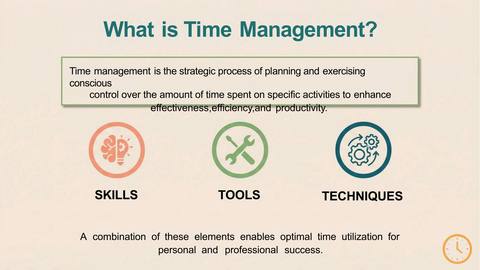}
    \includegraphics[width=0.32\textwidth]{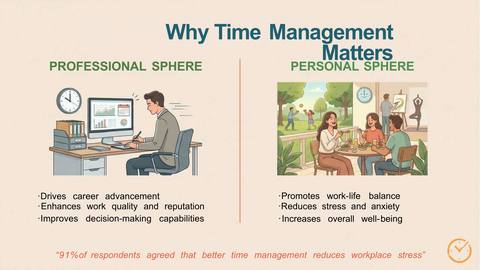}\\[3pt]
    \includegraphics[width=0.32\textwidth]{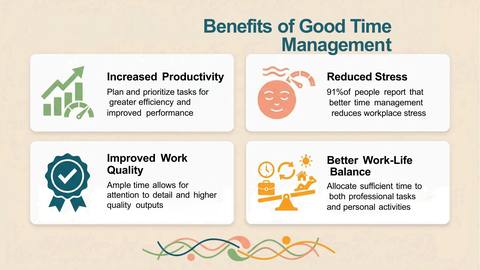}
    \includegraphics[width=0.32\textwidth]{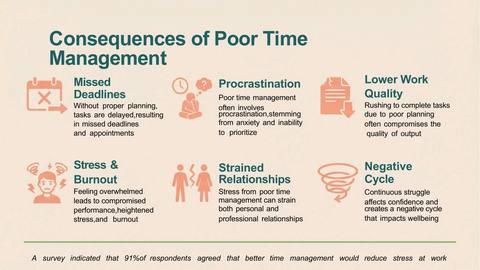}
    \includegraphics[width=0.32\textwidth]{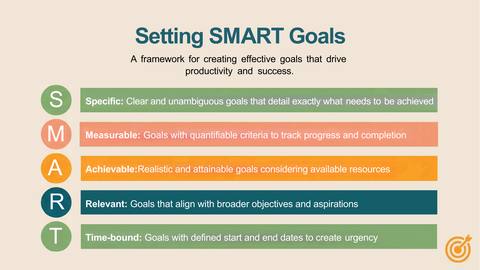}
    
    \textbf{(h) Skyworks-Banana}
    \end{minipage}
    
    \vspace{6pt}
    
    % Zhipu
    \begin{minipage}{\textwidth}
    \centering
    \includegraphics[width=0.32\textwidth]{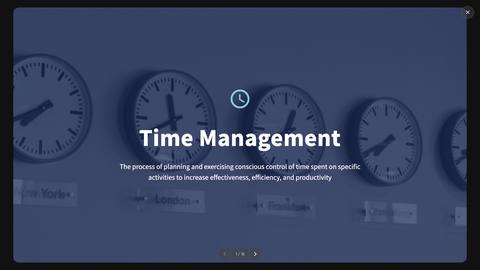}
    \includegraphics[width=0.32\textwidth]{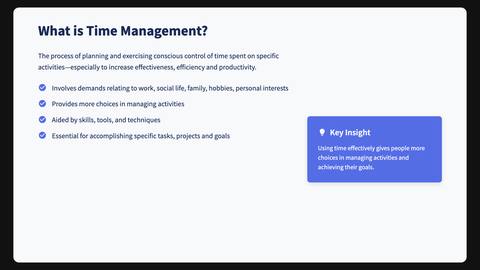}
    \includegraphics[width=0.32\textwidth]{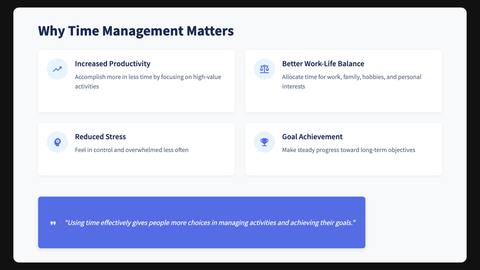}\\[3pt]
    \includegraphics[width=0.32\textwidth]{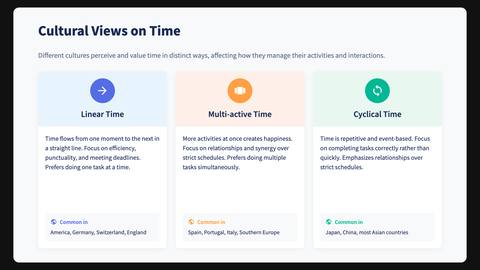}
    \includegraphics[width=0.32\textwidth]{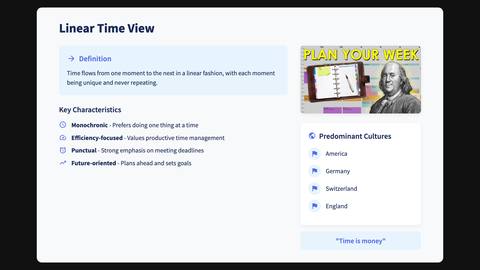}
    \includegraphics[width=0.32\textwidth]{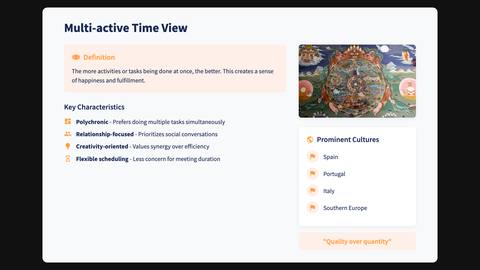}
    
    \textbf{(i) Zhipu}
    \end{minipage}
    
    \caption{Comparison of slides generated on ``Time Management'' (Part 3 of 3). Each subfigure shows 6 consecutive slides (2 rows $\times$ 3 columns) from a single product.}
    \label{fig:time-management-comparison-3}
\end{figure*}

Figures~\ref{fig:5g-comparison-1}--\ref{fig:time-management-comparison-3} present qualitative comparisons of slides generated by Gamma, Kimi-Banana, Kimi-Smart, Kimi-Standard, NotebookLM, Quake, Skywork, Skyworks-Banana, and Zhipu across three topics: 5G Technology (high complexity), Art Therapy (medium complexity), and Time Management (low complexity). 

% Figure 1: 5G Topic Comparison - Part 1

% Figure 2: 5G Topic Comparison - Part 2

% Figure 3: 5G Topic Comparison - Part 3

% Figure 4: Art Therapy Topic Comparison - Part 1

% Figure 5: Art Therapy Topic Comparison - Part 2

% Figure 6: Art Therapy Topic Comparison - Part 3

% Figure 7: Time Management Topic Comparison - Part 1

% Figure 8: Time Management Topic Comparison - Part 2

% Figure 9: Time Management Topic Comparison - Part 3

\end{document}